\documentclass[runningheads]{llncs}

\usepackage{eccv}

\usepackage{eccvabbrv}
\usepackage{supplement}

\usepackage{graphicx}
\usepackage{booktabs}

\usepackage[accsupp]{axessibility}  

\usepackage[dvipsnames]{xcolor}

\usepackage{mathtools}
\usepackage{overpic}
\usepackage{xcolor,colortbl}
\usepackage{subcaption}
\usepackage{algpseudocode}
\usepackage{algorithm}
\usepackage{nicefrac}

\usepackage{pict2e}

\usepackage{multirow} 
\usepackage{arydshln} 
\newcommand{\specialcell}[2][c]{%
	\begin{tabular}[#1]{@{}c@{}}#2\end{tabular}}

\usepackage{bm}

\usepackage{comment}

\usepackage{pifont} 
\usepackage{bbm}
\usepackage{dsfont}
\usepackage{wrapfig}

\newcommand{\manifold}{\mathcal{M}}
\newcommand{\tangspace}[1]{\mathcal{T}_{#1}\manifold}
\newcommand{\metrictensor}{V}
\newcommand{\mdist}{\text{dist}}
\newcommand{\innerprod}{v}

\newcommand{\counterfactual}{\mathcal{I}}

\newcommand{\conceptlog}{z}

\newcommand{\userfkt}{{\rho}}

\newcommand{\diag}{\mathop{\rm diag}}

\newcommand{\norm}[1]{\left\lVert#1\right\rVert}

\newcommand{\pdf}{p}

\newcommand{\rvxr}{{x}}
\newcommand{\rvyr}{{y}}
\newcommand{\rvzr}{{z}}

\newcommand{\ind}{\mathds{1}}

\newcommand{\indicator}[1]{\ind\left\{#1\right\}}

\newcommand{\covMatP}{ \Sigma}
\newcommand{\mean}{\mu}

\newcommand{\param}{\theta}
\newcommand{\paramA}{\phi}

\newcommand{\paramC}{\pi}
\newcommand{\paramD}{\tau}

\newcommand{\lag}{\lambda}

\newcommand{\expect}[2]{\mathbb{E}_{#1}\!\left[#2\right]}

\newcommand{\KL}[2]{{K\! L}\!\left(#1\vert\vert#2\right)}

\newcommand{\discriminantfkt}{f}
\newcommand{\discriminantfktb}{F}

\newcommand{\logitfkt}[1]{\log \frac{#1}{1-#1}}

\newcommand{\class}{c}

\newcommand{\target}{y}

\newcommand{\dimdata}{M}
\newcommand{\dimdatab}{N}
\newcommand{\dimlatentprior}{L}

\newcommand{\numdata}{D}
\newcommand{\numiter}{T}
\newcommand{\iter}{t}
\newcommand{\numsamples}{S}
\newcommand{\samples}{s}

\newcommand{\numclasses}{K}
\newcommand{\idxclass}{k}

\newcommand{\idxdata}{d}

\newcommand{\idxa}{i}
\newcommand{\idxb}{j}

\newcommand{\encoder}{h}
\newcommand{\decoder}{g}

\newcommand{\varpdf}{q}

\newcommand{\weight}{w}

\newcommand{\bias}{b}

\newcommand{\loss}{\mathcal{L}}

\newcommand{\lossC}{\mathcal{J}}

\newcommand{\categorical}{\mathrm{Cat}}

\newcommand{\gaussian}{\mathcal{N}}

\DeclareMathOperator*{\argmin}{arg\,min}
\DeclareMathOperator*{\argmax}{arg\,max}

\newcommand{\crefSubFigRef}[2]{\crefformat{figure}{Fig.~##2##1{#2}##3}%
  \cref{#1}\crefformat{figure}{Fig.~##2##1##3}}

\usepackage{hyperref}

\usepackage{orcidlink}

\begin{document}

\title{The Gaussian Discriminant Variational Autoencoder (GdVAE): A Self-Explainable Model with Counterfactual Explanations}

\titlerunning{The Gaussian Discriminant Variational Autoencoder}

\author{Anselm Haselhoff\inst{1,2}\orcidlink{0009-0007-9489-6395} \and
Kevin Trelenberg\inst{1} \and\\
Fabian Küppers\inst{3}\orcidlink{0009-0005-9856-7527} \and 
Jonas Schneider\inst{3}
}

\authorrunning{A.~Haselhoff et al.}

\institute{TrustIn.AI Lab, Ruhr West University of Applied Sciences, Germany \and
TML Lab, The University of Sydney, Australia, \quad 
\inst{3}\;\,e:fs TechHub GmbH, Germany\\
\email{\{name.surname\}@hs-ruhrwest.de}, \email{\{name.surname\}@efs-techhub.com}\\
\url{https://trustinai.github.io/gdvae}
}

\maketitle

\begin{figure}[ht]
	\centering
\begin{overpic}[grid=false, width=0.87\columnwidth]{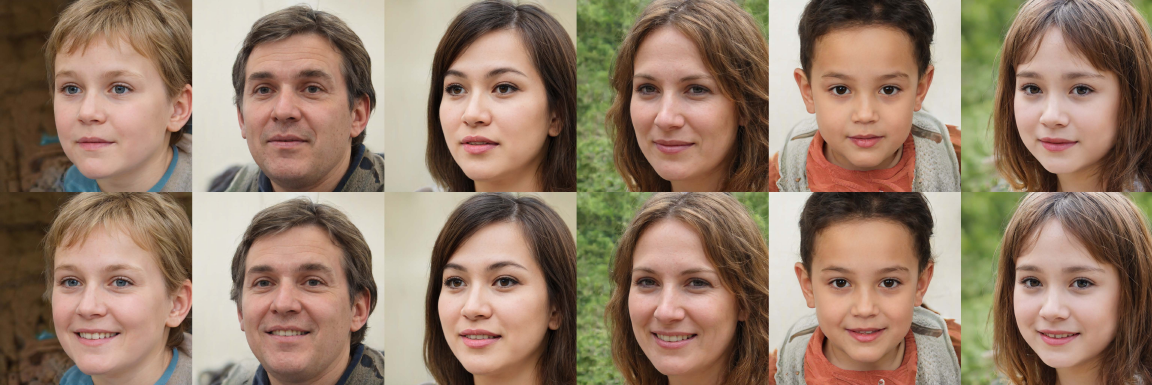}%
\put(1,30) {\textcolor{white}{\footnotesize{$\bf {x}^{\star}$}}}%
\put(1,12.5) {\textcolor{white}{\footnotesize{$\bf {x}^{\delta}$}}}%
\end{overpic}	
\caption{FFHQ high-resolution  (1024$\times$1024) counterfactuals ${x}^{\delta}$ for smiling.
}\label{fig:FFHQCFs}
\end{figure}
\begin{abstract}
Visual counterfactual explanation (CF) methods modify image concepts, \eg, shape, to change a prediction to a predefined outcome while closely resembling the original query image. Unlike self-explainable models (SEMs) and heatmap techniques, they grant users the ability to examine hypothetical "what-if" scenarios. 
Previous CF methods either entail post-hoc training, limiting the balance between transparency and CF quality, or demand optimization during inference.
To bridge the gap between transparent SEMs and CF methods, we introduce the GdVAE, a self-explainable model based on a conditional variational autoencoder (CVAE), featuring a Gaussian discriminant analysis (GDA) classifier and integrated CF explanations. Full transparency is achieved through a generative classifier that leverages class-specific prototypes for the downstream task and a closed-form solution for CFs in the latent space. 
The consistency of CFs is improved by regularizing the latent space with the explainer function.  
Extensive comparisons with existing approaches affirm the effectiveness of our method in producing high-quality CF explanations while preserving transparency. {Code and models are public.}
{}
\end{abstract}
\section{Introduction}\label{sec:intro} 
Deep neural networks (DNNs), such as generative adversarial networks (GANs) for image generation~\cite{StyleGan8953766} and DNN classifiers~\cite{pmlr-v97-tan19a}, have achieved notable success.
However, they suffer from limited interpretability, often being considered black boxes with decision processes not well understood by humans.

\textit{Generative explanation methods} identify meaningful latent space directions related to independent factors of variation (\eg, shape). Previous work finds these directions by enforcing disentanglement during training or analyzing the latent space \cite{StyleGan8953766, NEURIPS2021_bfd2308e, NEURIPS2018_1ee3dfcd, higgins2017betavae, Ding_2020_CVPR, Ren2021LearningDR, Bau2019, Shen2020InterpretingTL}. Explanations are obtained by visualizing the effect of changes in the latent space. Generative models are also used in \textit{counterfactual} (CF) reasoning, which answers questions like, "How can the example be changed to belong to category B instead of A?". This allows users to explore hypothetical "what-if" scenarios \cite{ghandeharioun2022dissect}. Recent advances combine generative models and classifiers to generate CF explanations, with enhanced techniques focusing on realism and consistency \cite{Lang_2021_ICCV, Singla2020Explanation, Goetschalckx_2019_ICCV, Samangouei18ExplainGAN,Khorram_2022_CVPR, ghandeharioun2022dissect, ecinn2021}. However, many methods lack transparency, as the CF generation often relies on a separate black-box model, and the classifier itself may not guarantee transparency either. 

\textit{Self-explainable models (SEMs)} provide explanations alongside their predictions without the need for post-hoc training \cite{Melis18SENN, NEURIPS2019_ProtoPNet, Gautam22ProtoVAE, Agarwal21NAM, chang2022nodegam}. Many SEMs are based on prototype learning, using these transparent and often visualizable prototypes as a bottleneck in a white-box classifier. This white-box classifier (\eg, linear predictor) is optimized end-to-end. However, generating CFs for these models is only feasible through post-hoc methods, potentially reducing transparency.

To bridge the gap between transparent SEMs and CF methods, we introduce GdVAE, a conditional variational autoencoder (CVAE) designed for transparent classification and CF explanation tasks. Full transparency is achieved with a generative classifier using class-specific prototypes and a closed-form solution for CFs in the latent space, inspired by Euclidean and Riemannian manifold perspectives. The prototype explanations come from the distributions provided by the CVAE's prior network, meaning the classifier has no additional trainable parameters. We solve the inference problem of the CVAE, which involves unknown classes, using expectation maximization that iteratively uses the classifier. Finally, we generate local CF explanations in the latent space using a transparent linear function that supports user-defined classifier outputs, and then use the decoder to translate them back to the input space. Joint training of the classifier and generative model regularizes the latent space for class-specific attributes, enabling realistic image and CF generation. An additional regularizer ensures consistency between query confidence and true confidence of the classified CF.

\textit{In summary, our contributions are:} (i) We introduce a SEM for vision applications, based on a CVAE, with an intrinsic ability to generate CFs; (ii) We offer global explanations in the form of prototypes directly utilized for the downstream task, visualizable in the input space; (iii) We provide transparent, realistic, and consistent local CF explanations, allowing users to specify a desired confidence value; (iv) We conduct a thorough comparative analysis of our method, analyzing performance, consistency, proximity, and realism on common vision datasets.

\section{Related Work}\label{sec:sota}
Since our work is a SEM with integrated visual CF explanations, we begin by outlining the categorization criteria. Subsequently, we review generative and CF explanations, as well as prototype-based SEMs tailored for vision tasks.
Generative models naturally serve as an integral component of an explainer function used for generating CF images. Typically, this function is learned through probing the classifier and optimizing it for specific properties.
In CF research, while various properties are discussed, \textit{realism}, \textit{proximity}, and \textit{consistency} stand out as widely accepted criteria.
To simplify, CFs should resemble natural-looking images (\textit{realism}), make minimal changes to the input (\textit{proximity}), and maintain query confidence consistency with the classifier's predictions when used as input (\textit{consistency}) \cite{Singla2020Explanation, ghandeharioun2022dissect,Khorram_2022_CVPR,black2022consistent}.
Similarly, in prototype-based SEMs, transparency is crucial, characterized by the visualization of prototypes (PT) in the input space and their utilization in a white-box classifier \cite{Gautam22ProtoVAE}. To align our work with CF methods and SEMs, we adopt the following predicates.
\begin{enumerate}
    \item \textit{Realism:} CFs should stem from the data manifold with a natural appearance.
    \item \textit{Consistency:} The explainer function  $\counterfactual_\discriminantfktb(\rvxr,\delta): (\mathbb{R}^\dimdatab,\mathbb{R})\rightarrow\mathbb{R}^\dimdatab$ should be conform with the desired classifier output $\discriminantfktb(\rvxr^{{\delta}})\approx \discriminantfktb(\rvxr)-\bar{\delta}=\delta$, where  $\bar{\delta}$ is the desired perturbation of the output function, $\delta$ the desired output, and $\rvxr^{{\delta}}=\counterfactual_\discriminantfktb(\rvxr,{\delta})$ the CF for the input $\rvxr$\;\cite{Singla2020Explanation}.
    \item \textit{Proximity}: The CF should minimally change the input.
    {\item \textit{Transparency:} A model should use explanations (e.g., prototypes) as intrinsic parts of a white-box predictor, and they should be visualizable in input space.}
\end{enumerate}

\noindent \textbf{Generative Explanations (a).} The first group of approaches aims to explain pre-trained generative models (\eg, GANs). Directions for interpretable control can be derived through unsupervised \cite{haerkoenen2020ganspace, voynov2020unsupervised, Plumerault2020Controlling, Esser2020} or supervised \cite{Goetschalckx_2019_ICCV, yang_semantic_2021,Shen2020InterpretingTL} analysis of generative models. {GANalyze} \cite{Goetschalckx_2019_ICCV} employs a pre-trained classifier to learn linear transformations in the latent space, whereas \cite{Shen2020InterpretingTL} directly use a linear classifier in the latent space to define the direction. Except for UDID \cite{voynov2020unsupervised}, all mentioned methods use linear explainer functions for manifold traversal.
Most of these methods, due to their linear explainer function, provide transparency in latent space manipulation. Transparent classification and CF generation aren't their primary focus, though they can generate CFs without optimizing for factors like \textit{realism}. {Our method aligns with these post-hoc methods by using a transparent linear explainer function for CF generation. In contrast, our approach excels by more effectively regularizing the latent space through end-to-end training.}

\noindent\textbf{Visual Counterfactual Explanations (b).} 
{The second category of methods focuses on CF generation, optimizing \textit{realism}, \textit{proximity}, and \textit{consistency}}. EBPE \cite{Singla2020Explanation} and its extension \cite{ghandeharioun2022dissect} explain pre-trained classifiers by using a GAN to generate CF images with user-defined confidence values. Similarly, works like \cite{Samangouei18ExplainGAN, Khorram_2022_CVPR, Lang_2021_ICCV, Jeanneret_2022_ACCV,jacob2022steex,ecinn2021}, train generators with a simpler consistency task, where the user pre-defines the class label only, without specifying the confidence. DiME \cite{Jeanneret_2022_ACCV} optimizes CFs iteratively, incurring significant computational costs. Unlike other methods, C3LT \cite{Khorram_2022_CVPR} only manipulates the latent space with neural networks, similar to methods in the first category, requiring access to a pre-trained generative model. A different line of research \cite{pmlr-v97-goyal19a,vandenhende2022making} seeks to replace image regions based on distractor images of the CF class. 
In \cite{Lang_2021_ICCV} and \cite{ecinn2021}, the classifier and generator are closely coupled during training to enforce a latent space that encodes class-specific information. StylEx \cite{Lang_2021_ICCV}, like \cite{Jeanneret_2022_ACCV}, requires time-consuming inference-time optimization and classifier probing to identify influential coordinates for each input image. In contrast, ECINN \cite{ecinn2021} is unique in its use of a transparent linear explainer function and an invertible model. Our method is closely related to ECINN, with the distinction that they require a post-hoc analysis of the training data to determine the parameters of the explainer function. Consequently, unlike our model, they approximate the true decision function of their classifier for CF generation, resulting in a loss of transparency. In contrast, all the other methods described employ complex DNNs for CF generation and the classifier, limiting their transparency. {Our approach mirrors these CF generation processes but stands out with a transparent, linear explainer function analytically linked to our white-box classifier's decision function.}

\begin{table}[t!]
\caption{Comparison of explanation methods. "Design" column groups approaches according to the headings: (a), (b), and (c). 
The symbol $\sim$ indicates that most methods use a transparent linear function for latent space traversal and may not be explicitly designed for generating CFs. 
Explanations are categorized into Counterfactuals ("CF") and Prototype-based ("PT"). $\dagger$: some works \cite{NEURIPS2019_ProtoPNet,Wang21TesNET} use alternating optimization.}\label{tab:categorize}
\begin{center}
\begin{tabular}{c|c|c|m{7mm}m{7mm}|c}
      \hline
{\multirow{2}{*}{Design}} & \multirow{2}{*}{Approach} & \multirow{2}{*}{Transparency} & \multicolumn{2}{c|}{Explanation} & \multicolumn{1}{c}{\multirow{2}{*}{Optimization}} \\ 
                      &                           &                               & 
                      \multicolumn{1}{c}{CF}            &  \multicolumn{1}{c|}{PT}           &                              \\ \hline
(a)  &  \parbox{3.5cm}{\centering \cite{voynov2020unsupervised, haerkoenen2020ganspace,Esser2020,Plumerault2020Controlling,Goetschalckx_2019_ICCV,yang_semantic_2021,Shen2020InterpretingTL} }     &       $\sim$                    &         \multicolumn{1}{c}{  $\sim$}     &                  &   post-hoc                                             \\ 
(b)   &  \parbox{3.6cm}{\centering  \cite{Khorram_2022_CVPR,ghandeharioun2022dissect,Singla2020Explanation,Samangouei18ExplainGAN,Lang_2021_ICCV,ecinn2021,Jeanneret_2022_ACCV,jacob2022steex}}        &                               &         \multicolumn{1}{c}{\checkmark}      &                  &   post-hoc                                             \\ 
(c) &  \cite{NEURIPS2019_ProtoPNet,Wang21TesNET,Gautam22ProtoVAE}                       &         \checkmark                  &               &           \multicolumn{1}{c|}{\checkmark }       &      end-to-end$^\dagger$ 
                                            \\ 
                                            \hline
  & \textbf{GdVAE} (ours) & \checkmark &  \multicolumn{1}{c}{\checkmark} &  \multicolumn{1}{c|}{\checkmark} & end-to-end    \\
          \hline
\end{tabular}
\end{center}
\end{table}

\noindent\textbf{Self-explainable Models (c).} The classifier and CF generation of our GdVAE are closely tied to the same prototypical space. A line of works that comprises this prototype-based learning can be found in SEM research \cite{Melis18SENN,FlintParekh2020AFT, NEURIPS2021_SITE, NEURIPS2019_ProtoPNet,Wang21TesNET, Gautam22ProtoVAE,CVNet2023}. In \cite{Gautam22ProtoVAE}, a categorization of SEMs was introduced, and our specific focus is on methods prioritizing the \textit{transparency} property \cite{NEURIPS2019_ProtoPNet, Wang21TesNET, Gautam22ProtoVAE}. To maintain interpretability, these SEMs employ similarity scores that measure the likeness between features and prototypes within the latent space. Afterwards, these scores are employed within a linear classifier, which encodes the attribution of each prototype to the decision. Unlike ProtoPNet  \cite{NEURIPS2019_ProtoPNet} and TesNET \cite{Wang21TesNET}, ProtoVAE \cite{Gautam22ProtoVAE} uses end-to-end training, utilizing a model capable of decoding learned prototypes, resulting in a smooth and regularized prototypical space. 

Our GdVAE employs one prototype per class with a linear Bayes’ classifier, implicitly utilizing Mahalanobis distance instead of a 2-norm-based similarity. Unlike ProtoVAE, our SEM enhances transparency and CF generation, unifying
these research areas effectively. Refer to \cref{tab:categorize} for an overview.

\section{Method}\label{sec:method}
\textbf{Notation.} We address a supervised learning problem with input samples $\rvxr\in \mathbb{R}^\dimdatab$ (\eg, images) and class labels $\rvyr\in \{1, \ldots, \numclasses\}$. The latent variable $\rvzr\in \mathbb{R}^\dimdata$ is used for both autoencoding and classification. {Model parameters $\param$ and $\paramA$ define the neural networks (NNs) for probabilistic models. For example, we use a Gaussian posterior $\varpdf_\paramA(\rvzr|\rvxr,\rvyr)= \gaussian\left(\mu_\rvzr(\rvxr,\rvyr;\paramA), \Sigma_\rvzr(\rvxr,\rvyr;\paramA)\right)$, with $\mu_\rvzr(\rvxr,\rvyr;\paramA)$ and $\Sigma_\rvzr(\rvxr,\rvyr;\paramA)$ as NNs}. In discussions involving encoders and decoders, we omit the class input $\rvyr$ for simplicity and employ shorthand notations for encoders and decoders, such as $\encoder(\rvxr)=\mu_\rvzr(\rvxr;\paramA)$ and $\decoder(\rvzr)=\mu_\rvxr(\rvzr;\param)$. We express a probabilistic classifier for discrete variables as $\pdf_\param(\rvyr | \rvzr)$, which can be transformed into discriminant functions, denoted as $\discriminantfkt^{(\idxa)}(\rvzr) = \log \pdf_\param(\rvyr = \idxa | \rvzr)$. For the two-class problem we can use a single discriminant $\discriminantfkt(\rvzr)=\discriminantfkt^{(\class) }(\rvzr)-\discriminantfkt^{(\idxclass)}(\rvzr)$, where positive values correspond to class $\class$ and negative values to class $\idxclass$. The following explanation methods are discussed solely for the two-class problem. {The composition of the encoder $\encoder(\rvxr)$ and the discriminant $\discriminantfkt(\rvzr)$ can be used as an input-dependent discriminant function $\discriminantfktb(\rvxr)=(\discriminantfkt\circ\encoder)(\rvxr)$. Similarly, we can obtain CF images by generating CFs in the latent space with respect to $\discriminantfkt(\rvzr)$ and using the decoder to transform them into the image space $\counterfactual_\discriminantfktb(\rvxr,\delta)=(\decoder\circ\counterfactual_\discriminantfkt)(\rvzr,\delta)$.}
\begin{figure*}[!t]
	\centering
\includegraphics[width=0.92\textwidth]{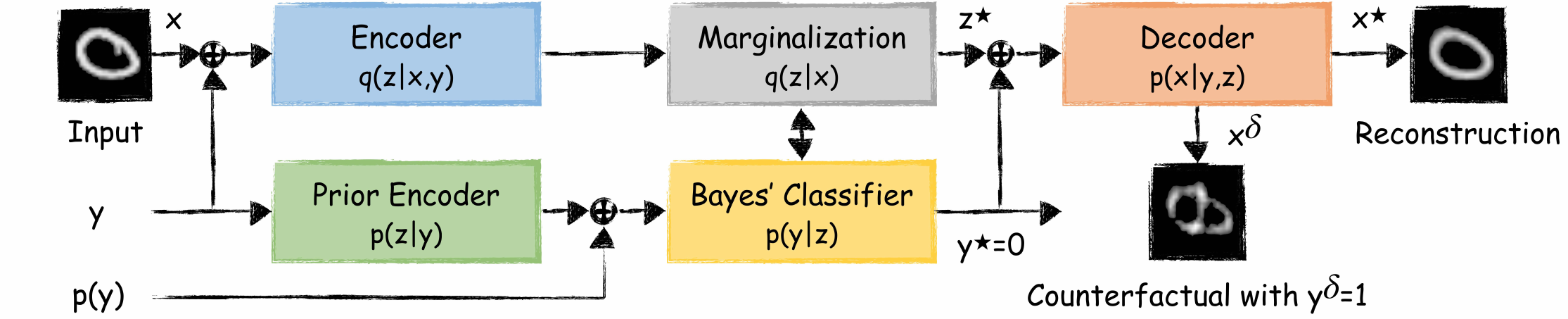}
\caption{{The GdVAE has three branches: \textit{1.) Feature Detection $\&$ Reconstruction:} The encoder, akin to a recognition network in a CVAE, generates latent code $\rvzr$. During inference, with an unknown class $\rvyr$, the marginal $\varpdf(\rvzr|\rvxr)$ acts as a feature detection module. The decoder reconstructs the input image $\rvxr$ using samples $\rvzr^\star$ from the marginal and $\rvyr^\star$ from the classifier. \textit{2.) Prior Encoder $\&$ Classifier:} The prior encoder learns the latent feature distribution independently of the input image, providing necessary distributions for the generative classifier. \textit{3.) Explanation:} During inference, the model generates a class prediction $\rvyr^\star$ and a latent variable $\rvzr^\star$. The user requests a CF by defining a desired confidence value and uses a linear function $\rvzr^\delta=\counterfactual_\discriminantfkt(\rvzr^\star,\delta)$ to modify $\rvzr^\star$ to $z^\delta$. The CF $\rvxr^\delta$ is obtained by transforming $z^\delta$ to image space using the decoder. The CF illustrates crossing the decision boundary, showing features of digits 0 and 1.}}
	\label{fig:overview}
\end{figure*}

\noindent\textbf{Overview.} The GdVAE enhances an autoencoder with an integrated generative classifier. We consider a generative model $\pdf_\param(\rvxr,\rvyr,\rvzr)=\pdf_\param(\rvxr|\rvyr,\rvzr) \pdf_\param(\rvyr,\rvzr)$ with two distinct factorizations for $\pdf_\param(\rvyr,\rvzr)=\pdf_\param(\rvzr|\rvyr)\pdf_\param(\rvyr)=\pdf_\param(\rvyr|\rvzr)\pdf_\param(\rvzr)$, defining coupled processes. {The first factorization establishes a class conditional prior $\pdf_\param(\rvzr|\rvyr)$ for the latent variable $\rvzr$ and delineates an autoencoder (M1), while the second integrates a discriminative classifier $\pdf_\param(\rvyr|\rvzr)$ (M2) using the latent variable.} Later, we'll employ a generative classifier using the prior encoder's mean values as decision prototypes that will benefit from the discriminative learning signal. See an overview and description in~\cref{fig:overview}. 

\subsection{Autoencoding and Generative Classification}\label{sec:VAEandGDA}
\noindent\textbf{Model Distributions.} \textit{CVAE including a class prior (M1):} For the first factorization of $\pdf_\param(\rvxr,\rvyr,\rvzr)$ we assume the observed variable $\rvxr$ to be generated from the set of latent variables $\rvzr$ and $\rvyr$ through the following process
\begin{align}
	\rvyr & \sim \pdf_\param(\rvyr)=\categorical_\rvyr\!\left(\paramC(\param)\right),\label{eq:model11}\\
	\rvzr|\rvyr & \sim \pdf_\param(\rvzr|\rvyr) = \gaussian\left(\mu_\rvzr(\rvyr;\param), \Sigma_\rvzr(\rvyr;\param)\right), \label{eq:model12}\\
	\rvxr|\rvyr,\rvzr &\sim \pdf_\param(\rvxr|\rvyr,\rvzr) = \gaussian\left(\mu_\rvxr(\rvyr,\rvzr ;\param), \Sigma_\rvxr(\rvyr,\rvzr ;\param)\right),
	\label{eq:model13}  
\end{align}
with categorical distribution $\categorical_\rvyr\!\left(\paramC(\param)\right)\!=\!\prod_{k=1}^{K} \paramC(\param)_k^{\ind\{{\rvyr=k}\}}$, where $\paramC$ is a probability vector and $\ind\{{\cdot}\}$ is the indicator function. This process defines a CVAE~\cite{NIPS2015_CVAE} with an added class prior $\pdf_\param(\rvyr)$, capturing class frequency. Thus, we capture both a prior encoder $\pdf_\param(\rvzr|\rvyr)$ and a class prior, which are used by our classifier. 

\textit{GDA model with latent prior (M2):}
The second factorization of $\pdf_\param(\rvxr,\rvyr,\rvzr)$ describes our classification model, where the target class $\rvyr$ (observable during training) is generated by the latent code $\rvzr$ according to our second process
\begin{align}
	\rvzr & \sim \pdf(\rvzr)=\gaussian\left(0, I\right), \label{eq:model21}\\
	\rvyr|\rvzr & \sim \pdf_\param(\rvyr|\rvzr) =\categorical_\rvyr\!\left(\paramD(\rvzr; \param)\right), \label{eq:model22}\\ 
	\rvxr|\rvyr,\rvzr &\sim  \pdf_\param(\rvxr|\rvyr,\rvzr) = \gaussian\left(\mu_\rvxr(\rvyr,\rvzr ;\param), \Sigma_\rvxr(\rvyr,\rvzr ;\param)\right).
	 \label{eq:model23}
\end{align}
Instead of using a separate NN to estimate $\paramD$, we reuse M1's distributions to obtain the categorical distribution $\pdf_\param(\rvyr|\rvzr)\!=\!\eta \pdf_\param(\rvzr|\rvyr)\pdf_\param(\rvyr)$, where $\eta$ is a normalization constant in the context of Bayes' theorem. In addition to this coupling, both models are jointly trained using a unified learning objective.

\noindent\textbf{Learning Objective.}\label{sec:objective}
Our generative models feature non-conjugate dependencies, making it intractable to maximize the conditional log-likelihood. Thus, we employ a surrogate posterior $\varpdf_\paramA(\rvzr|\rvxr,\rvyr)$ to approximate the true posterior $\pdf_\param(\rvzr|\rvyr)$~\cite{Kingma2014}. The surrogate, also called the recognition model, adapts the latent code distribution based on $\rvxr$. Instead of maximizing the log-likelihood $\log \pdf_\param(\rvxr,\rvyr)$ of our model, we use the evidence lower bound (ELBO) to define our loss. The resulting per sample loss for the GdVAE is $\loss^{gd}=\alpha 	\loss^{M1} + \beta 	\loss^{M2}$, with 
{\begin{align} 	
\loss^{M1}&\!=\!	-\expect{\rvzr,\rvyr\sim\varpdf_\paramA}{\log \pdf_\param(\rvxr|\rvyr,\rvzr)}+\KL{\varpdf_\paramA(\rvzr|\rvxr,\rvyr)}{\pdf_\param(\rvzr|\rvyr)} -\log \pdf_\param(\rvyr), \label{eq:lossgdvae}\\
\loss^{M2}&\!=\!-\expect{\rvzr,\rvyr\sim\varpdf_\paramA}{\log \pdf_\param(\rvxr|\rvyr,\rvzr)}+\KL{\varpdf_\paramA(\rvzr|\rvxr,\rvyr)}{\pdf(\rvzr)} -\expect{\rvzr\sim \varpdf_\paramA}{\log \pdf_\param(\rvyr|\rvzr)}.\label{eq:lossgdvae2}	
\end{align}}

\noindent$\alpha$ and $\beta$ control the balance between M1 and M2, and $K\!L$ denotes the Kullback-Leibler divergence. The derivation of the loss and ELBO can be found in 
the Supplement. Note that during inference, we cannot directly sample from the encoder $\varpdf_\paramA(\rvzr|\rvxr,\rvyr)$ since the class $\rvyr$ is unknown. Instead, we conduct ancestral sampling by first sampling from  $\varpdf_\paramA(\rvyr|\rvxr)$ and afterwards from $\varpdf_\paramA(\rvzr|\rvxr, \rvyr)$ to approximate $\varpdf_\paramA(\rvzr|\rvxr)$. To ensure coherence between the training and inference processes, we compute the expectations relative to $\varpdf_\paramA(\rvzr|\rvxr)$ and $\varpdf_\paramA(\rvzr,\rvyr|\rvxr)$ during training, respectively. This alignment enhances the accuracy of predictions.

\noindent\textbf{Marginalization.}\label{sec:marg} The training process is straightforward when labels are observable, and we can directly sample from the conditional encoder $\varpdf_\paramA(\rvzr|\rvxr,\rvyr)$. 
Likewise, during inference with the model, we require an estimate of $\rvzr$ given $\rvxr$ and $\rvyr$. The challenge here is that $\rvyr$ is unknown during inference. 

Therefore, we draw inspiration from semi-supervised learning~\cite{NIPS2014_d523773c}, employ a factorized probabilistic model $\varpdf_\paramA(\rvzr,\rvyr|\rvxr)=\varpdf_\paramA(\rvzr|\rvxr,\rvyr)\varpdf_\paramA(\rvyr|\rvxr)$ and perform a marginalization $\varpdf_\paramA(\rvzr|\rvxr)=\sum_{\rvyr=1}^\numclasses \varpdf_\paramA(\rvzr|\rvxr,\rvyr)\varpdf_\paramA(\rvyr|\rvxr)$.  In practice, besides the conditional encoder $\varpdf_\paramA(\rvzr|\rvxr,\rvyr)$, a classifier $\varpdf_\paramA(\rvyr|\rvxr)$ is needed. To avoid the need for sampling in the image space~\cite{NIPS2015_CVAE}, we initialize the classifier with the class prior {$\pdf_\param(\rvyr)$} and iteratively refine both the classifier and the latent feature model. This
\begin{wrapfigure}[15]{r}{0.57\textwidth}
\vspace{-0.75cm}
\begin{minipage}[t]{.55\textwidth}
\begin{algorithm}[H]
	\caption{An EM-based classifier}\label{algo:marg}
	\begin{algorithmic}
		\State $\varpdf_\paramA(\rvyr|\rvxr)  \gets \pdf_\param(\rvyr)$
		 \For{iterations $\iter \in \{1,\ldots,\numiter\}$}  
		\State \text{E-Step: Ancestral sampling for GMM}
		\State $\rvzr^{(\samples)} \sim \varpdf_\paramA(\rvzr|\rvxr)=\sum_{\rvyr=1}^\numclasses \varpdf_\paramA(\rvzr|\rvxr,\rvyr)\varpdf_\paramA(\rvyr|\rvxr)$   
		\State \text{E-Step: GDA classifier}
		\State $\pdf_\param(\rvyr|\rvzr^{(\samples)})\gets  \eta \pdf_\param(\rvzr^{(\samples)}|\rvyr) \pdf_\param(\rvyr)$ 
		\State  \text{M-Step: Assign mean confidence to $\varpdf$}
		\State $\varpdf_\paramA(\rvyr|\rvxr) \gets \pdf_\param(\rvyr|\rvzr)= \frac{1}{\numsamples} \sum_{\samples=1}^\numsamples \pdf_\param(\rvyr|\rvzr^{(\samples)})$ 
		\EndFor
\State \textbf{return} $\varpdf_\paramA(\rvyr|\rvxr)$ 
	\end{algorithmic}
\end{algorithm}%
\end{minipage}
\end{wrapfigure} 
expectation-maximization (EM) approach is detailed in \cref{algo:marg}, with a proof in the Supplement. In contrast to a standard EM for a Gaussian mixture model (GMM), where we usually estimate mean and covariance values, we employ the GMM to generate $S$ data samples $\rvzr^{(s)}$. Subsequently, we perform a soft assignment using the fixed classifier $\pdf_\param(\rvyr|\rvzr)$ and, akin to \cite{falck2021multifacet}, reestimate $\varpdf_\paramA(\rvyr|\rvxr)$. The closer our estimate aligns with the true class of $\rvxr$, the more samples $\rvzr^{(s)}$ we obtain from the correct class, as $\varpdf_\paramA(\rvzr|\rvxr,\rvyr)$ is weighted by $\varpdf_\paramA(\rvyr|\rvxr)$. 

The algorithm yields the classifier $\varpdf_\paramA(\rvyr|\rvxr)$, used in the learning objective to estimate $\varpdf_\paramA(\rvzr|\rvxr)$.
We perform ancestral sampling, initially drawing samples from $\varpdf_\paramA(\rvyr|\rvxr)$, then from $\varpdf_\paramA(\rvzr|\rvxr, \rvyr)$ to approximate $\varpdf_\paramA(\rvzr|\rvxr)$ (see \cref{algo:marg}).

\noindent\textbf{Generative Classifier.} The generative classifier is built upon a Gaussian discriminant analysis model (GDA) \cite{Haselhoff_2021_CVPR} and does not have any additional parameters. Its purpose is to transform the features $\rvzr$ from the recognition network and marginalization process into an interpretable class prediction. 

During the training of the entire GdVAE, the prior network learns the class-conditional mean $\mean_\rvzr(\rvyr;\param)=\mean_{\rvzr|\rvyr}$ and covariance $\covMatP_\rvzr(\rvyr;\param)=\covMatP_{\rvzr|\rvyr}$ as the parameters of our distribution $\pdf_\param(\rvzr|\rvyr) = \gaussian\left(\mu_\rvzr(\rvyr;\param), \Sigma_\rvzr(\rvyr;\param)\right)$. We assume conditional independence and decompose the likelihood as $\pdf _\param(\rvzr |\rvyr)=\prod_{\idxb=1}^{\dimdata} \pdf_\param (\rvzr_\idxb |\rvyr)$. In practice, this results in a diagonal covariance matrix $\covMatP_{\rvzr|\rvyr}=\diag\left(\sigma^2_{\conceptlog_1| \target},\ldots, \sigma^2_{\conceptlog_\dimdata| \target} \right)$. {We use this distribution to determine the likelihood values for the GDA classifier. The class prior $\pdf_\param(\target)$ can be learned either jointly or separately as the final component of the GDA model. Thus, we use the mean values as class prototypes and the covariance to measure the distance to these prototypes.} 

To infer the class, we apply Bayes' theorem using the detected feature $\rvzr$ from the recognition model $\pdf_\param(\rvyr=\idxa|\rvzr) = \eta \pdf_\param(\rvzr |\rvyr=\idxa) \pdf_\param(\rvyr=\idxa)$, with the normalizer $\eta$.
For the explanation method, we further assume equal covariance matrices $\covMatP_\rvzr$ (independent of $\rvyr$), yielding linear discriminants $\discriminantfkt^{(\idxa)}(\rvzr)={\weight^{(\idxa)}}^T\rvzr+\bias^{(\idxa)}$, where the weight and bias are given by $\weight^{(\idxa)}=\covMatP_{\rvzr}^{-1}\mean_{\rvzr|\idxa}$ and $\bias^{(\idxa)}=-\frac{1}{2}\mean_{\rvzr|\idxa}^T\covMatP_{\rvzr}^{-1} \mean_{\rvzr|\idxa} +\log \pdf_\param(\rvyr=\idxa)$. For two classes we get $\discriminantfkt(\rvzr)=\discriminantfkt^{(\class)}(\rvzr)-\discriminantfkt^{(\idxclass)}(\rvzr)=\weight^T\rvzr+\bias$.
\subsection{Counterfactual Explanations (CF)}\label{sec:CFExplanation} Instead of directly employing a DNN to define an explainer function $\rvxr^{{\delta}}=\counterfactual_\discriminantfktb(\rvxr,{\delta})$, we generate CFs in the latent space and visualize the outcome using the decoder $\counterfactual_\discriminantfktb(\rvxr,\delta)=\decoder(\counterfactual_\discriminantfkt(\rvzr,\delta))$. Since the discriminant $\discriminantfkt(\rvzr)=\weight^T\rvzr+\bias$ of our classifier is linear by construction, we will see that the optimal explainer function is also linear $\counterfactual_\discriminantfkt(\rvzr, \kappa) = \rvzr + \kappa\overline{\weight}$, where the latent vector is adjusted in the direction of $\overline{\weight} \in \mathbb{R}^\dimdata$. Here, $\kappa \in \mathbb{R}$---a tuning knob for data traversal---represents the strength of the manipulation. Our proposed CF methods are shown in \cref{fig:counter}.

\textit{1.) Local counterfactuals:}
A local explanation should meet both consistency and proximity properties. Therefore, the optimal CF $z^\delta$ minimizes the distance to the current instance $z$ while ensuring the decision function matches the requested value
$\delta$. This involves solving the following constrained optimization problem \begin{align}\label{eq:optim}
 \counterfactual_\discriminantfkt(\rvzr, \delta)=\argmin_{z^\delta} \; & \mdist(z^\delta,z), \quad
 \textrm{subject to} \;  f(z^\delta)=\delta,
\end{align}
where $\mdist(.,.)$ is a distance metric that guarantees \textit{proximity} and the constraint ensures \textit{consistency}. Regardless of whether we choose the common L2-norm \cite{Singla2020Explanation, Jeanneret_2022_ACCV} or a Riemannian-based metric (Mahalanobis distance) induced by VAEs \cite{NEURIPS2022GeometricVAE}, the solution to \cref{eq:optim} is a linear explainer function
\begin{align}
\counterfactual_\discriminantfkt(\rvzr,\delta)&={\rvzr}^{{\delta}}=\rvzr+\kappa \overline{\weight}\text{, with }
\kappa=\frac{\delta-\weight^T\rvzr-\bias}{\weight^T\overline{\weight}},
\label{eq:localExplainer}
\end{align}
where $\weight=\covMatP_{\rvzr}^{-1}(\mean_{\rvzr|\class}-\mean_{\rvzr|\idxclass})$ is the gradient direction of our discriminant. In this approach, any negative value of $\delta$ would lead to a change in the class prediction, and $\delta=0$ corresponds to both classes having equal probability. To simplify user interaction, one can specify the value in terms of a probability using the logit function, such that $\delta=\logitfkt{\pdf_\class}$ with $\pdf_\class=\pdf(\rvyr=\class|\rvzr^\delta)$. 

Using the L2-norm, we obtain the intuitive solution where $\overline{\weight}=\weight$ {(local-L2)}. The CF is generated by using the shortest path (perpendicular to the decision surface) to cross the decision boundary (see \cref{fig:counter}). 
The theoretical analysis on Riemannian manifolds \cite{NEURIPS2022GeometricVAE} shows that samples close in the latent space with respect to a Riemannian metric lead to close images in terms of the L2-norm, thus optimizing proximity. A Riemannian-based solution using the Mahalanobis distance is $\overline{\weight}=\covMatP_{\rvzr}\weight$ {(local-M)}. Training with a spherical covariance $\Sigma_z\!=\!\sigma^2I$ instead of $\covMatP_{\rvzr}\!=\!\diag\left(\sigma^2_{\conceptlog_1},\ldots, \sigma^2_{\conceptlog_\dimdata} \right)$ yields equivalent functions and therefore equal empirical results for both Riemannian and L2-based CFs. Proofs, assumptions, and implications for non-linear methods are provided in the Supplement.

\textit{2.) Global counterfactuals:} The second CF approach is to move directly in the direction of the prototype of the opposing class, termed the \textit{counterfactual prototype}. In this scenario, we take a direct path from our current input $\rvzr$ to the CF prototype $\mean_{\rvzr|\idxclass}$, defining the direction as $\overline{\weight}=(\mean_{\rvzr|\idxclass}-\rvzr)$, and reuse the local explainer function from \cref{eq:localExplainer}. 

The local approach minimizes input attribute changes (\textit{proximity}), while global explanations gradually converge to common CF prototypes to reveal the overall model behavior for a category of examples. Both methods maintain the \textit{consistency} property in the latent space. For \textit{realism}, we argue that transitioning directly to the CF prototype or minimizing a distance function is the most effective way to stay within the data distribution, resulting in a natural appearance.

\noindent \textbf{Consistency Loss.}
Our explainer function implicitly assumes that the encoder and decoder act as inverses of each other. Consequently, it is imperative to ensure that a reconstruction $\rvxr^\delta=\decoder(\rvzr^\delta)$, based on the latent representation $\rvzr^\delta=\counterfactual_\discriminantfkt(\encoder(\rvxr),\delta)$, results in a similar latent representation when encoded once more, i.e., $\encoder(\rvxr^\delta)\approx \rvzr^\delta$. This alignment is crucial to ensure that the classifier provides the desired confidence when a CF is used as input. To enforce this property, similar to \cite{Lang_2021_ICCV,Singla2020Explanation, sinha2021consistency}, we introduce a tailored consistency loss
\begin{align}
	\loss^{con}&=\expect{\pdf(\delta)}{\KL{\varpdf_\paramA(\rvzr|\rvxr^\delta)}{\varpdf_\paramA(\rvzr^\delta|\rvxr)}},
 \label{eq:consitency}
\end{align}
where the term addresses classification consistency for generated CF inputs. Essentially, we are probing latent values between the distributions $\pdf_\param(\rvzr|\rvyr=\class)$ and $\pdf_\param(\rvzr|\rvyr=\idxclass)$ to optimize for the consistency property. $\varpdf_\paramA(\rvzr^\delta|\rvxr)$ is obtained by applying the linear transformation of the explainer function $\counterfactual_\discriminantfkt(\rvzr,\delta)$ to $\varpdf_\paramA(\rvzr|\rvxr)$. In other words, we simply shift the mean value and keep the variance. {We use both global and local explainer functions to generate training samples.} $\pdf(\delta)$ defines the desired perturbation of the latent variable and we use $\pdf(\delta)=\mathcal{U}(-\varepsilon,\varepsilon)$, where $\varepsilon$ can be specified in terms of a probability. The final loss is then given by $\loss = \loss^{gd}+\gamma \loss^{con}$, where $\gamma$ controls the impact of the consistency regularizer. 

\section{Experiments}
The empirical evaluation aims to validate the performance of our model, focusing on two components: the predictive performance of the GdVAE and the quality of the CFs.
We present quantitative results of the predictive performance and CFs in \cref{sec:resultspredictive,sec:resultsCFEval}, along with qualitative results in \cref{sec:Resultsqualitative}. 
In the Supplement, we conduct a hyperparameter investigation covering all method parameterizations. This includes exploring the model balance between M1 and M2, consistency loss, and presenting additional quantitative and qualitative results.

\noindent\textbf{Datasets and Implementation.} We employ four image datasets: MNIST \cite{mnist}, CelebA \cite{celeba}, CIFAR-10 \cite{cifar10}, and the high-resolution dataset FFHQ \cite{StyleGan8953766}. Our neural networks are intentionally designed to be compact. For CelebA, the encoder has five convolutional layers and one linear layer for $\mu_\rvzr(\rvxr,\rvyr;\paramA)$ and $\Sigma_\rvzr(\rvxr, \rvyr;\paramA)$, which define the distribution $\varpdf_\paramA(\rvzr|\rvxr,\rvyr)$. The decoder's architecture is symmetrical to that of the encoder. Prior encoders use fully connected networks with four layers to compute $\mu_\rvzr(\rvyr;\param)$ and $\Sigma_\rvzr(\rvyr;\param)$, defining our distribution $\pdf_\param(\rvzr|\rvyr)$. All baseline methods employ identical backbones as the GdVAE, and when feasible, publicly available code was adjusted to ensure a fair comparison. See the Supplement for details on datasets, models, and metrics. 

\subsection{Evaluation of Predictive Performance}\label{sec:resultspredictive}

\noindent\textbf{Methodology.} For a trustworthy SEM, performance should align with the closest black-box model~\cite{Gautam22ProtoVAE}. Thus, the goal of this evaluation is not to outperform state-of-the-art results on specific datasets but to offer a relative comparison for the GdVAE architecture and various training methods. In all approaches, both the classifier and autoencoder are jointly trained, sharing the same backbone.

 \begin{table}[t]
 \caption{{Predictive performance: Importance sampling (IS), ProtoVAE, and a black-box baseline. Classifier accuracy (ACC) and mean squared error (MSE) of reconstructions (scaled by $10^2$) are reported. Mean values and standard deviations are from four training runs with different seeds. $\dagger$: incl. ProtoVAE's augmentation and preprocessing.}}
\label{tab:gdvaeEMvsIS}
	\centering
	\renewcommand{\arraystretch}{0.94}
		\begin{tabular}{l|cc|cc|cc}
        \hline
        \multirow{2}{*}{Method} & \multicolumn{2}{c|}{MNIST}& \multicolumn{2}{c|}{CIFAR-10}& \multicolumn{2}{c}{CelebA - Gender}\\
        	&  $ACC\%$ $\uparrow$& $MSE$ $\downarrow$&$ACC\%$ $\uparrow$& $MSE$ $\downarrow$& $ACC\%$ $\uparrow$& $MSE$ $\downarrow$\\
         \hline 
         IS \cite{NIPS2015_CVAE, classIncremental, NEURIPS2020_02ed8122}    & \textbf{99.0\scriptsize$\pm$0.08} & \textbf{1.04\scriptsize$\pm$0.01}& 55.0\scriptsize$\pm$0.59 & 2.45\scriptsize$\pm$0.03   			 &  {94.7}\scriptsize$\pm$0.44 & 1.77\scriptsize$\pm$0.08  \\ 
	\textbf{Ours}     & \textbf{99.0\scriptsize$\pm$0.11} & 1.10\scriptsize$\pm$0.04 & \textbf{65.1\scriptsize$\pm$0.78} & \textbf{1.71\scriptsize$\pm$0.02}&\textbf{96.7\scriptsize$\pm$0.13}& \textbf{0.91\scriptsize$\pm$0.01}    \\ 
		\cdashline{1-7}
		Baseline& 99.3\scriptsize$\pm$0.04 & 1.12\scriptsize$\pm$0.02   &{69.0}\scriptsize$\pm$0.54 & 1.45\scriptsize$\pm$0.01   	& {96.7}\scriptsize$\pm$0.26 & {0.82}\scriptsize$\pm$0.00   				\\ 
\hline
	ProtoVAE \cite{Gautam22ProtoVAE} & \textbf{99.1\scriptsize$\pm$0.17} & 1.51\scriptsize$\pm$0.23& {76.6\scriptsize$\pm$0.35} & {2.69\scriptsize$\pm$0.02} & {96.6\scriptsize$\pm$0.24} & {1.32\scriptsize$\pm$0.10} \\
 	\textbf{Ours}$^\dagger$     & 98.7\scriptsize$\pm$0.05 & \textbf{0.93\scriptsize$\pm$0.01} &\textbf{76.8\scriptsize$\pm$0.91} & \textbf{1.18\scriptsize$\pm$0.02} & 
    \textbf{96.8\scriptsize$\pm$0.04} & \textbf{0.71\scriptsize$\pm$0.01} \\ 
  \hline 
        \end{tabular}
\end{table}
\noindent\textbf{Baselines.} 
First, optimal performance for the selected architecture is established using a black-box model, comprising a jointly trained CVAE and classifier as the \textit{baseline}. Next, GdVAE's inference method is evaluated against the leading CVAE technique, \textit{importance sampling (IS)} \cite{NIPS2015_CVAE, classIncremental, NEURIPS2020_02ed8122}. Lastly, \textit{ProtoVAE} \cite{Gautam22ProtoVAE} is referenced as a prototype-per-class VAE benchmark.

\noindent\textbf{Results.}
The results in \cref{tab:gdvaeEMvsIS} indicate good generalization in classification and reconstruction across MNIST and CelebA. The GdVAE's EM-based inference achieves performance close to the optimal baseline with a separate classifier, except for CIFAR where there is a four-percentage-point gap in accuracy. Comparing our EM and the IS approach suggests that our method is more efficient for higher-dimensional images, benefiting from sampling in the lower-dimensional latent instead of image space. 
With data augmentation and normalization from ProtoVAE, GdVAE achieves comparable results to ProtoVAE.

\textit{Takeaway:} The inference procedure of our SEM closely matches the performance of a discriminative black-box model. Furthermore, our method consistently delivers competitive results to state-of-the-art approaches, particularly when applied to higher-dimensional images. The class-conditional GdVAE offers better reconstructions compared to ProtoVAE, the only unconditional model.

\subsection{Quantitative Evaluation of CF Explanations}\label{sec:resultsCFEval}
\noindent\textbf{Methodology.} 
The experiments aim to evaluate the quality of CFs regarding \textit{realism}, \textit{consistency}, and \textit{proximity}. \textit{Realism}, as defined in \cite{Khorram_2022_CVPR, ghandeharioun2022dissect} or data consistency \cite{Singla2020Explanation}, refers to the CF images being realistic and capturing identifiable concepts. To measure realism, we employ the Fr\'{e}chet Inception Distance (FID) \cite{Khorram_2022_CVPR, ghandeharioun2022dissect, Singla2020Explanation} as a common metric. Akin to \cite{Khorram_2022_CVPR}, \textit{proximity} is assessed using the mean squared error (MSE) between the CF and the query image. 

The \textit{consistency} property, also known as compatibility \cite{Singla2020Explanation} or importance \cite{ghandeharioun2022dissect}, is evaluated using mean squared error (MSE), accuracy (ACC), as well as the Pearson correlation coefficient. We create CFs for every image by requesting confidences within the range $\pdf_\class\in[0.05,0.95]$, with a step size of $0.05$. The metrics compare the expected outcome of the classifier $\pdf_\class$ (desired probability score of CFs) with the actual probability $\widehat{\pdf}_\class$ obtained from the classifier for the CF. 

\noindent\textbf{Baselines.} We employ methods from different designs (see \cref{sec:sota}) as baselines with shared backbones. 
To ensure a fair comparison, we slightly modify methods that originally tackle the simpler consistency task \cite{Khorram_2022_CVPR,ecinn2021} or those intended for unsupervised scenarios \cite{voynov2020unsupervised}, aligning them with the consistency defined in \cref{sec:sota}. First, we apply generative explanation methods, including GANalyze \cite{Goetschalckx_2019_ICCV} and UDID \cite{voynov2020unsupervised}, while utilizing our pre-trained GdVAE as an autoencoder and classifier. Second, we adapt post-hoc CF methods to be compatible with our GdVAE architecture. We adapt the method from ECINN \cite{ecinn2021} to approximate our classifier. C3LT \cite{Khorram_2022_CVPR} is trained to generate CFs for our GdVAE model using a non-linear explainer function instead of our linear one. Finally, EBPE \cite{Singla2020Explanation} is adjusted to train an encoder and decoder based on the GdVAE architecture and the pre-trained classifier. These approaches are compared to our CF methods.
\begin{table*}[!t]
\caption{{
Evaluation of CF explanations using Pearson correlation ($\rho_p$), ACC, and MSE (scaled by $10^2$) for consistency, Fr\'{e}chet Inception Distance (FID) for realism, and MSE (scaled by $10^2$) for proximity. Mean values and standard deviations are from four runs with different seeds. The first and second best results are \textbf{bolded} and \underline{underlined}.}}
\label{tab:consistency}
	\centering
		\renewcommand{\arraystretch}{0.94}
		\begin{tabular}{cwl{2.5cm}|wc{1.5cm}wc{1.5cm}wc{1.5cm}|wc{2.cm}|wc{2.cm}}
			\cline{2-7}  
        & \multicolumn{1}{l|}{\multirow{2}{*}{Method}} & \multicolumn{3}{c|}{Consistency} & \multicolumn{1}{c|}{Realism} &{Proximity}\\
         &	&   {$\rho_p$ $\uparrow$} & $ACC\%$ $\uparrow$& $MSE$ $\downarrow$ & $FID$ $\downarrow$ &{ $MSE$ $\downarrow$ } \\
		\cline{2-7}
		\parbox[t]{2.5mm}{\multirow{8}{*}{\rotatebox[origin=c]{90}{\scriptsize MNIST - Binary 0/1}}} 
		&GANalyze \cite{Goetschalckx_2019_ICCV}  & 0.84\scriptsize$\pm$0.04 & 5.5\scriptsize$\pm$1.3 & 6.75\scriptsize$\pm$1.27  & \underline{54.89\scriptsize$\pm$4.19} & {6.33\scriptsize$\pm$1.73}\\ 
		&UDID \cite{voynov2020unsupervised} & 0.85\scriptsize$\pm$0.01 & 1.2\scriptsize$\pm$0.3 &  8.82\scriptsize$\pm$0.18  & \textbf{38.89\scriptsize$\pm$2.01} & 7.44\scriptsize$\pm$0.81 \\ 
		&ECINN \cite{ecinn2021}  & {0.93\scriptsize$\pm$0.02} & {33.0\scriptsize$\pm$7.5} &  {1.76\scriptsize$\pm$0.81} & 87.25\scriptsize$\pm$12.63 & \textbf{3.47\scriptsize$\pm$0.75}\\ 
	&EBPE \cite{Singla2020Explanation}  & \textbf{0.97\scriptsize$\pm$0.01} & \textbf{44.6\scriptsize$\pm$4.3} &  \textbf{0.50\scriptsize$\pm$0.13} & 108.94\scriptsize$\pm$13.61 & {25.73\scriptsize$\pm$20.69}\\ 
		&C3LT \cite{Khorram_2022_CVPR} & 0.89\scriptsize$\pm$0.03  & 3.6\scriptsize$\pm$0.8 &  6.32\scriptsize$\pm$1.39& 57.09\scriptsize$\pm$10.78 & {5.83\scriptsize$\pm$1.47}\\
         & \textbf{Ours} (local-L2) &  \underline{0.95\scriptsize$\pm$0.00} & \underline{42.9\scriptsize$\pm$2.7} & {0.95\scriptsize$\pm$0.11} & 91.22\scriptsize$\pm$11.04 & {4.58\scriptsize$\pm$1.00}\\
         & \textbf{Ours} (local-M) &  \underline{0.95\scriptsize$\pm$0.01} & \textbf{44.6\scriptsize$\pm$2.5} & \underline{0.87\scriptsize$\pm$0.13} & 89.91\scriptsize$\pm$5.58 & \underline{4.10\scriptsize$\pm$0.37}\\
         \cdashline{2-7}
        &\textbf{Ours} (global)  & {0.97\scriptsize$\pm$0.01} & {54.2\scriptsize$\pm$4.0} &    {0.55\scriptsize$\pm$0.13}  & 125.45\scriptsize$\pm$11.32 & {6.23\scriptsize$\pm$0.53}\\ 
	\cline{2-7}
  	\parbox[t][][t]{2.5mm}{\multirow{8}{*}{\rotatebox[origin=c]{90}{\scriptsize CelebA - Smiling}}}
	&GANalyze \cite{Goetschalckx_2019_ICCV}  &  0.78\scriptsize$\pm$0.03 & 
    15.2\scriptsize$\pm$3.3 &  5.42\scriptsize$\pm$0.97 & 147.43\scriptsize$\pm$19.49 & 13.47\scriptsize$\pm$9.36\\  
		&UDID \cite{voynov2020unsupervised}  & {0.86\scriptsize$\pm$0.06} & 15.8\scriptsize$\pm$9.2 & 4.22\scriptsize$\pm$2.17 & 178.23\scriptsize$\pm$75.84 & 13.73\scriptsize$\pm$9.41\\  
       &ECINN \cite{ecinn2021}  & 0.72\scriptsize$\pm$0.21 & 21.3\scriptsize$\pm$9.6 &  5.68\scriptsize$\pm$4.32 &  {95.35\scriptsize$\pm$14.48} & {1.16\scriptsize$\pm$0.22}\\
        &EBPE \cite{Singla2020Explanation}  & \textbf{0.94\scriptsize$\pm$0.01}& \textbf{41.9\scriptsize$\pm$3.1} & \textbf{1.22\scriptsize$\pm$0.16}  & 191.67\scriptsize$\pm$20.51 & 1.54\scriptsize$\pm$0.06\\ 
		&C3LT \cite{Khorram_2022_CVPR}  & \underline{0.90\scriptsize$\pm$0.01} & 11.8\scriptsize$\pm$5.5&  3.94\scriptsize$\pm$0.66 & {101.46\scriptsize$\pm$11.56}& 3.97\scriptsize$\pm$0.86\\ 
    &\textbf{Ours} (local-L2) & 0.81\scriptsize$\pm$0.04 & {25.0\scriptsize$\pm$4.9} &  {3.65\scriptsize$\pm$1.06} & \textbf{85.52\scriptsize$\pm$2.37} & 
  \underline{0.99\scriptsize$\pm$0.02}\\
    &\textbf{Ours} (local-M) & 0.82\scriptsize$\pm$0.05 & \underline{25.7\scriptsize$\pm$5.1} &  \underline{3.51\scriptsize$\pm$1.05} & \underline{85.56\scriptsize$\pm$2.39} & 
    \textbf{0.92\scriptsize$\pm$0.03}\\
 \cdashline{2-7}
     &\textbf{Ours} (global)  & {0.89\scriptsize$\pm$0.01}  & {45.9\scriptsize$\pm$12.3} & {2.08\scriptsize$\pm$0.54} & 128.93\scriptsize$\pm$4.94 & 
    5.81\scriptsize$\pm$0.53\\
	\cline{2-7}
\end{tabular}
\end{table*}
\begin{figure*}[!t]
 	\centering
 	\begin{subfigure}{0.9\linewidth}
 		\centering
 		\begin{overpic}[grid=false, width=1.0\textwidth]{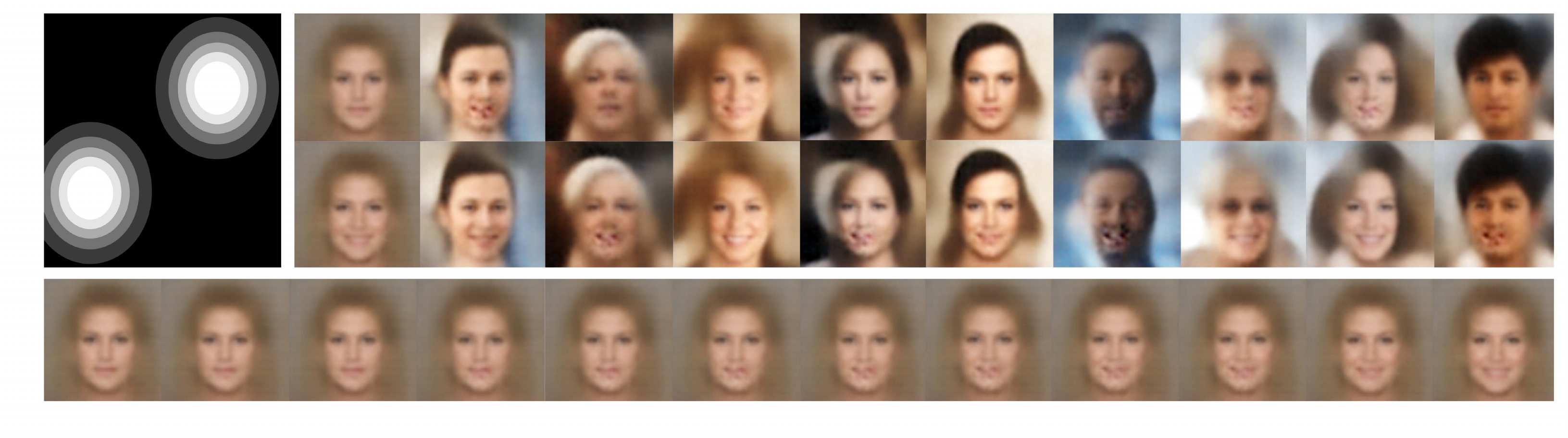}%
 		\put(5.1,27){\bf	\linethickness{0.35mm} \color{lightgray!100}\line(3,-5.2){9.4}}
        \put(3.5,24.7) {\color{white}  {a)}}%
        \put(19.5,24.7) {\color{white} {b)}}%
        \put(3.5,7.5) {\color{white} {c)}}%
     	\put(4.855,18.5) {\scriptsize \color{green}{\ding{72}}}%
 		\put(13.05,25.0) {\scriptsize\color{orange}{\ding{74}}}%
        \put(24.5,16.5){\scriptsize\color{green}{\ding{72}}}%
        \put(24.5,24.8){\scriptsize\color{orange}{\ding{74}}}%
        \put(5.745,15.8){\linethickness{0.3mm}          \color{red}\vector(0,1.0){3.}}
        \put(13.88,22.15){\linethickness{0.3mm}          \color{red}\vector(0,1.0){3.}}
        \put(2.5,16.) {\color{red}  { $\epsilon$}}%
        \put(10.8,22.8) {\color{red} { $\epsilon$}}%
        \put(13.4,21.85){\bf	\linethickness{0.35mm} \color{cyan}\vector(-3,-2.5){7.3}}
		\put(4.1,12.4) {\scriptsize\color{green}  { $\bf \mean_{\rvzr|s}$}}%
        \put(4.06,14.8) {\scriptsize\color{green}{ \ding{70}}}%
        \put(12.05,21.4) {\color{orange} \normalsize { $\bullet$}}%
        \put(12.0,19.0) {\scriptsize \color{orange} { $\bf\mean_{\rvzr|\bar{s}}$}}%
        \put(8.1,8.) {\scriptsize \color{orange}  \normalsize{$\bullet$}}%
        \put(95.7,7.9) {\scriptsize \color{green}{ \ding{70}}}%
        \put(18.9,19.1){\color{orange}\linethickness{0.5mm} \polygon(0,0)(80.1,0)(80.1,7.9)(0,7.9)}
        \put(18.9,18.7){\color{green}\linethickness{0.5mm} \polygon(0,0)(80.1,0)(80.1,-7.8)(0,-7.8)}
        \put(2.9,2.25){\color{orange}\linethickness{0.5mm} \polygon(0,0)(7.5,0)(7.5,7.8)(0.0,7.8)}
        \put(91.5,2.25){\color{green}\linethickness{0.5mm} \polygon(0,0)(7.5,0)(7.5,7.8)(0.0,7.8)}
        \put(10.8,2.25){\bf	\linethickness{0.5mm} \color{cyan}\vector(1,0){80.6}}
 	 	\put(40,-0.6) {\color{black}  \parbox{5in}{ $\rvxr^\delta=\decoder(\bar{\counterfactual}_\discriminantfkt(\mean_{\rvzr|\bar{s}},\delta)))$} }%
        \put(0.0,-0.6) {\color{black}  \parbox{5in}{ $\mean_{\rvxr|\bar{s}}=\decoder(\mean_{\rvzr|\bar{s}})$} }%
        \put(87.5,-0.6) {\color{black}  \parbox{5in}{ $\mean_{\rvxr|{s}}=\decoder(\mean_{\rvzr|{s}})$} }%
 	\end{overpic}
 \end{subfigure}
\caption{Regularized latent space. a) Distribution $\pdf_\param(\rvzr |\rvyr)$ with class-conditional mean values for not-smiling (orange, \textcolor{orange}{\large{$\bullet$}}) and smiling (green, \textcolor{green}{{\ding{70}}}), where $\rvyr=\bar{s}=not-smiling$ and $\rvyr=s=smiling$. b) Reconstructed random samples for not-smiling (top, orange \textcolor{orange}{\ding{74}}) and smiling (bottom, green \textcolor{green}{\ding{72}}), arranged in ascending order of their Mahalanobis distance from left to right. In each column, the Mahalanobis distance is made consistent by adding the same random vector $\epsilon$ (red vector in a) to the mean of both classes, aligning samples along isocontours.
c) The global explainer function interpolates between class-conditional means along the straight-line path (cyan arrow in a).}
 	\label{fig:LatentSpace}
 \end{figure*}
 
\noindent\textbf{Results.} The results in \cref{tab:consistency} reveal performance across diverse datasets in binary classification challenges. Considering that the GdVAE is the sole transparent model, it is essential to bear in mind that most models operate on the GdVAE's pre-regularized latent space (\cref{fig:LatentSpace}) when interpreting the results. Consequently, with the exception of EBPE, these methods face a less complex task and should approximate the "true" linear direction of our local CFs post-GdVAE training.
It becomes evident that our global CF method exhibits a higher degree of consistency with the classifier, albeit falling short in terms of realism when compared to the local {approaches}. This divergence is expected as the global method converges toward the mean representation of the CF prototype, producing relatively blurred representations with a notable distance from the query image (poor proximity). However, global CFs effectively uncover the model's overarching decision logic through its prototypes (see \crefSubFigRef{fig:LatentSpace}{c}). 

{Specifically, on the MNIST dataset, our local methods achieve the best or second-best results in all consistency metrics, producing CFs with well-calibrated confidence values.} The notably high accuracy values indicate that our methods generate CFs covering the entire confidence range, effectively capturing samples near the decision boundary. However, the realism metric is affected due to the absence of MNIST images near the decision boundary, notably those representing shared concepts of digits 0 and 1 (see \cref{fig:overview}). In summary, a favorable trade-off between consistency, realism, and proximity is achieved by EBPE, ECINN, and our local methods. A distinct perspective arises when considering the CelebA dataset, where our local {methods excel} in achieving optimal results for both proximity and realism, maintaining a low FID score. In terms of consistency metrics like ACC and MSE, our local {method (local-M) ranks} among the top two performers. Global CFs are distinguished with a separating line in \cref{tab:consistency}, indicating their deviation from query images by consistently approaching the same prototypes, which effectively reveals biases (e.g., toward female prototypes in CelebA, see \cref{sec:Resultsqualitative}).
\begin{figure}[!t]
  \centering
    \hfill
	\begin{subfigure}{0.47\linewidth}
 		\centering
 		\begin{overpic}[grid=false, width=1\linewidth]{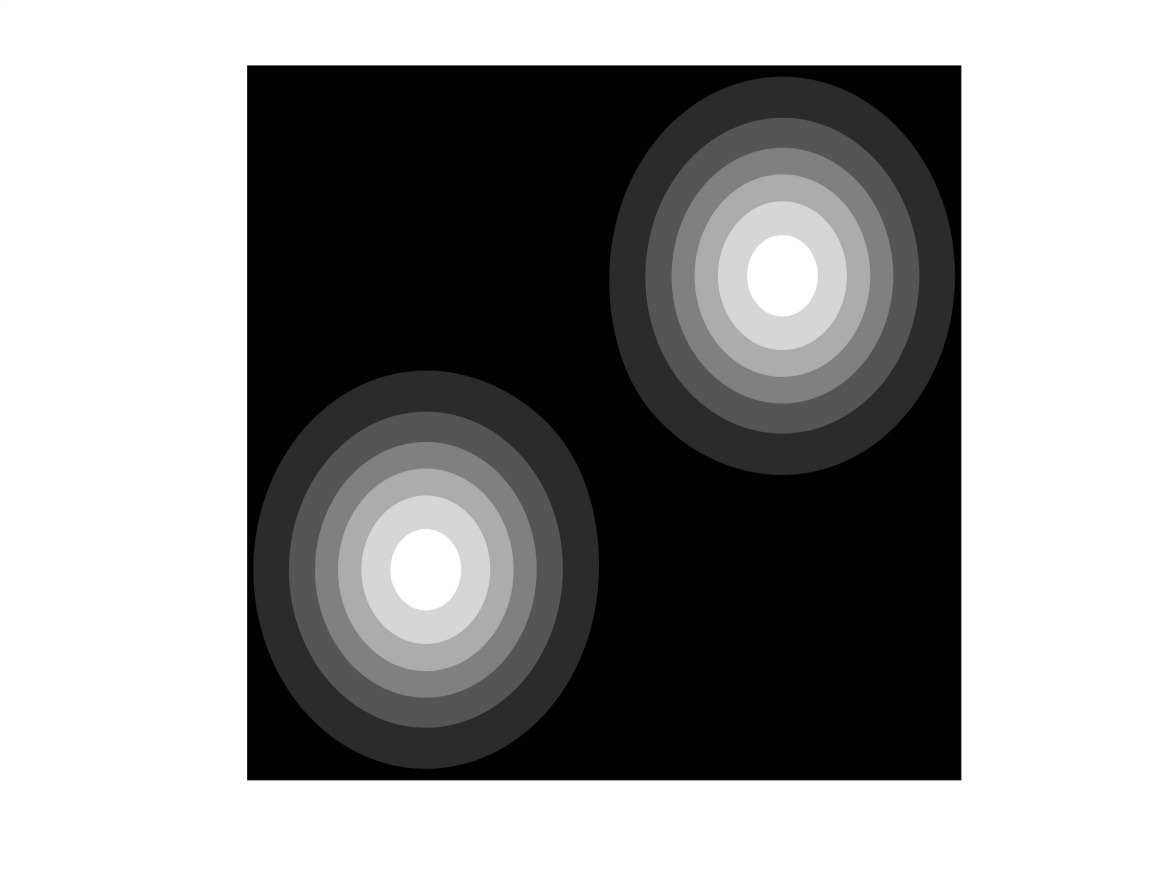}%
 		\put(34,69.5){\bf	\linethickness{0.6mm} \color{white!100}\line(3,-5.2){35.4}}
 		\put(48,21){\bf	\linethickness{0.4mm} \color{green!60}\vector(3,2){30}}
 		\put(73,32) {\color{white} \Large {$\bf \weight$}}%
 		\put(54.25,25.75) {\color{green}  { \ding{58}}}%
 		\put(62,25) {\color{green} \Large {$\bf {\rvzr}^{\delta}$}}%
 		\put(48,21){\bf	\linethickness{0.4mm} \color{cyan!60}\vector(3,4.75){18.8}}
 		\put(51.2,29.8) {\color{cyan}  { \ding{116}}}%
 		\put(53,38.2) {\color{cyan} \Large {$\bf \bar{\rvzr}^{\delta}$}}%
 		\put(46.9,19.9) {\color{lightgray!50} \large {$\bullet$}}%
 		\put(44,14) {\color{lightgray!50} \Large {$\bf \rvzr^\star$}}%
 		\put(30,35) {\color{blue} \large { $\bf\mean_{\rvzr|\class}$}}%
        \put(34.47,24.42) {\color{blue}\footnotesize{\ding{70}}}
 		\put(60,62) {\color{orange} \large { $\bf \mean_{\rvzr|\idxclass}$}}%
 		\put(65.35,49.61) {\color{orange} \large {$\bullet$}}%
 		\put(50,3) {\color{black} \large {$\rvzr_1$} }%
 		\put(11,39) {\color{black} \large {$\rvzr_2$} }%
 	\end{overpic}
  \caption{CF generation. Consider input $\rvxr$ with latent vector $\rvzr^\star$ and class prediction $\class$. Our first CF method, denoted as ${\rvzr}^{\delta}$ (green {\color{green}{\ding{58}}}), walks along the gradient direction $\weight$ {or the slightly rotated gradient direction $\Sigma_z\weight$ (omitted for clarity)}. The second method creates a CF, $\bar{\rvzr}^{\delta}$ (cyan {\color{cyan}{ \ding{116}}}), by moving from $\rvzr^\star$ to the prototype $\mean_{\rvzr|\idxclass}$ (orange \textcolor{orange}{\large$\bullet$}) of the contrasting class $\idxclass$. Here, we show a solution for $\delta=0$ where the decision boundary is crossed.}
  	\label{fig:counter}
  \end{subfigure}
  \hfill 
  	\begin{subfigure}{0.47\linewidth}
     \centering
 		\begin{overpic}[grid=false, width=0.95\textwidth]{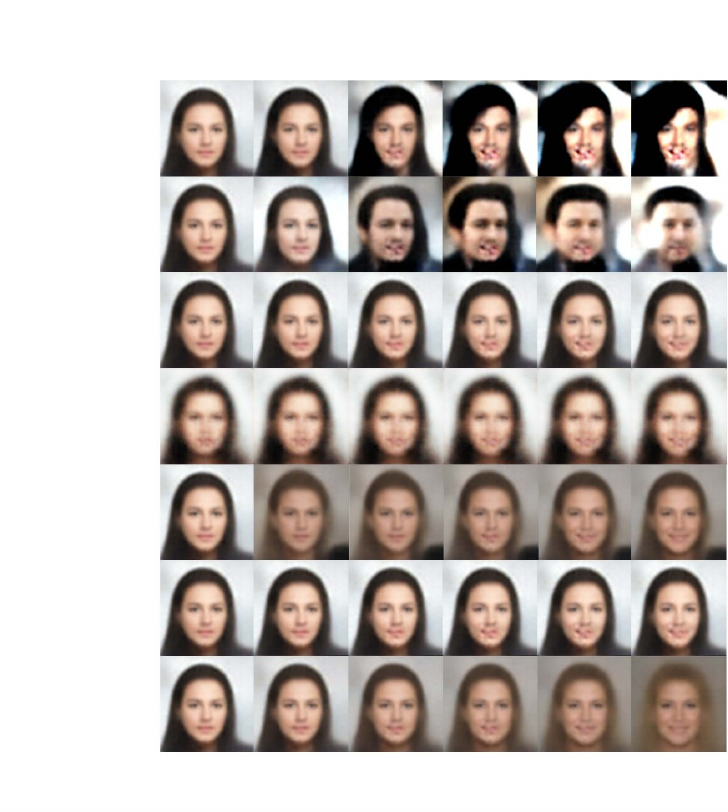}%
        \scriptsize	   	\put(19.5,7.5){\color{lightgray!100}\linethickness{0.4mm} \polygon(0,0)(12,0)(12,82.5)(0,82.5)}
        \put(1,82.5) {\color{black} \parbox{.5in}{GAN-\\alyze}}%
 		\put(1,70.75) {\color{black} {UDID} }%
 		\put(1,59.5) {\color{black} {ECINN} }%
 		\put(1,47.25) {\color{black} {EBPE} }%
 		\put(1,35.75) {\color{black} {C3LT} }%
 		\put(1,24) {\color{black} \parbox{.5in}{\textbf{Ours}\\(local)} }%
 		\put(1,12.) {\color{black} \parbox{.5in}{\textbf{Ours}\\(global)} }%
 		\put(24,93) {\color{black}  \normalsize {$\rvxr^\star$} }%
 		\put(58.5,93) {\color{black}  \normalsize {$\rvxr^\delta$} }%
 		\put(31.85,7.5){\bf	\linethickness{0.4mm} \color{cyan}\vector(1,0){57.6}}
 		\put(31.85,19.25){\bf	\linethickness{0.4mm} \color{green}\vector(1,0){57.6}}
        \put(31.85,90){\bf	\linethickness{0.4mm} \color{green}\vector(1,0){57.6}}
 	    \put(19.5,2) {\color{black}  { $\pdf_s=$} }%
 	    \put(31.,2) {\color{black}  {$[\;0.05$} }%
 		\put(45,2) {\color{black}  {$0.25$} }%
 		\put(57.5,2) {\color{black}  {$0.5$} }%
 	    \put(68,2) {\color{black}  {$0.75$} }%
 	    \put(79.5,2) {\color{black}  {$0.95\;]$} }%
 	\end{overpic}
 	\caption{We generate CelebA CFs ($\rvxr^\delta$) linearly for the input, with increasing confidence for smiling ($\rvyr=s$) from left to right. On the leftmost side, $\rvxr^\star$ denotes the reconstruction of the input  $\rvxr$. {Local denotes both local-M and local-L2 methods, as their images are indistinguishable.}}
  	\label{fig:counterExamples}
 	\end{subfigure}
  \caption{Left: Counterfactual generation. Right: Counterfactual examples.}
\end{figure}

\textit{Takeaway:} Our SEM, featuring both linear local and global explanations, yields results that stand on par with leading post-hoc explanation techniques such as C3LT and EBPE. Moreover, our model slightly outperforms ECINN across various metrics, with ECINN serving as an optimized post-hoc variant of our local CFs. The findings suggest a trade-off between consistency and realism, as no single method excels in all metrics. Notably, as realism decreases (higher FID), consistency (correlation) increases, resulting in different working points for each CF method. For further insights regarding this trade-off between consistency and realism, please refer to the Supplement.

\subsection{Qualitative Evaluation}\label{sec:Resultsqualitative}

\noindent\textbf{Prototypical Space and Bias Detection.} The prototypical space of the GdVAE is shown in \cref{fig:LatentSpace}.
This section's results reinforce GdVAE's transparency through easily comprehensible global explanations and latent space visualization. We achieve this by displaying the decoded prototypes and interpolating between them through our global explainer function (see \crefSubFigRef{fig:LatentSpace}{c}). The transparent classifier's prototypes directly uncover biases without the need for quantitative analysis of counterfactuals on simulated datasets, as shown in prior work (e.g., \cite{Singla2020Explanation}). 

\begin{wraptable}[8]{r}{0.325\textwidth}
 \vspace{-.5cm}
\caption{CelebA bias.} \label{tab:malefemale}
	\centering
\begin{tabular}{lcc}
\cline{1-3}
		{Attr.} & \footnotesize $ACC\%\uparrow$ & \footnotesize $MSE\downarrow$ \\
    	\midrule
 male & 89.9\scriptsize	$\pm$1.37 & 0.92\scriptsize$\pm$0.03   \\
    female & 91.1\scriptsize$\pm$0.33 & 0.88\scriptsize$\pm$0.02  	\\
    \cline{1-3}
 \end{tabular}
\end{wraptable}
Illustrated in \crefSubFigRef{fig:LatentSpace}{c}, the classifier's decision on smiling is shaped by female prototypes, revealing a potential bias or data imbalance not observed in our local CFs and other CF methods. The gender bias is exposed by evaluating the smile classifier across hidden attributes (\cref{tab:malefemale}), indicating reduced performance with increased uncertainty in males. Finally, we leverage the generative capabilities of the CVAE structure, generating samples for various classes. The results are presented in \crefSubFigRef{fig:LatentSpace}{b} and organized based on their distance to the corresponding prior. {Due to regularization, the latent space preserves the identity of individuals when generating samples for different classes using the same random vector.}

\noindent\textbf{CF Explanations.} Regarding visual quality, \cref{fig:counterExamples} directly compares our approach with other tested methods, using a random CelebA image. Our local method achieves visual quality comparable to state-of-the-art approaches, yielding results akin to ECINN. C3LT demonstrates smooth outcomes reminiscent of our global method, while EBPE preserves image concepts like our local method but with slight reconstruction variations. The alignment of C3LT with our global CFs, despite its expected approximation of the direction of our local CFs, highlights the advantage of our analytic link between the classifier and CFs.  

\noindent\textbf{High-resolution CFs.} We showcase our classifier's scalability in more complex scenarios, such as higher resolutions, by embedding it within a pre-trained StyleGAN architecture on the FFHQ dataset.
Local CF explanations for smiling with $\pdf_s=0.99$ are depicted in \cref{fig:FFHQCFs}. Similar to findings with CelebA data, our method retains the background while altering only pertinent attributes.
\section{Conclusion}
In this paper, we present a novel self-explainable model capable of delivering counterfactual explanations alongside transparent class predictions. Our approach uses a linear classifier in the latent space that utilizes visualizable prototypes for the downstream task. With the known linear structure, we can provide an analytical solution to generate counterfactual images. Our extensive experiments substantiate our method's ability to yield results that are on par with state-of-the-art approaches in terms of consistency, proximity, and realism while maintaining transparency. Furthermore, we illustrate how prototypes offer insight into decision logic and aid in identifying classifier bias. We see our method as a significant step toward the integration of self-explainable models and counterfactual explanation techniques. 
In contrast to previous work that requires post-hoc analysis for generating counterfactuals, our transparent model constrains the shared latent space to support consistency, proximity, and realism. Finally, resembling C3LT, our approach scales and seamlessly integrates with larger network architectures, as demonstrated on the FFHQ dataset. 
\section*{Acknowledgments} 
The authors thank e:fs TechHub GmbH, Germany, and Tongliang Liu's Trustworthy Machine Learning Lab at the University of Sydney for their support. Special appreciation goes to Muyang Li for insightful discussions and Niclas Hüwe for assistance with implementation.

%

\appendix
\clearpage
\setcounter{page}{1}

\maketitlesupplementary

\section*{Table of Contents}
The Supplement is structured as follows:
\begin{itemize}
\item Limitations and societal impacts are discussed in \cref{supp:limsoc}.
\item In \cref{supp:proof}, you can find the ELBO derivation (\cref{eq:lossgdvae,eq:lossgdvae2}), as well as proofs for the EM-based approach (\cref{algo:marg}) and the optimality of linear explainer functions (\cref{eq:localExplainer}).
\item Detailed information about the models used, the training process, and experimental specifics, such as compute resources, hyperparameters, dataset, and asset details, can be found in \cref{supp:modelandtraining}.
\item \cref{supp:metrics} delves into the metrics employed for predictive performance analysis and CF quality assessment.
\item Additional results related to hyperparameter tuning, trade-off between consistency and realism (\cref{suppfig:consistencyVsFID,fig:consistencyGammaALL}) as well as supplementary qualitative results are presented in \cref{supp:results}.
\end{itemize}

\section{Limitations and Societal Impacts}\label{supp:limsoc}
\subsection{Limitations}\label{supp:lim}

We acknowledge several limitations concerning the GdVAE and the evaluation methodology:
\begin{itemize}
    \item Our datasets for quantitative evaluation were selected to balance computational efficiency, alignment with previous studies, and adherence to the Reproducibility Checklist, emphasizing the provision of central tendency and variation (e.g., mean, standard deviation). Unlike prior CF research, our focus on analyzing both central tendency and variability has led to experiments that are four times more costly (see \cref{sec:compute}). By doing so, we acknowledge the limitations imposed by our dataset selection criteria, yet we argue that these limitations are counterbalanced by the gains in the reproducibility of our experimental results.
    \item Our method, unlike \cite{Lang_2021_ICCV} and \cite{ghandeharioun2022dissect}, enables simultaneous manipulation of all class-related attributes but lacks fine-grained control over individual image attributes. Incorporating multiple prototypes, as in \cite{Gautam22ProtoVAE}, could enhance GdVAE's predictive performance and capability to manipulate individual image attributes.
    \item Our quantitative evaluation of CF methods is currently limited to binary classification problems, and future evaluations of multi-class problems are needed to advance CF literature. Unfortunately, most visual CF methods (\cref{tab:categorize}) evaluate CFs for binary tasks like CelebA, where attributes are not mutually exclusive. Those that use multi-class settings do not scale well to a high number of classes without modification, requiring the training of a model for each binary CF class pair \cite{Khorram_2022_CVPR,Samangouei18ExplainGAN}. Other approaches require time-consuming inference-time optimization \cite{Lang_2021_ICCV,Jeanneret_2022_ACCV} or an additional image of the CF class to guide CF generation \cite{vandenhende2022making,pmlr-v97-goyal19a}. Our method can be expanded to multiple classes. For multiple classes, our CFs (\cref{eq:localExplainer}) can be generated without changes by choosing a reference class. Ambiguity between logits and softmax makes user interaction less convenient, yet setting $\delta < 0$ achieves CFs inducing class flips. Results for MNIST and CIFAR, presented in \cref{fig:suppMNISTMulti,fig:suppCIFARMulti}, demonstrate CFs for the simpler consistency task by swapping the logits of the predicted and counterfactual classes. This strategy is effective for high-confidence class predictions, such as with MNIST, though it formally only changes the class without necessarily switching to the specified CF class.
    \item Our approach utilizes a class-conditional encoder model $\varpdf_\paramA(\rvzr|\rvxr,\rvyr)$. Significantly reduced computational costs can be achieved by employing unconditional models, such as $\varpdf_\paramA(\rvzr|\rvxr)$. In \cite{falck2021multifacet}, a clustering solution is presented, directly leveraging unconditional encoder and decoder models. However, further analyses are needed to conclude the performance implications. 
    \item The EM-based method described in \cref{algo:marg} is an integral part of the error backpropagation process. Consequently, using a high number of iterations results in very deep computational graphs, which can introduce challenges, including issues like vanishing gradients and other forms of instability.
    \item All our experiments employ relatively simple network architectures, and we strive to maintain uniform training configurations as closely as possible. For instance, our baselines share the same backbone architecture, are trained with loss functions as faithful as feasible to the original implementations, and undergo training for a duration of 24 epochs. These limitations may result in methods like EBPE performing below their full potential, considering that in their original versions, both EBPE and its extension, DISSECT, were trained for 300 epochs. Results after additional training epochs are in \cref{tab:baselineModels}.
\end{itemize}

\subsection{Societal Impacts}\label{supp:soc}
Our GdVAE is not inherently associated with specific applications that directly cause negative societal impacts. However, it does possess the potential for misuse in unethical ways. For instance, given our generative models, there exists the possibility of their modified use for generating deep fake images with altered attributes. Furthermore, although our SEM provides insights into the model, it does not guarantee the detection of all fairness issues, existing biases, or results that align with attribution-based explanations.
Our methods complement existing fairness (\hspace{1sp}\cite{Louizos2015TheVF}) and explainability (\hspace{1sp}\cite{Shap2017, Rigorouspmlr-v162-lundstrom22a}) enhancement techniques rather than replacing them.

\section{Proofs}\label{supp:proof}
\subsection{Variational Lower Bound of the Joint Log-Likelihood}
In this section, we derive the Evidence Lower Bound (ELBO) for our main paper's loss function, which is represented by \cref{eq:lossgdvae,eq:lossgdvae2}. We define the key variables as follows with $\rvxr$ representing the input data (e.g., an image), $\rvyr$ corresponding to the target class, and $\rvzr$ representing our latent variables. The general ELBO for the model is derived by 
\begin{align*}
	\log \pdf_\param(\rvxr,\rvyr)&=\log \int \pdf_\param(\rvxr,\rvyr,\rvzr) d \rvzr=\log \int \pdf_\param(\rvxr,\rvyr,\rvzr) \frac{\varpdf_\paramA(\rvzr|\rvxr,\rvyr)}{\varpdf_\paramA(\rvzr|\rvxr,\rvyr)}d \rvzr\\
	&=\log \expect{ \varpdf_\paramA(\rvzr|\rvxr,\rvyr)}{\frac{\pdf_\param(\rvxr,\rvyr,\rvzr)}{\varpdf_\paramA(\rvzr|\rvxr,\rvyr)}}	\\
	&\geq 	\expect{ \varpdf_\paramA(\rvzr|\rvxr,\rvyr)}{\log \frac{\pdf_\param(\rvxr,\rvyr,\rvzr)}{\varpdf_\paramA(\rvzr|\rvxr,\rvyr)}}	(ELBO)\\
 &= 	\expect{ \varpdf_\paramA(\cdot)}{\log \pdf_\param(\rvxr,\rvyr,\rvzr)}-\expect{ \varpdf_\paramA(\cdot)}{\log \varpdf_\paramA(\rvzr|\rvxr,\rvyr)}\\
	&= 	
 \expect{ \varpdf_\paramA(\cdot)}{\log \pdf_\param(\rvxr|\rvyr,\rvzr)}+\expect{ \varpdf_\paramA(\cdot)}{\log \pdf_\param(\rvyr,\rvzr)} -\expect{ \varpdf_\paramA(\cdot)}{\log \varpdf_\paramA(\rvzr|\rvxr,\rvyr)}\\
	&=-\loss^{M1,2}(\param,\paramA;\rvxr,\rvyr),
\end{align*}
with $\varpdf_\paramA(\cdot)=\varpdf_\paramA(\rvzr|\rvxr,\rvyr)$.
The loss function for the first model M1 including a CVAE with an additional class prior $\pdf_\param(\rvyr,\rvzr)=\pdf_\param(\rvzr|\rvyr)\pdf_\param(\rvyr)$ is given by
\begin{align*}
	\loss^{M1}&=	- \expect{ \varpdf_\paramA(\cdot)}{\log \pdf_\param(\rvxr|\rvyr,\rvzr)}-\expect{ \varpdf_\paramA(\cdot)}{\log \pdf_\param(\rvyr,\rvzr)} +\expect{ \varpdf_\paramA(\cdot)}{\log \varpdf_\paramA(\rvzr|\rvxr,\rvyr)}\\
 &=	- \expect{ \varpdf_\paramA(\cdot)}{\log \pdf_\param(\rvxr|\rvyr,\rvzr)}-\expect{ \varpdf_\paramA(\cdot)}{\log \pdf_\param(\rvzr|\rvyr)}\\
 &\quad \,-\expect{\varpdf_\paramA(\cdot)}{\log \pdf_\param(\rvyr)} +\expect{ \varpdf_\paramA(\cdot)}{\log \varpdf_\paramA(\rvzr|\rvxr,\rvyr)}\\
	&=-\expect{ \varpdf_\paramA(\cdot)}{\log \pdf_\param(\rvxr|\rvyr,\rvzr)}+ \KL{\varpdf_\paramA(\rvzr|\rvxr,\rvyr)}{\pdf_\param(\rvzr|\rvyr)} -\log \pdf_\param(\rvyr).
\end{align*}
The loss for the second model M2 with a classifier and latent prior $\pdf_\param(\rvyr,\rvzr)=\pdf_\param(\rvyr|\rvzr)\pdf_\param(\rvzr)$ is given by
\begin{align*}
	\loss^{M2}&=-\expect{ \varpdf_\paramA(\cdot)}{\log \pdf_\param(\rvxr|\rvyr,\rvzr)}-\expect{ \varpdf_\paramA(\cdot)}{\log \pdf_\param(\rvyr,\rvzr)}+\expect{ \varpdf_\paramA(\cdot)}{\log \varpdf_\paramA(\rvzr|\rvxr,\rvyr)}\\
	&=	-\expect{ \varpdf_\paramA(\cdot)}{\log \pdf_\param(\rvxr|\rvyr,\rvzr)}-\expect{ \varpdf_\paramA(\cdot)}{\log \pdf_\param(\rvyr|\rvzr)}\\
  &\quad \,-\expect{\varpdf_\paramA(\cdot)}{\log \pdf_\param(\rvzr)}+\expect{ \varpdf_\paramA(\cdot)}{\log \varpdf_\paramA(\rvzr|\rvxr,\rvyr)}\\
	&=-\expect{ \varpdf_\paramA(\cdot)}{\log \pdf_\param(\rvxr|\rvyr,\rvzr)}+ \KL{\varpdf_\paramA(\rvzr|\rvxr,\rvyr)}{\pdf_\param(\rvzr)} 
-\expect{ \varpdf_\paramA(\cdot)}{\log \pdf_\param(\rvyr|\rvzr)}.
\end{align*}
The loss function $\tilde{\loss}^{gd}=\tilde{\loss}^{gd}(\param,\paramA;\rvxr,\rvyr)$ for a joint training of the conditional variational autoencoder and classifier can be obtained by combining $\loss^{M1}=\loss^{M1}(\param,\paramA;\rvxr,\rvyr)$ and $\loss^{M2}=\loss^{M2}(\param,\paramA;\rvxr,\rvyr)$. We then obtain 
\begin{align}\label{eq:gdvaeLossOriginal}
\begin{split}
	\tilde{\loss}^{gd}=&\alpha \loss^{M1}+\beta \loss^{M2}\\ 
	=&	-(\alpha+\beta)\expect{ \varpdf_\paramA(\cdot)}{\log \pdf_\param(\rvxr|\rvyr,\rvzr)}+ \alpha\left(\KL{\varpdf_\paramA(\rvzr|\rvxr,\rvyr)}{\pdf_\param(\rvzr|\rvyr)} -\log \pdf_\param(\rvyr)\right)  \\ &+\beta\left(\KL{\varpdf_\paramA(\rvzr|\rvxr,\rvyr)}{\pdf_\param(\rvzr)} 
   -\expect{ \varpdf_\paramA(\cdot)}{\log \pdf_\param(\rvyr|\rvzr)}\right),
   \end{split}
\end{align}
with $\varpdf_\paramA(\cdot)=\varpdf_\paramA(\rvzr|\rvxr,\rvyr)$.
\subsection{Variational Expectation Maximization for the Marginalization Process (\cref{algo:marg})}

In this section, we employ variational expectation maximization (EM) to provide a proof for the iterative algorithm, as depicted in \cref{algo:marg}. This algorithm serves as a fundamental component for marginalization. It's worth noting that this EM process is nested within the overarching variational optimization of the GdVAE. Consequently, we switch the roles of distributions, with $\pdf(\cdot)$ representing the variational distribution and $\varpdf(\cdot)$ signifying the model distribution.

Consider the variable pair $\left(\rvxr, \rvzr, \rvyr \right)$ within the model distribution $\varpdf_\paramA(\rvxr, \rvzr, \rvyr)$, where only $\rvxr$ is observable. The objective of variational EM is to optimize the model parameters $\paramA$ by maximizing the marginal likelihood $\varpdf_\paramA(\rvxr)$. This optimization is achieved by leveraging a lower bound on the marginal likelihood, using the 'variational' distribution $\pdf_\param(\rvzr, \rvyr|\rvxr)$. This lower bound is defined by
\begin{align*}
	\log \varpdf_\paramA(\rvxr) &\geq  \expect{\pdf_\param(\rvzr, \rvyr|\rvxr)}{\log \varpdf_\paramA(\rvxr, \rvzr, \rvyr)} \quad \,-  \expect{\pdf_\param(\rvzr, \rvyr|\rvxr)}{\log \pdf_\param(\rvzr, \rvyr|\rvxr)}\\
	&=-\expect{\pdf_\param(\rvzr, \rvyr|\rvxr)}{\log \frac{\pdf_\param(\rvzr, \rvyr|\rvxr)}{\varpdf_\paramA(\rvzr, \rvyr|\rvxr)}}+ \log \varpdf_\paramA(\rvxr).
 \end{align*}
By rearranging the lower bound, we obtain the function $\lossC(\pdf, \paramA)$, which serves as the objective for the EM algorithm. This function is meant to be maximized and is defined by
\begin{align*}
	\Rightarrow 0\geq  \lossC(\pdf, \paramA) &= -\KL{\pdf_\param(\rvzr, \rvyr|\rvxr)}{\varpdf_\paramA(\rvzr, \rvyr|\rvxr)}\\ 
 &= -\expect{\pdf_\param(\rvzr, \rvyr|\rvxr)}{\log \frac{\pdf_\param(\rvzr, \rvyr|\rvxr)}{\varpdf_\paramA(\rvzr, \rvyr|\rvxr)}}= -\expect{\pdf_\param(\rvzr, \rvyr|\rvxr)}{\log \frac{\pdf(\rvyr| \rvzr)\pdf_\param(\rvzr|\rvxr)}{\varpdf_\paramA(\rvyr| \rvxr)\varpdf_\paramA(\rvzr|\rvxr)}}.
\end{align*}
We further assume conditional independence and apply the following factorizations $\varpdf_\paramA(\rvzr, \rvyr|\rvxr)=\varpdf_\paramA(\rvyr| \rvzr,\rvxr)\varpdf_\paramA(\rvzr|\rvxr)=\varpdf_\paramA(\rvyr| \rvxr)\varpdf_\paramA(\rvzr|\rvxr)$, with  \newline{$\varpdf_\paramA(\rvzr|\rvxr)=\sum_{\rvyr=1}^\numclasses \varpdf(\rvzr|\rvxr,\rvyr)\varpdf_\paramA(\rvyr|\rvxr)$}. 
Likewise, we define $\pdf_\param(\rvzr, \rvyr|\rvxr)=\pdf(\rvyr| \rvzr)\pdf_\param(\rvzr|\rvxr)$. The parameters $\param$ and $\paramA$ were omitted for the distributions assumed to remain constant throughout the EM procedure. These distributions include $\pdf(\rvyr|\rvzr)$, representing the classifier employing the prior encoder, and $\varpdf(\rvzr|\rvxr,\rvyr)$, representing the recognition model of the GdVAE. The iterative EM procedure for step $\iter\in\{1,\ldots,\numiter\}$ can be expressed as follows:

\noindent\textbf{E-Step:}$\,$Choose a distribution $\pdf\!=\!\pdf_\param(\rvzr, \rvyr|\rvxr)$ that maximizes $\lossC(\pdf, \paramA)$ for fixed $\paramA^\iter$.
\begin{itemize}
	\item Since $\pdf(\rvyr|\rvzr)$ is fixed the optimum is given by choosing $\pdf_\param(\rvzr|\rvxr)=\varpdf_{\paramA^\iter}(\rvzr|\rvxr)$. 
\end{itemize}

\noindent\textbf{M-Step:}$\,$Choose parameters $\paramA^{\iter+1}$ that maximize $\lossC(\pdf, \paramA)$ for fixed $\pdf\!=\!\pdf_\param(\rvzr,$\,$\rvyr|\rvxr)$.
\begin{itemize}
	\item The optimum is given by choosing $\varpdf_{\paramA^{\iter+1}}(\rvyr| \rvxr)= \expect{\varpdf_{\paramA^\iter}(\rvzr|\rvxr)}{\pdf(\rvyr| \rvzr)}$. This choice defines $\varpdf_{\paramA^{\iter+1}}(\rvzr|\rvxr)=\sum_{\rvyr=1}^\numclasses \varpdf(\rvzr|\rvxr,\rvyr)\varpdf_{\paramA^{\iter+1}}(\rvyr|\rvxr)$ as well. 
\end{itemize}
Proof for the M-Step: We can simplify the optimization process by assuming that both $\varpdf_\paramA(\rvzr|\rvxr)$ and $\pdf_\param(\rvzr|\rvxr)$ are Gaussian mixture models with the same number, denoted as $\numclasses$, of mixture components. Drawing inspiration from \cite{1186768}, we can next derive a lower bound, denoted as $\loss(\pdf, \paramA)$, for $\lossC(\pdf, \paramA)$ through the application of the \textit{log-sum} inequality and by inserting $\pdf_\param(\rvzr|\rvxr)=\varpdf_{\paramA^\iter}(\rvzr|\rvxr)$ into $\lossC(\pdf, \paramA)$
\begin{align*}
	\lossC(\pdf, \paramA)&=-\int \sum_{\rvyr=1}^\numclasses \pdf(\rvyr| \rvzr)\varpdf_{\paramA^\iter}(\rvzr|\rvxr)\cdot \log \frac{ \pdf(\rvyr| \rvzr)\sum_{\rvyr^\star=1}^\numclasses  \varpdf(\rvzr|\rvxr,\rvyr^\star )\varpdf_{\paramA^\iter}(\rvyr^\star |\rvxr)}{\varpdf_\paramA(\rvyr|\rvxr)\sum_{\rvyr^\star=1}^\numclasses  \varpdf(\rvzr|\rvxr,\rvyr^\star)\varpdf_\paramA(\rvyr^\star|\rvxr)} d\rvzr \\ 
	&\geq \loss(\pdf, \paramA) = -\int \sum_{\rvyr=1}^\numclasses \pdf(\rvyr| \rvzr) \varpdf_{\paramA^\iter}(\rvzr|\rvxr) \cdot\log \frac{ \pdf(\rvyr| \rvzr)\varpdf(\rvzr|\rvxr,\rvyr)\varpdf_{\paramA^\iter}(\rvyr|\rvxr)}{\varpdf_\paramA(\rvyr|\rvxr) \varpdf(\rvzr|\rvxr,\rvyr)\varpdf_\paramA(\rvyr|\rvxr)} d\rvzr\\
	&=-\expect{\varpdf_{\paramA^\iter}(\rvzr|\rvxr)}{\sum_{\rvyr=1}^\numclasses \pdf(\rvyr| \rvzr) \log \frac{ \pdf(\rvyr| \rvzr)\varpdf_{\paramA^\iter}(\rvyr|\rvxr)}{\varpdf_\paramA(\rvyr|\rvxr) \varpdf_\paramA(\rvyr|\rvxr)} }.
\end{align*} 
To maximize $\loss(\pdf, \paramA)$, we employ the Lagrange multiplier $\lag\in \mathbb{R}$ and set the derivative with respect to a specific class $\class$ to zero
\begin{align*}
0&=\frac{\partial}{\partial \varpdf_\paramA(\class|\rvxr)} \left[\loss(\pdf, \paramA) - \lag(\sum_{\rvyr=1}^\numclasses \varpdf_\paramA(\rvyr|\rvxr)-1) \right]=\expect{\varpdf_{\paramA^\iter}(\rvzr|\rvxr)}{2 \frac{\pdf(\class| \rvzr)}{\varpdf_\paramA(\class|\rvxr)}} - \lag,\\
	&\Rightarrow \varpdf_{\paramA^{\iter+1}}(\rvyr|\rvxr)= \expect{\varpdf_{\paramA^\iter}(\rvzr|\rvxr)}{\pdf(\rvyr| \rvzr)}.
\end{align*}
We determine the value of $\lambda$ by solving this equation for $\numclasses$ classes, incorporating the property ${\sum_{\rvyr=1}^\numclasses \varpdf_\paramA(\rvyr|\rvxr)=1}$. Subsequently, we arrive at the solution $\varpdf_{\paramA^{\iter+1}}(\rvyr|\rvxr) = \expect{\varpdf_{\paramA^\iter}(\rvzr|\rvxr)}{\pdf(\rvyr| \rvzr)}$ or, equivalently, the parameter $\paramA^{\iter+1}$ that maximizes $\lossC(\pdf, \paramA)$ while keeping $\pdf_\param(\rvzr, \rvyr|\rvxr)$ fixed. This optimization is performed for a single input $\rvxr$, and we approximate the expectation through Monte Carlo integration. Therefore, we sample $\rvzr^{(\samples)}$ from $\varpdf_{\paramA^\iter}(\rvzr|\rvxr)$ and calculate  $\varpdf_{\paramA^{\iter+1}}(\rvyr|\rvxr)= \expect{\varpdf_{\paramA^\iter}(\rvzr|\rvxr)}{\pdf(\rvyr| \rvzr)} \approx \frac{1}{\numsamples} \sum_{\samples=1}^\numsamples \pdf_\param(\rvyr|\rvzr^{(\samples)})$. 

\subsection{Optimality of Linear Explainer Functions (\cref{eq:optim} and \cref{eq:localExplainer})}

\noindent\textbf{Euclidean Space ($\metrictensor:=I$).} Our SEM is regularized to produce a linear separating hyperplane due to the linear classifier we employ. This results in a linear path for CF generation, where the classifier's gradient vector $w$ describes the \textit{shortest path} for CF generation (see \cref{fig:counter}). This \textit{latent space closeness} is also used as an L2-based metric $\mdist^2_I(z^{(1)},z^{(2)})$ to measure CF proximity \cite{Singla2020Explanation} or to directly optimize a perceptual loss for CF generation \cite{Jeanneret_2022_ACCV}. 

\noindent\textbf{Riemannian Manifolds ($\metrictensor:=\Sigma^{-1}_\rvzr$).} 
A generalized perspective on the distance function arises from considering the theoretical analysis on Riemannian manifolds presented in \cite{NEURIPS2022GeometricVAE, NIPS2012_MetricLearning}. According to \cite{NIPS2012_MetricLearning}, a Riemannian manifold, denoted as the pair $(\manifold, \innerprod_\rvzr)$, can be understood as a smoothly curved space $\manifold$ (\eg, latent space) equipped with a Riemannian metric $\innerprod_\rvzr$. The Riemannian metric is an inner product $\innerprod_\rvzr(a,b)=\langle a,b\rangle_\rvzr=a^T\metrictensor(\rvzr) b$ on the tangent space $\tangspace{\rvzr}$ for each $\rvzr\in \manifold$. The metric tensor $\metrictensor(\rvzr)$ is a positive definite matrix that induces a distance measure. Assuming a constant metric tensor or single metric learning, the Mahalanobis distance is obtained \cite{NIPS2012_MetricLearning,NEURIPS2022GeometricVAE}
\begin{align}\label{eq:distmeasure}
\mdist^2_{\metrictensor}\left(\rvzr^{(1)},\rvzr^{(2)}\right)&=\norm{\rvzr^{(1)}-\rvzr^{(2)}}^2_{\metrictensor}=\left(\rvzr^{(1)}-\rvzr^{(2)}\right)^T\metrictensor \left(\rvzr^{(1)}-\rvzr^{(2)}\right). 
\end{align}
If the metric tensor is chosen to be the identity matrix $\metrictensor:=I$, we obtain the L2-based Euclidean distance. To define a smooth continuous Riemannian metric with a metric tensor in every point $\rvzr$, \cite{NIPS2012_MetricLearning} propose
\begin{align}
    \label{eq:metrictensor}
    \metrictensor(\rvzr)&= \sum_{k=1}^K \pi_k(\rvzr) \metrictensor_k,
\end{align}
where $\metrictensor_k$ are pre-trained metric tensors, \eg, obtained by a Large Margin Nearest Neighbor classifier, which are associated with the mean of each class. $\paramC_k$ are weights that change smoothly with $\rvzr$, where each $\paramC_k>0$ and $\sum_{k=1}^K\paramC_k=1$. As an example of a smooth weight function, they use the following
\begin{align}
\pi_k(\rvzr)&\propto\exp{\left(-\frac{\rho}{2} \norm{\rvzr-\rvzr^{(k)}}^2_{\metrictensor_k}\right)},
\end{align}
with the constant $\rho$ and class means $\rvzr^{(k)}$. Based on this continuous Riemannian metric (see \cref{eq:metrictensor}), \cite{NEURIPS2022GeometricVAE} propose using the trained covariance matrices from a VAE’s encoder $\varpdf_\paramA(\rvzr|\rvxr)=\gaussian\left(\mu_\rvzr(\rvxr;\paramA), \Sigma_\rvzr(\rvxr;\paramA)\right)$ to define the metric. Specifically, they employ a weighted linear combination of $\metrictensor_k=\Sigma^{-1}_\rvzr(\rvxr^{(k)};\paramA)$, where the training data, or a subset thereof, is used to approximate the metric in the latent space. In addition to \cref{eq:metrictensor}, there is a supplementary additive component which is approximately zero. They further observe that, due to the Evidence Lower Bound (ELBO) objective, variables that are close in the latent space with respect to $\metrictensor(\rvzr)$ will also produce samples that are close in the image space in terms of L2 distance, which is crucial for ensuring counterfactual proximity.

\noindent\textbf{Optimization and Assumptions.}
In our analysis, we assume that the regularizer included in the ELBO induces a surrogate posterior  $\varpdf_\paramA(\rvzr|\rvxr,\rvyr)$ that closely approximates the true posterior $\pdf_\param(\rvzr|\rvyr)= \gaussian\left(\mu_\rvzr(\rvyr;\param), \Sigma_\rvzr(\rvyr;\param)\right)$. Given this approximation, we may consider using pre-trained metric tensors from the GDA (Gaussian discriminant analysis) classifier instead of those derived from a Large Margin Nearest Neighbor classifier \cite{NIPS2012_MetricLearning}. By doing so, as referenced in \cref{eq:metrictensor}, we obtain a Riemannian metric
\begin{align}
    \label{eq:metrictensorCVAE}
    \metrictensor(\rvzr)&= \sum_{k=1}^K \pi_k(\rvzr) \Sigma^{-1}_\rvzr(\rvyr=k;\param),
\end{align}
with $\pi_k(\rvzr)=\pdf_\param(\rvyr=k|\rvzr)$. In our experiments, we use only two classes and have chosen the covariance to be independent of the class $\rvyr$ in order to obtain linear discriminants. Consequently, this results in a constant metric tensor, effectively employing a single metric
\begin{align}
    \label{eq:metrictensorCVAEtwoclass}
    \metrictensor(\rvzr)&= \pi_1(\rvzr) \Sigma^{-1}_\rvzr+(1-\pi_1(\rvzr))\Sigma^{-1}_\rvzr=\Sigma^{-1}_\rvzr=\metrictensor.
\end{align}
Having defined two types of distance metrics, the L2 and Mahalanobis distances, we can now optimize the objective
\begin{align}
 \counterfactual_\discriminantfkt(\rvzr, \delta)=\argmin_{z^\delta} \; & \mdist^2_\metrictensor(z^\delta,z), \quad
 \textrm{subject to} \;  f(z^\delta)=\delta.
\end{align}
To minimize the objective $\mdist^2_\metrictensor(z^\delta,z)$, we use a Lagrange multiplier $\lag\in \mathbb{R}$ and set the derivatives with respect to the counterfactual $z^\delta$ and $\lag$ to zero
\begin{align}
    \loss(z^\delta)&=\left(\rvzr^{\delta}-\rvzr\right)^T\metrictensor \left(\rvzr^{\delta}-\rvzr\right) + \lag(f(z^\delta)-\delta),\\
    \frac{\partial \loss(z^\delta)}{\partial z^\delta}&=2\left(\rvzr^{\delta}-\rvzr\right)\metrictensor + \lag \weight=0, \quad
    \Rightarrow z^\delta =z + \frac{\lambda}{2}\metrictensor^{-1}\weight,\label{eq:optiz}\\
    \frac{\partial \loss(z^\delta)}{\partial \lambda}&=0, \quad \Rightarrow f(z^\delta)=\weight^T z^\delta + \bias=\delta.\label{eq:optil}
\end{align}
By combining \cref{eq:optiz,eq:optil}, we arrive at the solution
\begin{align}
    \weight^T z^\delta &=\weight^Tz + \frac{\lambda}{2}\weight^T\metrictensor^{-1}\weight=\delta-\bias,\\
    \Rightarrow z^\delta&=z+\kappa \metrictensor^{-1}\weight, \; \text{with}\; \kappa=\frac{\delta-\weight^T\rvzr-\bias}{\weight^T \metrictensor^{-1}\weight}.
\end{align}
Given the assumptions used, we obtain a linear explainer function to generate counterfactuals, regardless of whether we choose the common L2-based metric  $\metrictensor=I$ (Euclidean space) or the Riemannian metric $\metrictensor=\Sigma^{-1}_\rvzr$.
Training with $\Sigma_z\!=\!\sigma^2I$ instead of $\covMatP_{\rvzr}\!=\!\diag\left(\sigma^2_{\conceptlog_1},., \sigma^2_{\conceptlog_\dimdata} \right)$ results in equivalent explainer functions, thus yielding equal empirical results for both L2-based and Riemannian-based metrics. Consequently, a linear function is optimal, and thus there is no superior solution for generating counterfactuals under these conditions. If the assumption that $\varpdf_\paramA(\rvzr|\rvxr,\rvyr)\approx \pdf_\param(\rvzr|\rvyr)$ is not met, a non-linear CF method (\eg, C3LT) may better fulfill the proximity property in the image space.

\newpage 
\section{Training and Model Details}\label{supp:modelandtraining}

The software to train and perform inference with our GdVAE model is available in the supplemental code.

\subsection{Datasets}\label{supp:datasetAndCode}

\noindent\textbf{MNIST \cite{mnist}.} The MNIST dataset comprises two sets: a training set with 60,000 labeled examples and a test set with 10,000 labeled examples. Each example is a 28x28 pixel grayscale image representing a handwritten digit ranging from 0 to 9. For the predictive performance analysis (\cref{tab:gdvaeEMvsIS}) we use all classes and for the evaluation of counterfactuals (\cref{tab:consistency}) we exclusively utilize the digits 0 and 1 for the training and to generate counterfactuals (denoted by  "MNIST-Binary 0/1"). We adhere to the standard data splits for both training and testing.

\noindent\textbf{CIFAR-10 \cite{cifar10}.} 
The CIFAR-10 dataset comprises over 60,000 images, along with annotations for ten classes. Each example is a 32x32 pixel image with 3 color channels. We adhere to the standard data splits for both training and testing. This dataset is used to analyze the predictive performance in \cref{tab:gdvaeEMvsIS}.

\noindent\textbf{CelebA \cite{celeba}.} 
The CelebA dataset comprises over 200,000 face images, along with annotations that cover a range of attributes, including gender, age, and facial expression. In our analysis, we employ a center crop of 128x128 pixels for the images and resize them to 64x64 pixel images with 3 color channels. For counterfactual evaluation (\cref{tab:consistency}), we use the attributes of "smiling" (labeled as 1) and "not smiling" (labeled as 0), while the "gender" attribute is used to assess predictive performance (\cref{tab:gdvaeEMvsIS}). We adhere to the standard data splits for both training and testing. For licensing details and information on human subject data collection, please see the reference \cite{celeba}.

\noindent\textbf{FFHQ \cite{StyleGan8953766}.}
The Flickr-Faces-HQ (FFHQ) dataset comprises over 70,000 high-resolution (1024x1024 pixels) images of human faces, curated for diversity in age, ethnicity, and accessories. The dataset was developed by NVIDIA. For counterfactual generation (\cref{fig:FFHQCFs}), we use the attributes of "smiling" (labeled as 1) and "not smiling" (labeled as 0) available at  \href{https://github.com/DCGM/ffhq-features-dataset/}{https://github.com/DCGM/ffhq-features-dataset/}. For licensing details and information on human subject data collection, please see the reference \cite{StyleGan8953766}.

\begin{figure*}[!ht]
	\centering
		\begin{overpic}[grid=false, width=0.95\textwidth]{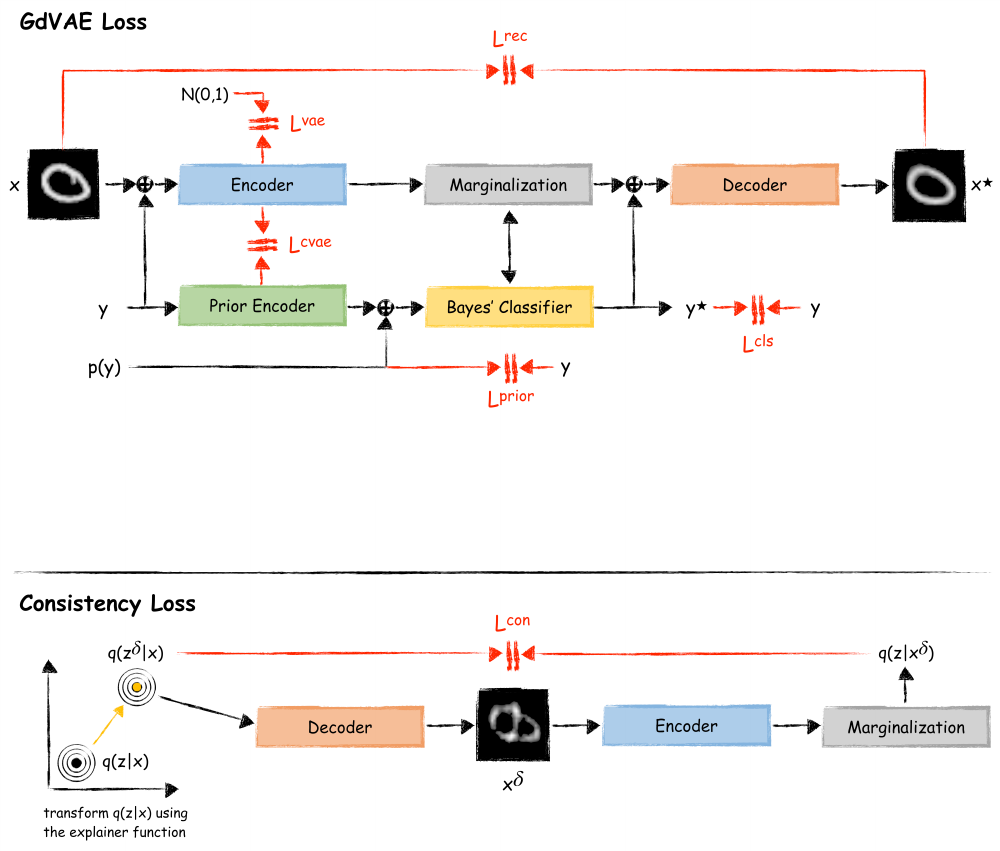}%
 		 \put(2.0,40.5) {$\loss^{gd}\quad =	(\alpha+\beta)\loss^{rec}+\alpha\left(\loss^{cvae} +\loss^{prior}\right) +\beta\left(\loss^{vae}+\loss^{cls}\right) \text{, with}$}%
     
     \put(1.0,35.5) { $\loss^{rec}\;\;\:=-\expect{\varpdf_\paramA(\rvzr,\rvyr|\rvxr)}{\log \pdf_\param(\rvxr|\rvyr,\rvzr)},\quad \loss^{cvae}=\KL{\varpdf_\paramA(\rvzr|\rvxr,\rvyr)}{\pdf_\param(\rvzr|\rvyr)},$}
     \put(1.2,30.5) { $\loss^{prior}=-\log \pdf_\param(\rvyr), \quad \loss^{vae}=\KL{\varpdf_\paramA(\rvzr|\rvxr,\rvyr)}{\pdf(\rvzr)},\quad \loss^{cls}=-\expect{\varpdf_\paramA(\rvzr|\rvxr)}{\log \pdf_\param(\rvyr|\rvzr)}.$}
     
 		\put(30.1,1.8) { \color{black}  { $\loss^{con}=\expect{\pdf(\delta)}{\KL{\varpdf_\paramA(\rvzr|\rvxr^\delta)}{\varpdf_\paramA(\rvzr^\delta|\rvxr)}}$}}%
 		\end{overpic}
\caption{{Diagram illustrating the GdVAE model components and their interactions, highlighting the key elements that contribute to loss function computation.}}
	\label{fig:GdVAELosses}
\end{figure*}
\subsection{GdVAE Details}\label{supp:GdVAE}
\textbf{GdVAE Training.} The training algorithm is outlined in the supplemental code. The fundamental GdVAE training procedure, does not include the consistency loss. This particular training method is employed in \cref{tab:gdvaeEMvsIS} and in \cref{tab:consistency} the consistency regularizer is incorporated. During the training process, we employ both the global and local-L2 explainer functions to generate counterfactual (CF) examples. To generate samples, we follow a random selection procedure, sampling from either explainer function with equal probability. \cref{fig:GdVAELosses} illustrates the inputs to the loss functions.

The hyperparameters used for training and architectural decisions to replicate the results from the paper are summarized in \cref{tab:hyper}. The architectures are presented in \cref{tab:archiCeleb,tab:archiCifar,tab:archiMNIST,tab:archiFFHQ,tab:archiPriorEncoder}. We maintain consistent hyperparameters  ($\nicefrac{\alpha}{\beta}=1$, $\gamma=1$, $\numiter=3$, $\numsamples=20$) across all datasets. Baseline models adopt dataset-specific settings for improved results, as indicated in \cref{tab:baselineModels}.

\noindent\textbf{GdVAE Architectures.} The architecture of the neural network for the CelebA GdVAE is inspired by the encoder and decoder configurations presented in \cite{Ding_2020_CVPR}. When dealing with CelebA, we use $4\times4$ kernels for every convolutional layer, with a stride of 2 and padding set to 1. However, the last convolutional layer deviates from this pattern and employs the default stride of 1 with no padding. In between these layers, Rectified Linear Units (ReLU) are applied.

For more detailed model specifications related to CelebA, please consult \cref{tab:archiCeleb}. The notation "FC ($\times 2$)" signifies that two distinct fully connected networks are in use for both the mean and log variance calculations. In our architecture, "Conv2d" stands for 2D convolution, while "ConvT2d" corresponds to 2D transposed convolution. The stride and padding settings for these operations are represented as "s" and "p", respectively.
These operations are implemented using the torch.nn package in PyTorch. 

It's worth noting that the prior encoder and label embeddings remain consistent across all models and datasets, as displayed in \cref{tab:archiPriorEncoder}. Architectural details for other datasets are provided in \cref{tab:archiCifar,tab:archiMNIST,tab:archiFFHQ}. {For FFHQ we use the pre-trained StyleGAN architecture based on \cite{yu2022diverse}. 
Our GdVAE model, as shown in Table \ref{tab:archiFFHQ}, is integrated between StyleGAN's encoder and decoder.} For a comprehensive understanding of the hyperparameters used in the experiments, please refer to \cref{tab:hyper} in the case of experiments detailed in \cref{tab:gdvaeEMvsIS,tab:consistency}. An ablation study regarding the hyperparameters can be found in \cref{supp:GdVAEAnalysis}.
 
\begin{table}
    \caption{Hyperparameters and architectural decisions for our GdVAE model configurations on MNIST, CIFAR-10, and CelebA. Values specific to binary classification scenarios (MNIST - binary 0/1, CelebA) that include the use of consistency loss are enclosed in parentheses, if they differ from the parameters used in other experiments. The ADAM optimizer is consistently employed across all datasets.}
    \label{tab:hyper}
    \centering
    \small 
    \begin{tabular}{lccc}
    \toprule
        & {MNIST} & {CIFAR-10}& {CelebA}  \\ 
    \midrule
    Batch size    & 64 & 64 & 64  \\
    Learning rate    & 0.0005 & 0.0005 & 0.0005  \\
    Epochs    & 24 & 24 & 24  \\
    Number classes $\numclasses$ &  10(2) & 10 & 2 \\
    Latent dimension $\dimdata$    & 10 & 64 & 64 \\
    Latent dimension $\dimlatentprior$   & 10(4) & 10 & 4 \\
    Samples $\numsamples$ in \cref{algo:marg}  & 20 & 20  & 20  \\
    Samples for $\mathbb{E}$ in \cref{eq:consitency}  & (10) & -  & (10)  \\
    $\epsilon$ for $\pdf(\delta)=\mathcal{U}(-\epsilon,\epsilon)$ & (2.94) & - & (2.94)  \\
    Samples $O$ for $\mathbb{E}$ in \cref{eq:lossgdvae,eq:lossgdvae2}  & 1 & 1  & 1  \\
    Iterations $\numiter$ in \cref{algo:marg}   & 3 & 3  & 3  \\
         \bottomrule
    \end{tabular}
\vspace{-0.5cm}
\end{table}
\begin{table}
    \caption{Prior encoder and label embeddings. In binary classification tasks, we opt for a class-independent choice of $\Sigma_\rvzr(\rvyr;\param) = \Sigma_\rvzr(\param)$. Consequently, we employ a distinct encoder network for the covariance $\Sigma_\rvzr(\param)$, which is obtained by utilizing a constant input, $\rvyr=1$.}
    \label{tab:archiPriorEncoder}
    \centering
    \vspace{-0.2cm}
    \begin{tabular}{l}
    \toprule
          \textbf{Prior Encoder $\pdf_\param(\rvzr|\rvyr) = \gaussian\left(\mu_\rvzr(\rvyr;\param), \Sigma_\rvzr(\rvyr;\param)\right)$} \\
        \midrule
           Input: $\numclasses$ values\\
           FC: $\dimlatentprior$ values $\Rightarrow$ FC: $\dimlatentprior$ values  $\Rightarrow$ FC: $\dimlatentprior$ values\\
           FC ($\times 2$): $\dim(z)=\dimdata$ values\\
          \toprule
        {\textbf{Encoder Label Embedding} for $\rvyr$ in $\varpdf_\paramA(\rvzr|\rvxr,\rvyr)$} \\
        \midrule
           Input: $\numclasses$ values $\Rightarrow$
           FC: $1$ channel, $W\times H$ image \\
    \toprule
          {\textbf{Decoder Label Embedding} for $\rvyr$ in $\pdf_\param(\rvxr|\rvyr,\rvzr)$} \\
        \midrule
           Input: $\numclasses$ values $\Rightarrow$ FC: $1$ value\\
        \bottomrule
    \end{tabular}

\end{table}

\begin{table*}[!htbp]
  \caption{Architecture for CelebA. We center crop and resize the images to have a width and height of $(W\times H) = (64\times64)$ and keep the three channels $C=3$. The class label $\rvyr$ is incorporated into the input image as a fourth channel through a label embedding (\cref{tab:archiPriorEncoder}). The latent space's dimensionality is defined as $\dim(z)=\dimdata=64$. We utilize $\varpdf_\paramA(\rvzr|\rvxr,\rvyr)=\gaussian\left(\mu_\rvzr(\rvxr,\rvyr;\paramA), \Sigma_\rvzr(\rvxr, \rvyr;\paramA)\right)$, with diagonal covariance matrix $\Sigma_\rvzr(\rvxr,\rvyr;\paramA)=\diag\left(\sigma^2_{\conceptlog_1| \rvxr,\target},\ldots, \sigma^2_{\conceptlog_\dimdata| \rvxr, \target} \right)$.}
    \label{tab:archiCeleb}
    \vspace{-0.2cm}
     \centering
    \resizebox{\textwidth}{!}{\begin{tabular}{ll}
    \toprule
        \textbf{Encoder $\varpdf_\paramA(\rvzr|\rvxr,\rvyr)$} &   \textbf{Decoder $\pdf_\param(\rvxr|\rvyr,\rvzr)$} \\
         \midrule
         Input: $4\times64\times64$ & Input: $\dim(\rvzr)+1=\dimdata+1$ values\\
         Conv2d: $32$ channels, $4\times4$ kernel, s=$2$, p=$1$ & FC: $256$ channels, $1\times1$ image \\
         Conv2d: $32$ channels, $4\times4$ kernel, s=$2$, p=$1$ &ConvT2d: $64$ channels, $4\times4$ kernel\\
         Conv2d: $64$ channels, $4\times4$ kernel, s=$2$, p=$1$ &ConvT2d: $64$ channels, $4\times4$ kernel, s=$2$, p=$1$\\
         Conv2d: $64$ channels, $4\times4$ kernel, s=$2$, p=$1$ &ConvT2d: $32$ channels, $4\times4$ kernel, s=$2$, p=$1$\\
         Conv2d: $256$ channels, $4\times4$ kernel & ConvT2d: $32$ channels, $4\times4$ kernel, s=$2$, p=$1$\\
         FC: $2\cdot \dim(\rvzr)=2\cdot\dimdata$ values& ConvT2d: $3$ channels, $4\times4$ kernel, s=$2$, p=$1$\\
Output:  $\dimdata$ values for $\mu_\rvzr(\rvxr,\rvyr;\paramA)$ and $\log \Sigma_\rvzr(\rvxr,\rvyr;\paramA)$& Output: $3\times64\times64$ image for $\mu_\rvxr(\rvyr,\rvzr;\param)$\\ 
        \bottomrule
    \end{tabular}}
\end{table*}
\begin{table*}[!htbp]
    \caption{Architecture for CIFAR-10.}
    \label{tab:archiCifar}
     \vspace{-0.3cm}
    \centering
      \resizebox{\textwidth}{!}{\begin{tabular}{ll}
    \toprule
        \textbf{Encoder $\varpdf_\paramA(\rvzr|\rvxr,\rvyr)$}&  \textbf{Decoder $\pdf_\param(\rvxr|\rvyr,\rvzr)$} \\
         \midrule
         Input: $4\times32\times32$  & Input: $\dim(\rvzr)+1=\dimdata+1$ values\\
         Conv2d: $64$ channels, $4\times4$ kernel, s=$2$, p=$1$ & FC: $256$ channels, $1\times1$ image \\
         Conv2d: $128$ channels, $4\times4$ kernel, s=$2$, p=$1$ & ConvT2d: $256$ channels, $2\times2$ kernel\\
         Conv2d: $256$ channels, $4\times4$ kernel, s=$2$, p=$1$ & ConvT2d: $256$ channels, $2\times2$ kernel, s=$2$\\
         Conv2d: $256$ channels, $2\times2$ kernel, s=$2$ & ConvT2d: $128$ channels, $4\times4$ kernel, s=$2$, p=$1$\\
         Conv2d: $256$ channels, $2\times2$ kernel & ConvT2d: $64$ channels, $4\times4$ kernel, s=$2$, p=$1$\\
        FC ($\times 2$): $\dim(z)=\dimdata$ values & ConvT2d: $3$ channels, $4\times4$ kernel, s=$2$, p=$1$\\
      Output:  $\dimdata$ values for $\mu_\rvzr(\rvxr,\rvyr;\paramA)$ and $\log \Sigma_\rvzr(\rvxr,\rvyr;\paramA)$& Output: $3\times32\times32$ image for $\mu_\rvxr(\rvyr,\rvzr;\param)$\\ 
        \bottomrule
    \end{tabular}}
\end{table*}
\begin{table*}[!htbp]
    \caption{Architecture for MNIST. }
    \label{tab:archiMNIST}
     \vspace{-0.3cm}
    \centering
    \resizebox{\textwidth}{!}{\begin{tabular}{ll}
    \toprule
        \textbf{Encoder $\varpdf_\paramA(\rvzr|\rvxr,\rvyr)$}  & \textbf{Decoder $\pdf_\param(\rvxr|\rvyr,\rvzr)$} \\
         \midrule
         Input: $2\times28\times28$  & Input: $\dim(\rvzr)+1=\dimdata+1$ values\\
         Conv2d: $64$ channels, $6\times6$ kernel, s=$2$  &  FC: $256$ channels, $4\times4$ image \\
         Conv2d: $128$ channels, $5\times5$ kernel & ConvT2d: $128$ channels, $5\times5$ kernel\\
         Conv2d: $256$ channels, $5\times5$ kernel& ConvT2d: $64$ channels, $5\times5$ kernel\\
        FC ($\times 2$): $\dim(z)=\dimdata$ values& ConvT2d: $1$ channel, $6\times6$ kernel, s=$2$\\
         Output:  $\dimdata$ values for $\mu_\rvzr(\rvxr,\rvyr;\paramA)$ and $\log \Sigma_\rvzr(\rvxr,\rvyr;\paramA)$ & Output: $1\times28\times28$ image for $\mu_\rvxr(\rvyr,\rvzr;\param)$\\ 
        \bottomrule
    \end{tabular}}
\end{table*}
\begin{table*}[!htbp]
    \caption{Architecture for FFHQ. }
    \label{tab:archiFFHQ}
     \vspace{-0.3cm}
    \centering
  \resizebox{\textwidth}{!}{\begin{tabular}{ll}
    \toprule
        \textbf{Encoder $\varpdf_\paramA(\rvzr|\rvxr,\rvyr)$} &   \textbf{Decoder $\pdf_\param(\rvxr|\rvyr,\rvzr)$} \\
         \midrule
         Input: $2\times96\times96$ & Input: $\dim(\rvzr)+1=\dimdata+1$ values\\
         Conv2d: $32$ channels, $4\times4$ kernel, s=$2$, p=$1$ & FC: $256$ channels, $1\times1$ image \\
         Conv2d: $32$ channels, $4\times4$ kernel, s=$2$, p=$1$ &ConvT2d: $128$ channels, $3\times3$ kernel, s=$1$, p=$0$\\
         Conv2d: $64$ channels, $4\times4$ kernel, s=$2$, p=$1$ &ConvT2d: $64$ channels, $4\times4$ kernel, s=$2$, p=$1$\\
         Conv2d: $64$ channels, $4\times4$ kernel, s=$2$, p=$1$ &ConvT2d: $64$ channels, $4\times4$ kernel, s=$2$, p=$1$\\
         Conv2d: $128$ channels, $4\times4$ kernel, s=$2$, p=$1$ &ConvT2d: $32$ channels, $4\times4$ kernel, s=$2$, p=$1$\\
         Conv2d: $256$ channels, $3\times3$ kernel, s=$1$, p=$0$ &ConvT2d: $32$ channels, $4\times4$ kernel, s=$2$, p=$1$\\
         FC: $2\cdot \dim(\rvzr)=2\cdot\dimdata$ values& ConvT2d: $1$ channel, $4\times4$ kernel, s=$2$, p=$1$\\
Output:  $\dimdata$ values for $\mu_\rvzr(\rvxr,\rvyr;\paramA)$ and $\log \Sigma_\rvzr(\rvxr,\rvyr;\paramA)$& Output: $1\times96\times96$ feature map for $\mu_\rvxr(\rvyr,\rvzr;\param)$\\
        \bottomrule
    \end{tabular}}
\end{table*}

\subsection{Details of Baseline Models and Assets}\label{supp:baselines}

\noindent\textbf{Black-Box Baseline.}
The optimal baseline for evaluating the predictive performance of the GdVAE in \cref{tab:gdvaeEMvsIS} involves training a discriminative classifier and a CVAE  jointly. Unlike the GdVAE architecture, this classifier employs a separate neural network as its backbone. This backbone network replicates the structure of the CVAE's encoder (refer to  \cref{tab:archiCeleb,tab:archiCifar,tab:archiMNIST}), allowing the CVAE to learn an optimal representation for reconstruction, while the discriminative classifier learns an independent representation for classification. 

The classifier leverages the CVAE encoder without the additional class input to generate a latent representation $\rvzr$. Subsequently, this latent vector is processed by a one-layer fully connected neural network, which uses a softmax function to map $\rvzr$ to the class output.

\noindent\textbf{Importance Sampling (IS) \cite{NIPS2015_CVAE, classIncremental, NEURIPS2020_02ed8122} for Generative Classification.} An alternative approach to our EM-based inference is presented in \cite{classIncremental}, where they use separate VAEs for different classes and perform an importance sampling strategy to obtain a generative classifier. We extend this approach to CVAEs and use the importance sampling strategy as a baseline in \cref{tab:gdvaeEMvsIS}. The importance sampling strategy for CVAEs to approximate the likelihood $\pdf_\param(\rvxr|\rvyr)$ is given by
\begin{align}
	\pdf_\param(\rvxr|\rvyr)&=\expect{\rvzr\sim 	\pdf_\param(\rvzr|\rvyr) }{	\pdf_\param(\rvxr|\rvyr, \rvzr) } = \int \pdf_\param(\rvxr|\rvyr, \rvzr) \pdf_\param(\rvzr|\rvyr) d\rvzr \\
 &\approx \frac{1}{\numsamples} \sum_{\samples=1}^\numsamples \frac{\pdf_\param(\rvxr|\rvyr, \rvzr^{(\samples)}) \pdf_\param(\rvzr^{(\samples)}|\rvyr)}{\varpdf_\paramA(\rvzr^{(\samples)}|\rvxr,\rvyr)},
\end{align}
where the samples $\rvzr^{(s)}$ are drawn from $\varpdf_\paramA(\rvzr|\rvxr,\rvyr)$ and the likelihood is afterwards used in a Bayes' classifier $\pdf_\param(\rvyr|\rvxr)=  \eta \pdf_\param(\rvxr|\rvyr) \pdf_\param(\rvyr)$. The drawback of this approach is that in addition to the encoder it requires to invoke the decoder model for each sample and since the dimensionality of the  input space $\rvxr\in \mathbb{R}^\dimdatab$ is typically larger than the dimensionality of the latent space $\rvzr\in\mathbb{R}^\dimdata$, more samples are required for a good approximation. 

This baseline method is denoted by importance sampling (IS) in \cref{tab:gdvaeEMvsIS}. The architecture and learning objective remain unchanged, but instead of using \cref{algo:marg}, importance sampling in the image space is employed. For a fair comparison, we utilize $\numsamples=60$ for the importance sampling method. This is in line with the GdVAE, which uses $\numsamples=20$ samples but conducts $\numiter=3$ iterations, resulting in a total of 60 samples as well. 

\noindent\textbf{ProtoVAE \cite{Gautam22ProtoVAE} }(\href{https://github.com/SrishtiGautam/ProtoVAE}{https://github.com/SrishtiGautam/ProtoVAE}). We utilized the original implementation as detailed in \cite{Gautam22ProtoVAE}. In our adaptation, we substituted the backbone network with the GdVAE backbone and configured the model to work with a single prototype. Furthermore, we explored various hyperparameter settings, such as adjusting the loss balance to match our reconstruction loss and binary cross-entropy loss. The results are presented in \cref{tab:protoVAEParams}. 
 
\noindent\textbf{GANalyze \cite{Goetschalckx_2019_ICCV}} (\href{http://ganalyze.csail.mit.edu/}{http://ganalyze.csail.mit.edu/}). We utilized the original implementation as detailed in \cite{Goetschalckx_2019_ICCV}. The code already includes a PyTorch version, which is mainly used for the implementation of the so called transformer $T(z, \alpha)$. The transformer is equivalent to a linear explainer function
\begin{align}
\counterfactual_\discriminantfkt(\rvzr,\alpha, \idxclass)&={\rvzr}^{{\alpha}}=\rvzr+\alpha \weight^{(\idxclass)},
\end{align}
where $\alpha$ is the requested confidence for the counterfactual class and $\weight^{(\idxclass)}$ is the direction that is learned for each counterfactual class $\idxclass$. 
As in the original implementation we use a quadratic loss between the desired confidence and the classifier output and utilize the per sample loss
\begin{align}
\loss_{cf}(\rvxr^\alpha,\alpha,\idxclass)=\left[  \pdf_\param(\rvyr=\idxclass|\rvxr^\alpha)-\alpha\right]^2,
\label{eq:lossGANalyze}
\end{align}
where $\idxclass$ is the counterfactual class with respect to the class label $\class$ of the input image $\rvxr$, $\rvxr^\alpha=\counterfactual_\discriminantfkt(\encoder(\rvxr),\alpha, \idxclass)$ the counterfactual, and $\encoder$ the encoder. We optimize GANalyze by approximating the expectation in 
\begin{align}
\loss^{GANalyze}(\weight)&=\expect{\alpha\sim \mathcal{U}(0,1)}{\loss_{cf}(\rvxr^\alpha,\alpha,k)}
\end{align}
with 10 samples for each training image. GANalyze's primary objective is to learn a vector $\weight^{(\idxclass)}$ for each class, as opposed to a single vector as seen in GdVAE. Instead of using a GAN, we leverage our pre-trained GdVAE as an encoder, decoder, and classifier. The conditional decoder takes the counterfactual class $\idxclass$ as its input. Moreover, we have explored various hyperparameter settings. For instance, one such setting involves the use of normalization, a practice not documented in the paper but applied in the code. 

This normalization enforces the counterfactual distance from the origin to be identical to the original input $\rvzr=\encoder(\rvxr)$
\begin{align}
\counterfactual^{norm}_\discriminantfkt(\rvzr,\alpha,\idxclass)&= \frac{\counterfactual_\discriminantfkt(\rvzr,\alpha,\idxclass)}{\norm{\rvzr}_2}.
\end{align}
The results of various hyperparameter settings are presented in \cref{tab:baselineModels}. 

\noindent\textbf{UDID \cite{voynov2020unsupervised}} (\href{https://github.com/anvoynov/GANLatentDiscovery}{https://github.com/anvoynov/GANLatentDiscovery}). We utilized the original implementation as detailed in \cite{voynov2020unsupervised}. The code already includes a PyTorch version, which is mainly used for the implementation of the so called  latent deformator $A(\alpha e_\idxclass)$ and reconstructor $R(\rvxr,\rvxr^\alpha)$ that are employed for unsupervised latent space analysis. The deformater is a non-linear explainer function 
\begin{align}
\counterfactual_\discriminantfkt(\rvzr,\alpha, \idxclass)&={\rvzr}^{{\alpha}}=\rvzr+A(\alpha e_\idxclass),
\end{align}
where $e_\idxclass\in \mathbb{R}^\numclasses$ are standard unit vectors defining the directions, one for each class. $\alpha\sim \mathcal{U}(0,1)$ defines the strength of manipulation and $A(\cdot)\in ^{\dimdata\times \numclasses}$ is defined by a neural network. The primary objective of the reconstructor $R(\rvxr,\rvxr^\alpha)=(\hat{k}, \hat{\alpha})$ is to compare the original image $\rvxr$ and the manipulated version $\rvxr^\alpha$, aiming to replicate the manipulation in the latent space. Consequently, the method seeks to discover disentangled directions. Instead of employing a GAN, we utilize our pre-trained GdVAE as an encoder, decoder, and classifier. Additionally, we introduce a supervised learning component $\loss_{cf}$ to align the method with other supervised CF methods. We therefore employ a per-sample loss defined by
\begin{align}
\loss^{UD}(A)&=\expect{\alpha}{\loss_{cl}(k,\hat{k})+\lambda \loss_{r}(\alpha,\hat{\alpha})+\loss_{cf}(\rvxr^\alpha,\alpha,k)},
\end{align}
where $\lambda=0.25$ represents the weight for the regression task, $\loss_{cl}$ corresponds to the cross-entropy function, $\loss_{r}$ pertains to the mean absolute error, and $\loss_{cf}$ denotes the supervised loss for the counterfactuals. The supervised loss aligns with the one used in GANalyze (as in \cref{eq:lossGANalyze}). We use a sample size of 10 for $\alpha$ to approximate the expectation. The results of various hyperparameter settings are presented in \cref{tab:baselineModels}. To enhance the method and refine the proximity property, we incorporated a proximity loss $\loss_{prox}(\rvxr,\rvxr^\alpha)=\norm{\rvxr-\rvxr^\alpha}_2^2$, aiming to maintain close resemblance between the counterfactual and query images.

\noindent\textbf{ECINN \cite{ecinn2021}.}
ECINN employs an invertible neural network and applies a post-hoc analysis of class conditional means within the latent space to determine an interpretable direction. They make the assumption that the covariance matrix $\Sigma_\rvzr$ (see \cref{sec:VAEandGDA}) is the identity matrix. Their post-hoc method is similar to our local-L2 function, but it approximates the true classifier using empirical mean values. Consequently, we view ECINN as an empirical implementation of our method, additionally estimating the empirical covariance.

Furthermore, they create two counterfactuals: one with high confidence for the opposing class and another precisely at the decision boundary. We replicate and extend their approach by using our explainer function (refer to \cref{eq:localExplainer}), wherein we empirically determine class conditional mean values and the covariance matrix after training the GdVAE. Thus, we employ $\overline{\weight}=\overline{\covMatP}_{\rvzr}^{-1}(\overline{\mean}_{\rvzr|\idxclass}-\overline{\mean}_{\rvzr|\class})$, utilizing the empirical means $\overline{\mean}_{\rvzr|\cdot}$ and covariance $\overline{\covMatP}_{\rvzr}$ with the predicted class of the GdVAE as label. The formula for the counterfactual at the decision boundary is exactly the one that ECINN would employ. 

\noindent\textbf{AttFind \cite{Lang_2021_ICCV}} (\href{https://github.com/google/explaining-in-style}{https://github.com/google/explaining-in-style}).  We utilized the original implementation as detailed in \cite{Lang_2021_ICCV}. To accommodate our PyTorch-based model, we translated their TensorFlow implementation of the \textit{AttFind} method. It's important to emphasize that our usage of the AttFind method is solely for identifying significant directions within the latent space of our GdVAE. We do not employ their GAN architecture.

The AttFind method operates by iterating through latent variables, evaluating the impact of each variable on the classifier's output for a given image, and subsequently selecting the top-k significant variables. An image is considered explained once AttFind identifies a manipulation of the latent space that leads to a substantial alteration in the classifier's output, effectively changing the classifier's class prediction. We employ their \textit{Subset} search strategy, which focuses on identifying the top-k variables where jointly modifying them results in the most significant change in the classifier's output. This method is limited to producing class flips and is not intended for generating CFs with desired confidence values. Hence, we have labeled this method with AttFind$^\dagger$ to denote its original purpose of effecting class changes only. The results are presented in \cref{tab:consistencySimple}, though their accuracy and MSE may not be directly comparable to those in \cref{tab:consistency}.

\noindent\textbf{EBPE \cite{Singla2020Explanation}} (\href{https://github.com/batmanlab/Explanation_by_Progressive_Exaggeration}{https://github.com/$\ldots$by{\_}Progressive{\_}Exaggeration}).  
We utilized the original implementation as detailed in \cite{Singla2020Explanation}. To accommodate our PyTorch-based model, we translated their TensorFlow implementation. The most significant modification involves replacing the GAN with the CVAE architecture utilized by all approaches. EPBE, aside from GdVAE, uniquely required decoder training, resulting in distinct latent space characteristics and reconstruction quality. EBPE solely uses the pre-trained GdVAE for classification.

The original EBPE version was designed to work with the CelebA dataset, which is more complex than the MNIST dataset. The MNIST classifier we aimed to explain achieved remarkably high accuracy and confidence, resulting in some bins within the EBPE training having no samples for generating images at specific confidence values. When a bin lacked any samples, it was impossible to generate additional images from it. This issue could be addressed through hyperparameter tuning. For hyperparameter analysis, we define the loss
\begin{align}
\loss^{EBPE}&= \lambda_{cGAN}\loss_{cGAN}(G,D)+ \lambda_{rec}\loss_{rec}(G)\nonumber\\
&+ \lambda_{cyc}\loss_{cyc}(G)
+\lambda_{cfCls}\loss_{cfCls}(G)+\lambda_{recCls}\loss_{recCls}(G).
\end{align}

\noindent Here, $\loss_{cGAN}$ represents EBPE's conditional GAN loss, $\loss_{rec}$ is the reconstruction loss, $\loss_{cyc}$ denotes the cycle loss, $\loss_{cfCls}$ is the classification loss for the counterfactual, and $\loss_{recCls}$ stands for the classification loss for the reconstruction of the input image. $\lambda$ represents the weight of the corresponding loss. For further details, please see \cite{Singla2020Explanation} and the original implementation. Detailed results for different configurations can be found in \cref{tab:baselineModels}. 

As highlighted in the 'Limitations' section, the results showcased in the main paper utilized 24 epochs to maintain uniformity across all methods. For improved EBPE performance with extended training, see the 96-epoch results in \cref{tab:baselineModels}. 

\noindent\textbf{C3LT \cite{Khorram_2022_CVPR}} (\href{https://github.com/khorrams/c3lt}{https://github.com/khorrams/c3lt}).  
We utilized the original implementation as detailed in \cite{Khorram_2022_CVPR}. C3LT was initially designed for use with pre-trained models, and we applied this implementation to our GdVAE models. 

While C3LT was originally tailored for the simpler consistency task of class modification without explicitly requesting a user-defined confidence, we extended its functionality by introducing a supervised loss term, denoted as $\loss_{cf}$. This loss term serves to align the user-requested confidence level ($\alpha$) with the predicted confidence ($\hat{\alpha}$), with $\hat{\alpha}=\pdf_\param(\rvyr=\idxclass|\counterfactual_\discriminantfkt(\encoder(\rvxr),\alpha,\idxclass))$. For $\loss_{cf}$, we experimented with both quadratic (\cref{eq:lossGANalyze}) and cross-entropy functions to assess their performance. The results of these experiments can be found in \cref{tab:baselineModels}.

In conclusion, we leveraged C3LT to facilitate these modifications and added a supervised loss term to ensure alignment between the requested confidence and the model's predictions. The original version, without the added loss term, is referred to as C3LT$^\dagger$ in \cref{tab:consistencySimple}. 

\newpage 

\subsection{Compute Resources}\label{sec:compute}
For training our GdVAE and baseline models, we had the option to utilize two high-performance PCs equipped with multiple GPUs. The first PC featured an Nvidia RTX 3090 and a Titan RTX, both boasting 24 GB of memory each. The second PC made use of two Nvidia A6000 GPUs, each equipped with 48 GB of memory. To establish a reference point for the required computational time, we considered the GdVAE exclusively for the counterfactual tasks, as they represent an upper limit for the models' demands.

For an individual epoch on binary MNIST, it took 5 minutes, and for CelebA, it took 60 minutes on the A6000. In the context of our hyperparameter sweep, we explored 25=(5+8+6+6) different model configurations for MNIST and 12 for CelebA, encompassing parameters like the number of samples $\numsamples$, iterations $\numiter$, the balance between $\nicefrac{\alpha}{\beta}$, and the consistency weight $\gamma$. Each of these configurations underwent four separate training runs to calculate the mean and standard deviations, ultimately leading to the training of 100 models for MNIST and 48 for CelebA, respectively.

Consequently, the total computational time required for this parameter analysis sums up to (5 $\nicefrac{min}{epochs}$ $\cdot$ 100 $\cdot$ 24 ${epochs}$ + 60 $\nicefrac{min}{epochs}$ $\cdot$ 48 $\cdot$ 24 ${epochs}$) = 81120 ${min}$, which is equivalent to 1352 $h$ of GPU time. 

\section{Evaluation Metrics}\label{supp:metrics}

\subsection{Metrics for Predictive Performance in \cref{tab:gdvaeEMvsIS}}\label{supp:metricsPred}

\noindent\textbf{Accuracy (ACC).} In \cref{tab:gdvaeEMvsIS}, we assess the predictive performance of the classifier using the accuracy (ACC) metric, defined as
\begin{align}
ACC &= \frac{1}{\numdata}\sum_{\idxdata=1}^\numdata \indicator{\hat{\rvyr}^{(\idxdata)}=\rvyr^{(\idxdata)}}, 
\label{eq:acc}
\end{align}
with the indicator function $\ind\{\cdot\}$. Here, $\hat{\rvyr}$ is the class prediction, determined as the class with the highest probability according to $\hat{\rvyr}=\argmax_\idxa \pdf_\param(\rvyr=\idxa|\rvzr)$. The ground-truth class labels are denoted by $\rvyr\in \{1, \ldots, \numclasses\}$ and we have $\numdata$ test data samples. $\rvyr^{(\idxdata)}$ represents the $\idxdata$-th sample.  

\noindent\textbf{Mean Squared Error (MSE).} To evaluate the reconstruction quality in \cref{tab:gdvaeEMvsIS}, we calculate the mean squared error (MSE) between the ground-truth input image $\rvxr$ and the reconstructed image $\hat{\rvxr}$
\begin{align}
MSE &= \frac{1}{\numdata\cdot \dimdatab} \sum_{\idxdata=1}^\numdata \sum_{\idxa=1}^\dimdatab \left(\rvxr_\idxa^{(\idxdata)}-\hat{\rvxr}_\idxa^{(\idxdata)}\right)^2. 
\label{eq:mse}
\end{align}
We assume images to be vectorized, with $\rvxr\in \mathbb{R}^\dimdatab$. Here, $\dimdatab$ is defined as the product of the image's width ($W$), height ($H$), and number of channels ($C$), i.e., $\dimdatab=W\cdot H\cdot C$. $\rvxr^{(\idxdata)}$ represents the $\idxdata$-th sample.  

\subsection{Metrics for CF Explanations in \cref{tab:consistency}}\label{supp:metricsCF}

\textbf{Realism.} Realism in counterfactuals is essential, as they should resemble natural data. To assess realism, we employ the \textit{Fr\'{e}chet Inception Distance (FID)} metric, a standard measure for this purpose \cite{Khorram_2022_CVPR, ghandeharioun2022dissect, Singla2020Explanation}. We compute the FID values using the PyTorch implementation from \cite{Seitzer2020FID}. 

\noindent\textbf{Consistency.} CF explanations aim to influence the behavior of a classifier to obtain desired outcomes. In case of our method the consistency task is to guarantee that the requested confidence value ${\pdf}_\class$ accurately matches to the confidence prediction of the classifier $\widehat{\pdf}_\class=\pdf_\param(\rvyr=\class|\encoder(\decoder(\rvzr^\delta)))$ for the CF ${\rvzr}^{\delta}=\counterfactual_\discriminantfkt(\rvzr,\delta)$. $\class$ is consistently assigned to the class label of the original input. 

As demonstrated in \cite{Singla2020Explanation}, a method for assessing consistency is to create a plot that compares the expected classifier outcomes with the confidence predictions of the classifier for the generated CF. The optimum is reached when we obtain an identity relationship (${\pdf}_\class=\widehat{\pdf}_\class$) between the two quantities. We use kernel density estimates (KDE) to visualize this relationship (\cref{fig:consistencyGammaALL,fig:KDEMnist}). 

Similar to \cite{ghandeharioun2022dissect}, we quantitatively evaluate the existence of a linear relationship using the \textit{Pearson correlation coefficient}. In addition, we utilize the \textit{mean squared error (MSE)} between the desired ${\pdf}_\class$ and the estimated confidence $\widehat{\pdf}_\class$
\vspace{-0.5cm}
\begin{align}
MSE &= \frac{1}{\numdata^\star} \sum_{\idxdata=1}^{\numdata^\star}  \left({\pdf}_\class^{(\idxdata)}-\widehat{\pdf}_\class^{(\idxdata)}\right)^2.  
\end{align}
Here, we have $\numdata^\star$ samples and ${\pdf}_\class^{(\idxdata)}$ represents the $\idxdata$-th sample. Given that we request confidence values within the range $\pdf_\class\in[0.05,0.95]$, with a step size of $0.05$, we acquire 19 counterfactuals per test image. As a result, $\numdata^\star=19\cdot\numdata$, where $\numdata$ represents the number of test images.

\textit{Accuracy (ACC).} Finally, the accuracy metric, as defined in \cite{Khorram_2022_CVPR} and denoted by "Val", is designed for the simpler consistency task, which assesses only class flips as a binary classification problem. To evaluate continuous confidence requests, we employ 12 bins $b\in\{1,\ldots,12\}$ and treat the bin assignment of the prediction as a multi-class problem. Therefore, we use \cref{eq:acc} with ground-truth bins $b^{(\idxdata)}$ and predicted bins $\hat{b}^{(\idxdata)}$ for the $\numdata^\star$ samples. 

In particular, ACC and MSE are used to gauge whether the counterfactuals are predominantly generated at the extreme confidence levels, near one or zero. While the results may not be directly comparable, in the context of methods that focus solely on generating class flips, such as C3LT$^\dagger$ \cite{Khorram_2022_CVPR} and AttFind$^\dagger$ \cite{Lang_2021_ICCV}, we adopt accuracy with two bins as a reference point, as per the definition in \cite{Khorram_2022_CVPR} for binary classification. Please note that this task is considerably simpler, resulting in accuracy values in \cref{tab:consistencySimple} being close to $100\%$ for methods marked with $\dagger$.

\noindent\textbf{Proximity.} The CF should only change the input in a minimal way $\rvxr^{{\delta}}=\argmin_{\rvxr'} \userfkt(\rvxr,\rvxr')$, with respect to some user defined quantity $\userfkt(\cdot)$, \eg, $\userfkt(\rvxr,\rvxr')=\norm{\rvxr-\rvxr'}_2$ \cite{black2022consistent}. 

\textit{Mean Squared Error (MSE).} In \cite{Khorram_2022_CVPR} they measure proximity by means of the $L1$-norm between the query image $\rvxr$ and the counterfactuals $\rvxr^\delta$. We adopt this metric by using \cref{eq:mse} for the $\numdata^\star$ counterfactual samples.   

\section{Additional Results}\label{supp:results}
\subsection{GdVAE Hyperparameter Analysis}\label{supp:GdVAEAnalysis}
\textbf{Parameterization of the Loss Function.} Before arriving at the final loss, we conducted several initial experiments on the MNIST dataset. Initially, we compared our loss formulation (A) derived from \cref{eq:gdvaeLossOriginal} with the loss formulation (B1) outlined in \cref{sec:objective}, where the training process is streamlined with the inference process. Additionally, we fine-tuned the reconstruction and classification loss by adjusting the scale of the reconstruction loss using $\pdf_\param(\rvxr|\rvyr,\rvzr)=\gaussian\left(\mu_\rvxr(\rvyr,\rvzr ;\param), \Sigma_\rvxr(\rvyr,\rvzr ;\param)\right)$, with $\Sigma_\rvxr:=\Sigma_\rvxr(\rvyr,\rvzr ;\param)=0.6^2\cdot I$ for likelihood calculation (as in \cite{falck2021multifacet}), instead of the standard $\Sigma_\rvxr=I$. Furthermore, the cross-entropy loss for classification was rescaled in proportion to the image size, using the factor $0.1\cdot W\cdot H\cdot C$ (B2), where $W$, $H$, and $C$ represent the width, height, and channels of the input image, respectively. Finally, using two priors $p(z|y)$ and $p(z)$ is not essential, but they are part of the model. In our probabilistic view, the used factorization (see \cref{sec:method}) justifies including $p(y|z)$ (and thereby $p(z)$), in contrast to works that merely append $p(y|.)$. However, using an uniform prior for $p(z)$, akin to excluding $p(z)$ from the loss, is valid. For MNIST, incorporating the normal prior results in a lower reconstruction error. 

The results pertaining to these settings, including classification accuracy ($ACC$) and mean squared error ($MSE$) as a metric for reconstruction quality, are presented in \cref{tab:initalTests}. We adopted the B2 setting for all other experiments and datasets without dataset-specific fine-tuning.

\begin{table}
 \vspace{-0.4cm}
\caption{Evaluation of different loss functions and settings of the GdVAE on the MNIST dataset. We use $\alpha=\beta=1$, $\numsamples=20$, and $\numiter=3$. MSE is scaled by a factor of $10^2$. Mean values, including standard deviation, are reported over four training processes with different seed values.}\label{tab:initalTests}
	\centering
 \vspace{-0.2cm}
		\renewcommand{\arraystretch}{1.2}
	\begin{tabular}{cccc}
 	      \toprule
		 Setting & {Details} &  $ACC\%\uparrow$ & $MSE\downarrow$ \\
    	\midrule
    \textbf{A:} & \cref{eq:gdvaeLossOriginal}, $\Sigma_\rvxr=I$&88.6\footnotesize$\pm$1.25 & 1.19\footnotesize$\pm$0.04   \\
    \textbf{B1:} & \cref{eq:lossgdvae,eq:lossgdvae2}, $\Sigma_\rvxr=I$ & 96.0\footnotesize$\pm$0.85 & 1.04\footnotesize$\pm$0.02   \\
    \textbf{B2:} & \cref{eq:lossgdvae,eq:lossgdvae2}, $\Sigma_\rvxr=0.6^2\cdot I$ & 99.0\footnotesize$\pm$0.11 & 1.10\footnotesize$\pm$0.04   \\	
	\textbf{B2 w/o $p(z)$:} & \cref{eq:lossgdvae,eq:lossgdvae2}, $\Sigma_\rvxr=0.6^2\cdot I$ &  99.0\scriptsize$\pm$0.08 & 1.14\scriptsize$\pm$0.03    \\	
	\bottomrule
 \end{tabular}
  \vspace{-0.2cm}
\end{table}

\begin{figure}[!htbp]
    \centering
    \includegraphics[width=0.5\linewidth]{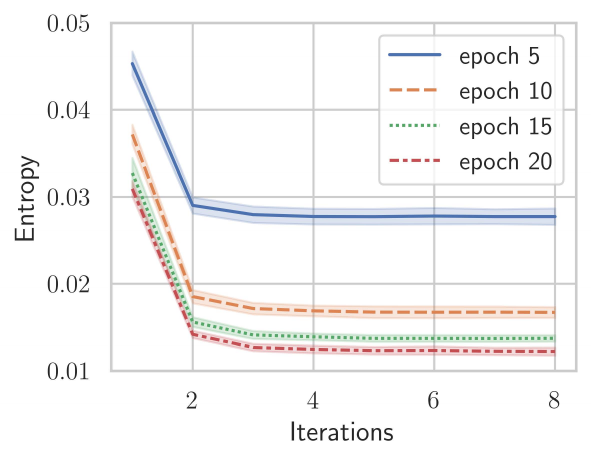}
	\caption{Evaluation of number of iterations $\numiter$ during the training process on the MNIST dataset. Mean values, including standard error, are reported over four training processes with different seed values.}
	\label{fig:iteration_relative_entropy}
\end{figure}

\noindent \textbf{How to Parameterize the EM-Based Algorithm?} The GdVAE (\cref{algo:marg}) requires user-defined values for the number of iterations ($\numiter$) and samples ($\numsamples$). To determine the optimal number of iterations, we assess the stability of the training process on the MNIST dataset. The evaluation involves measuring the entropy of $\varpdf_\paramA(\rvyr|\rvxr)$ during training epochs and comparing successive epochs. The results, depicted in \cref{fig:iteration_relative_entropy}, illustrate the change in entropy with increasing iterations, averaged over the entire test dataset. The error bars represent the standard error across four training runs with different seeds. Convergence is observed after 3 iterations. Other parameters are set to $\alpha=\beta=\numsamples=1$. Based on the convergence analysis, we fix the value of $\numiter$ to 3 for subsequent experiments.

\begin{figure*}[ht]
	\centering
	\begin{subfigure}{0.33\textwidth}
		\begin{overpic}[grid=false, width=0.96\textwidth]{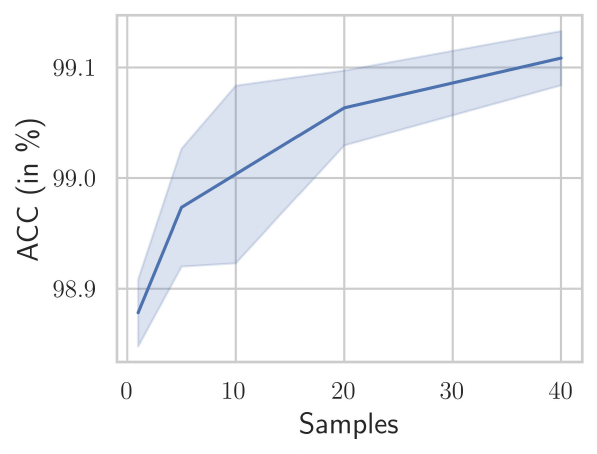}%
		\end{overpic}
	\end{subfigure}%
	\hfill
	\begin{subfigure}{0.33\textwidth}
		\begin{overpic}[grid=false, width=0.98\textwidth]{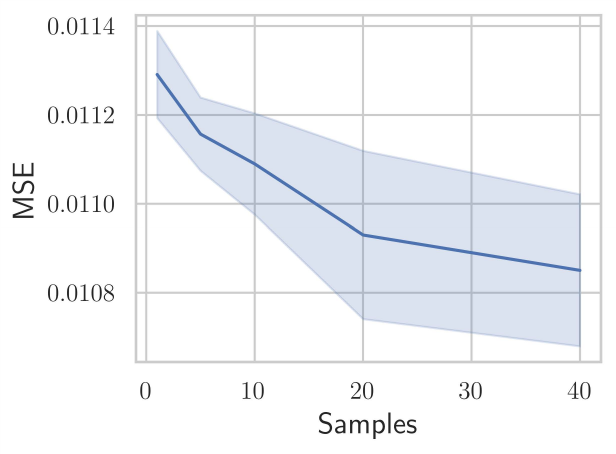}
		\end{overpic}
	\end{subfigure}
	\hfill
	\begin{subfigure}{0.33\textwidth}
		\begin{overpic}[grid=false, width=1.0\textwidth]{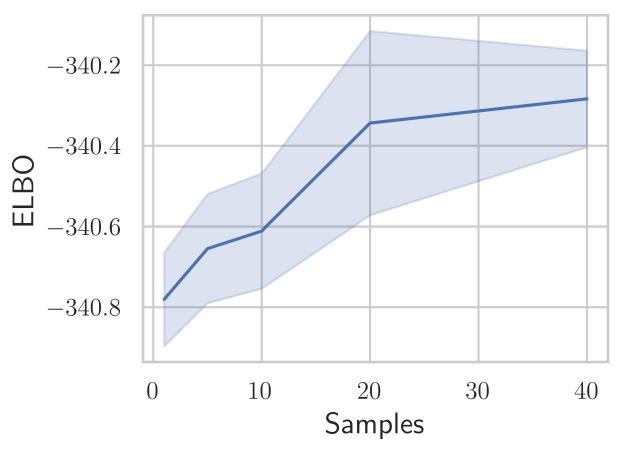}
		\end{overpic}
	\end{subfigure}
	
	\caption{Quality of models trained on the MNIST dataset with a different number of samples $\numsamples$.}
	\label{fig:sampleamount}
\end{figure*}

After assessing convergence, we examine the influence of the number of samples $\numsamples$ drawn from $\varpdf_\paramA(\rvzr|\rvxr)$ on performance. We evaluate classification accuracy ($ACC$), reconstruction error ($MSE$), and the average ELBO of M1 and M2 through multiple training sessions with different random seeds using $\numsamples \in \{1,5,10,20,40\}$ and report the average values along with the standard error. The results, depicted in \cref{fig:sampleamount}, demonstrate the expected improvement in performance with an increasing number of samples. All model trainings are stable and result in accuracy values $\geq 98\%$ and for $S\in \{20,40\}$ comparable results are obtained with a small standard error regarding classification accuracy. By selecting $\numsamples=20$, a balanced trade-off between computational load and performance is achieved, as indicated by the evidence lower bound (ELBO).

\noindent\textbf{How to Balance the Loss for Models M1 and M2?}
The proposed GdVAE model incorporates two types of loss functions: the M1 and M2 losses. Both losses include the reconstruction loss, but differ primarily in their impact on the recognition model $\varpdf_\paramA(\rvzr|\rvxr,\rvyr)$ and the classifier. While M1 focuses on training the likelihood model for purely generative classification, M2 provides a discriminative training signal for the generative classifier and enforces a standard normal distribution in the latent space. As described in \cref{sec:objective}, we ensure alignment between the training and inference processes, requiring both models to utilize the EM-based classifier for calculating the reconstruction loss using $\pdf_\param(\rvxr|\rvzr,\rvyr)$.
For the EM algorithm to function properly, a prerequisite is that, given an input image $\rvxr$, the recognition model $\varpdf_\paramA(\rvzr|\rvxr,\rvyr)$ must either have the lowest Kullback-Leibler divergence $\KL{\varpdf_\paramA(\rvzr|\rvxr,\rvyr)}{\pdf_\param(\rvzr|\rvyr)}$ for the correct class or be independent of $\rvyr$ (\eg, \cite{falck2021multifacet}). The M1 loss does not consider these aspects, as it solely focuses on minimizing $\KL{\varpdf_\paramA(\rvzr|\rvxr,\rvyr)}{\pdf_\param(\rvzr|\rvyr)}$ for the correct class without enforcing it to be smaller than for the other classes. The desired behavior is enforced by adding the M2 loss, which incorporates a discriminative training signal and a KL divergence term that promotes independence. Furthermore, the classifier utilized in M2 relies on the likelihood function learned in M1, highlighting the interdependence between both models.

\begin{figure*}[!t]
	\centering
	\begin{subfigure}{0.32\textwidth}
		\begin{overpic}[grid=false, width=1\textwidth]{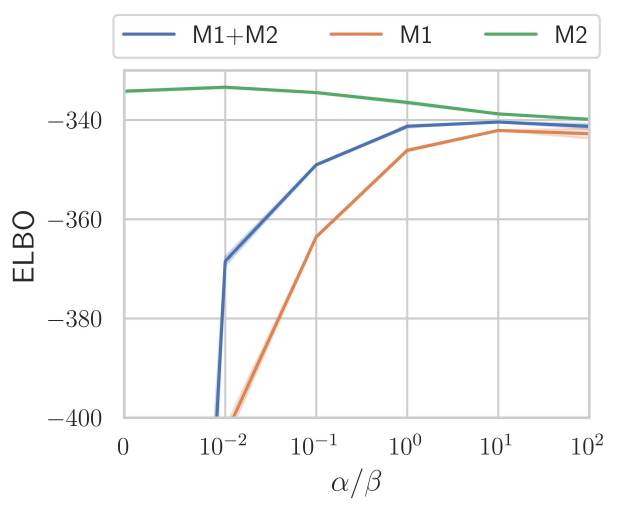}%
		\end{overpic}
		  \caption{MNIST}
	\end{subfigure}%
	\hfill
	\begin{subfigure}{0.32\textwidth}
		\begin{overpic}[grid=false, width=1\textwidth]{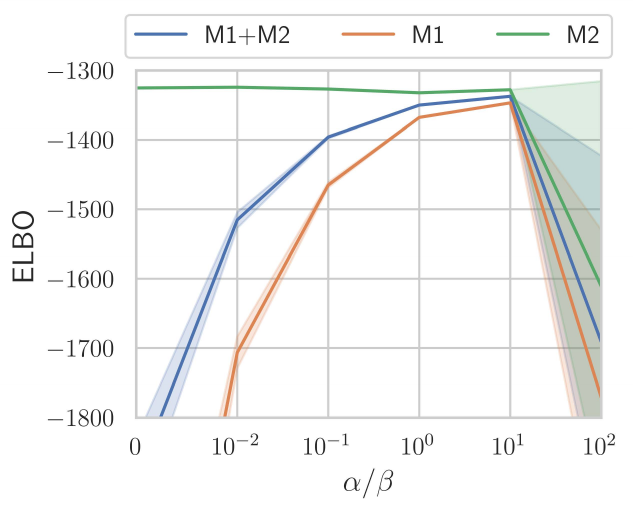}%
		\end{overpic}
			\caption{CIFAR-10}
	\end{subfigure}
	\hfill
	\begin{subfigure}{0.32\textwidth}
		\begin{overpic}[grid=false, width=1\textwidth]{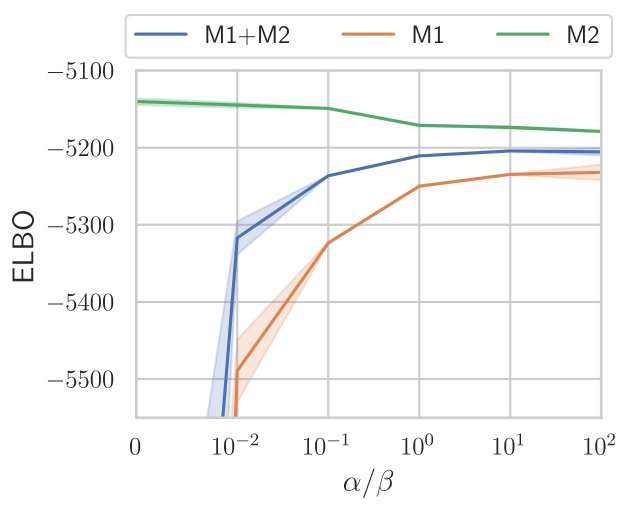}%
	\end{overpic}
	\caption{CelebA}
	\end{subfigure}
	\caption{Analysis of the optimal balance between models M1 and M2 on MNIST, CIFAR-10, and CelebA datasets. Mean values, including standard error, are reported over four training processes with different seed values.}
	\label{fig:ablationAB}
\end{figure*}

We analyze the optimal interplay between models M1 and M2 by evaluating the parameterization of the loss function defined by \cref{eq:lossgdvae,eq:lossgdvae2}, where we select suitable values for  $\alpha$ and $\beta$. To keep notation concise, we use the ratio $\nicefrac{\alpha}{\beta}$ instead of the individual values. We assess combinations of $\alpha$ and $\beta$ from the set $\left\{0,1,10,100\right\}$ and present the ELBO results graphically in \cref{fig:ablationAB}. The results indicate that assigning equal weight values to the models ($\nicefrac{\alpha}{\beta}=1$) or using $\nicefrac{\alpha}{\beta}=10$ generally yields good performance across diverse datasets. To slightly enhance classification accuracy, we opted for $\nicefrac{\alpha}{\beta}=1$ in all experiments discussed in the main paper. For the MNIST dataset, opting for $\nicefrac{\alpha}{\beta}=1$ led to an accuracy of $99.0\%$, as opposed to $98.8\%$ with nearly identical reconstruction error. Similar accuracy improvements were observed for CIFAR-10 ($65.1\%$ and $63.4\%$) and CelebA ($96.7\%$ and $96.4\%$) datasets. These results are used in \cref{tab:gdvaeEMvsIS}.

\noindent\textbf{How to Balance the Consistency Loss?} The experiments aim to assess the quality of counterfactuals (CF) when varying the impact of the consistency loss. As in the main paper, we employ the Fréchet Inception Distance (FID) to gauge \textit{realism} and Pearson and Spearman's rank correlation coefficients to gauge \textit{consistency}. Following a similar approach to \cite{Singla2020Explanation}, we also visualize the requested versus the actual response of the classifier using a kernel density estimate plot. 

\begin{figure*}[!t]
	\centering
	\begin{subfigure}{0.33\textwidth}
		\begin{overpic}[grid=false, width=1\textwidth]{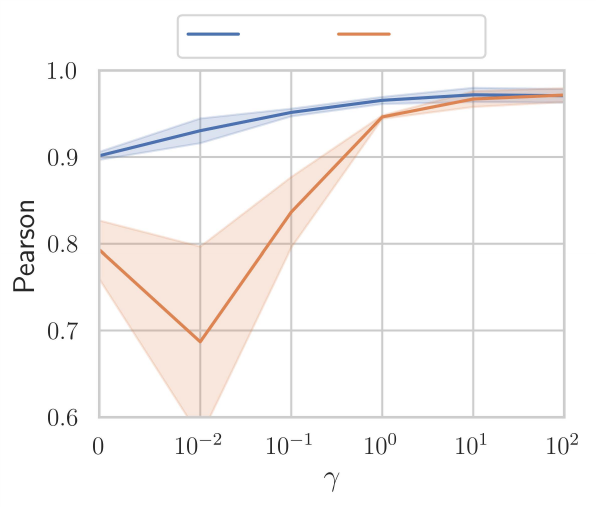}			\put(40.5,78.2) {\tiny {global}}
		\put(67.5,78.2)  {\tiny  {local}}
		\end{overpic}
	\end{subfigure}%
	\hfill
	\begin{subfigure}{0.33\textwidth}
	\begin{overpic}[grid=false, width=1\textwidth]{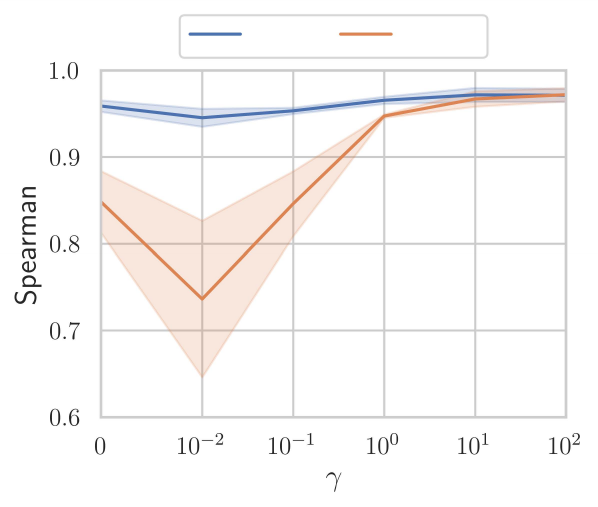}
 \put(40.7,77.6) {\tiny {global}}
		\put(67.,77.6)  {\tiny  {local}}
	 	\end{overpic}
	  \end{subfigure}
	\hfill
	 \begin{subfigure}{0.33\textwidth}
		\begin{overpic}[grid=false, width=1.0\textwidth]{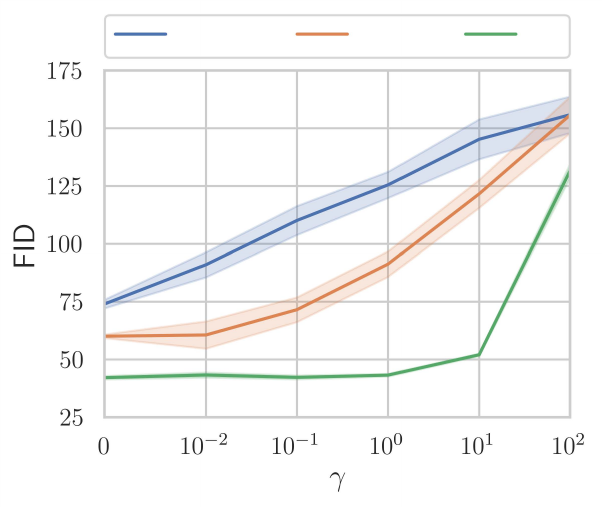}
            \put(28.4,77.3) {\tiny {global}}
		\put(59.8,77.3)  {\tiny  {local}}	
  		\put(87.8,77.3)  {\tiny  {$\rvxr^\star$}}	
		\end{overpic}
	\end{subfigure}
	\begin{subfigure}{1\textwidth}
		\begin{overpic}[grid=false, width=1.0\textwidth]{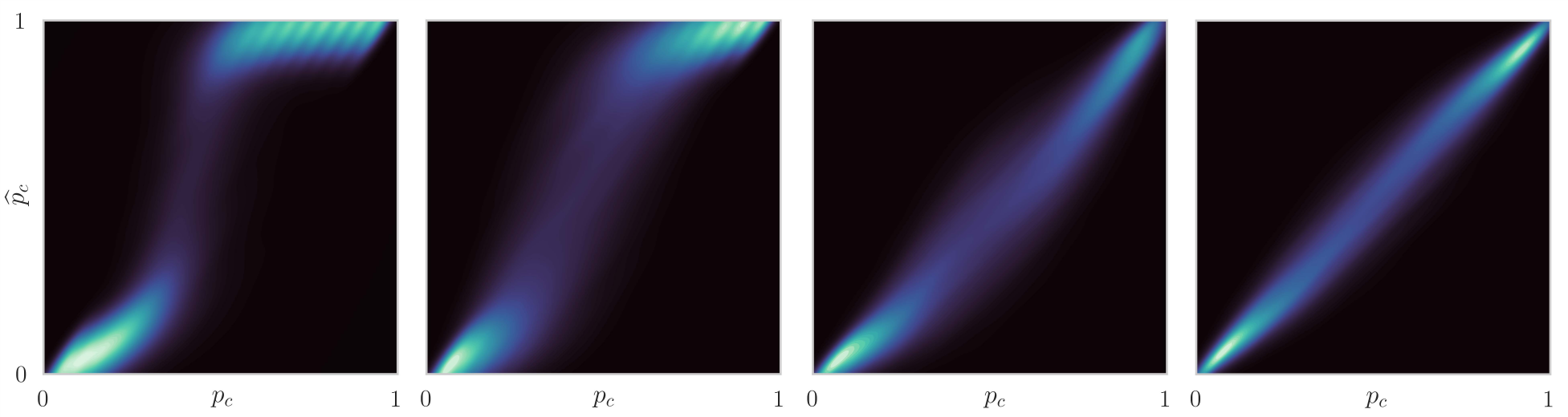}%
			\put(4.5,22) {\color{white}  {$\bm\gamma=0.0$}}%
			\put(4.5,19) {\color{white}  {$\bm\rho_p=0.9$}}%
			\put(29,22) {\color{white} {$\bm\gamma=0.01$}}%
			\put(29,19) {\color{white} {$\bm\rho_p=0.93$}}%
			\put(54,22) {\color{white} {$\bm\gamma=0.1$}}%
			\put(54,19) {\color{white} {$\bm\rho_p=0.95$}}%
			\put(78,22) {\color{white} {$\bm\gamma=1.0$}}%
			\put(78,19) {\color{white} {$\bm\rho_p=0.97$}}%
		\end{overpic}
	\end{subfigure}
	\caption{Consistency of counterfactual examples using the MNIST dataset. 
Top: Pearson's $\rho_p$ and Spearman correlation coefficients assess the relationship between the requested confidence and the classifier's output for counterfactual examples. Significance in correlation is indicated by a $p-value \leq 0.001$. Similarly, the Fr\'{e}chet Inception Distance (FID) quantifies the image quality of the generated counterfactuals in comparison to the real data. Mean values, including standard error, are reported over four training processes with different seed values. Bottom: Consistency of the requested confidence versus the actual classifier confidence is visually depicted using a kernel density estimate (KDE) plot of the observations. The desired confidence output of the classifier for a counterfactual example $\rvxr^{\delta}$ is specified by $\pdf_\class$, while the actual confidence acquired by inputting the counterfactual to our classifier is represented by $\widehat{\pdf}_\class = {\pdf}(\rvyr = \class | \rvxr^{\delta})$. Additionally, the corresponding Pearson correlation coefficient $\rho_p$ is visualized.}
	\label{fig:consistencyGammaALL}
\end{figure*}

The ablation study, examining the influence of the consistency loss, is depicted in \cref{fig:consistencyGammaALL}. We assess weight values $\lambda$ for the consistency loss from the set $\{0.0, 0.01, 0.1, 1.0, 10, 100\}$. The error bars represent the standard error across four training runs with different seeds. We present results for both our local-L2 and global CF generation methods, as described in the main paper \cref{sec:CFExplanation}. $\rvxr^\star$ serves as the FID baseline, representing values obtained by applying the encoder and decoder directly to the test data. Thus, the FID score for $\rvxr^\star$ reflects the reconstructed test set. 

\begin{figure}[!t]
    \centering
    \includegraphics[width=0.5\linewidth]{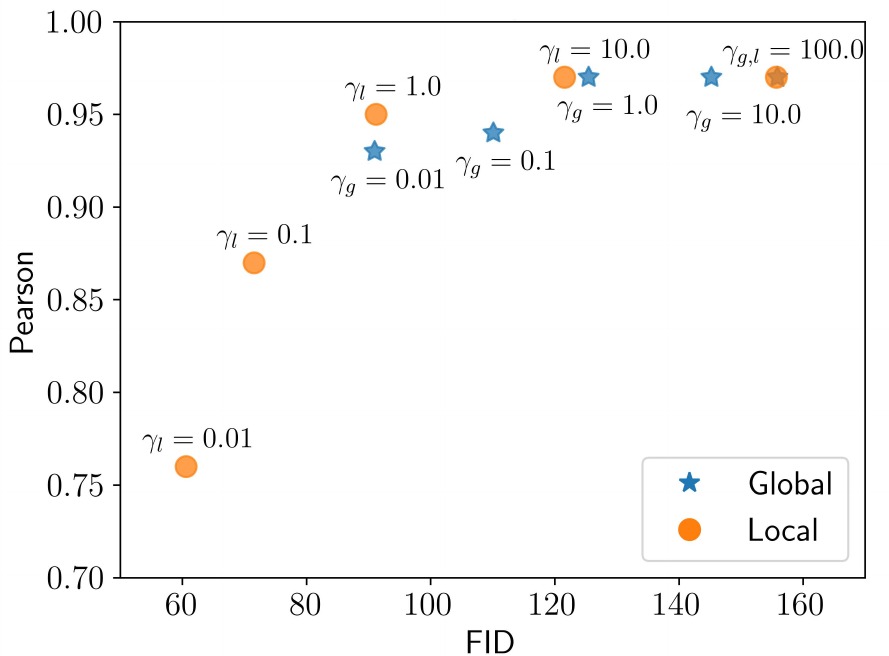}
    \caption{Pearson correlation against FID scores on the MNIST dataset, where $\gamma_l$ represents the consistency weight for the local method, and $\gamma_g$ for the global one.}
    \label{suppfig:consistencyVsFID}
\end{figure}

The ablation study on the consistency loss in \cref{fig:consistencyGammaALL} reveals a trade-off between \textit{consistency} and \textit{realism}. When the consistency parameter exceeds $\lambda=1$, the FID score of the ground truth reconstructions is significantly affected. Similarly, as the influence of the consistency regularizer increases, the correlation and FID value also increases. In \cref{suppfig:consistencyVsFID}, we visualize this trade-off by directly plotting the Pearson correlation against the FID scores.  

\begin{figure*}[!ht]
    \centering
    \includegraphics[width=0.75\linewidth]{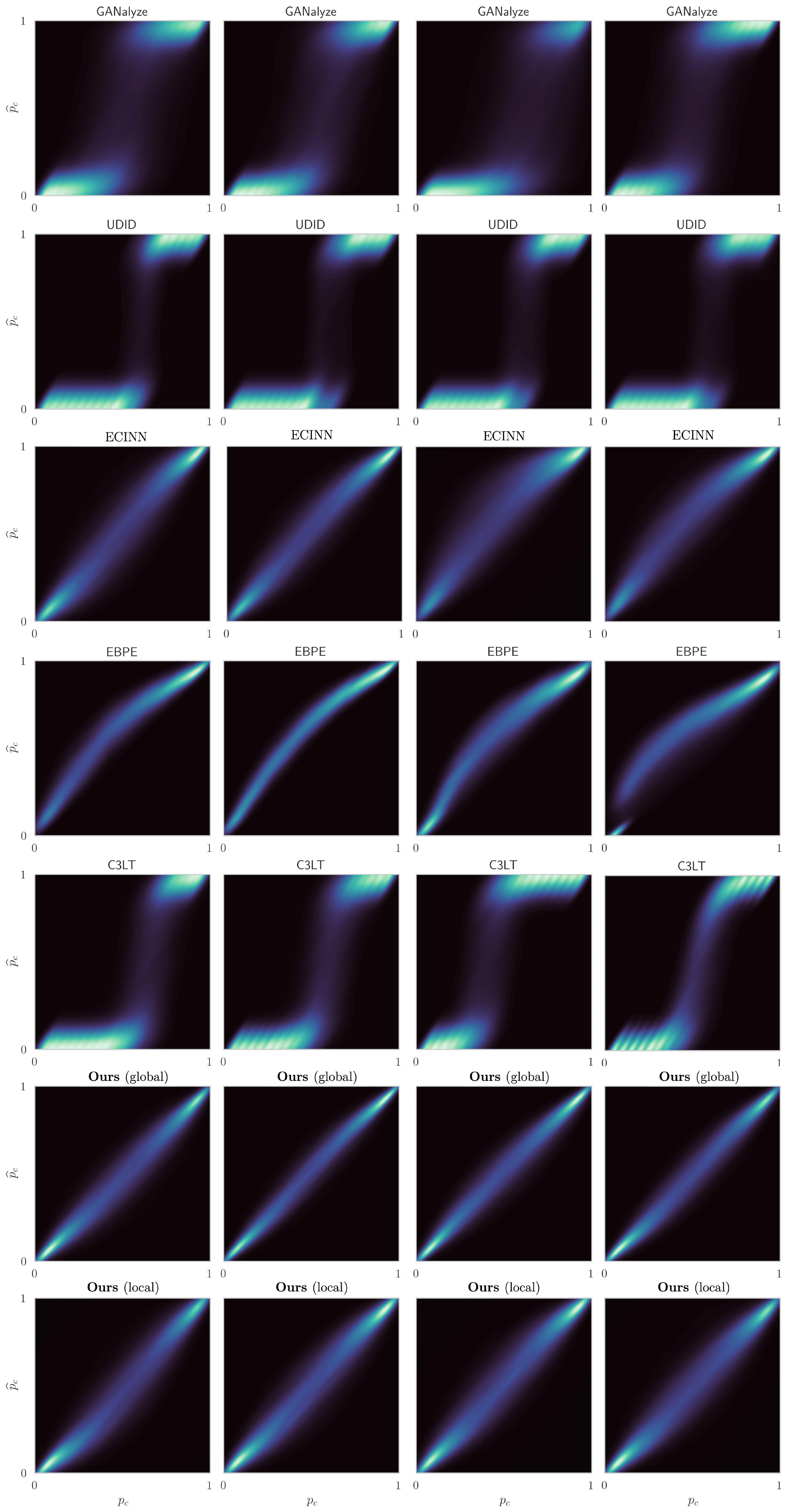}
    \caption{MNIST KDE plots are employed to visually depict the relationship between the requested confidence value ${\pdf}_\class$ and the predicted confidence by the classifier $\widehat{\pdf}_\class = \pdf_\param(\rvyr=\class|\encoder(\decoder(\rvzr^\delta)))$ for the counterfactual ${\rvzr}^{\delta} = \counterfactual_\discriminantfkt(\rvzr,\delta)$.}
    \label{fig:KDEMnist}
\end{figure*}

The KDE plot uncovers an intriguing observation: even without utilizing the consistency regularizer, reasonable counterfactual examples can be generated with high correlation values ($\rho_p\approx0.9$). Interestingly, counterfactuals tend to be generated at the extreme ends of the confidence range, near one or zero. Therefore, the method effectively flips the class of the query image, but the confidence values are not well calibrated. With increasing correlation values and improved calibration, the realism measure (FID) yields inferior results. One possible explanation is that more counterfactuals are generated near the decision boundary, where there is less real data available, resulting in a compromised natural appearance. It becomes evident that, for the consistency task \cite{Samangouei18ExplainGAN, Khorram_2022_CVPR, Lang_2021_ICCV} in which the user solely pre-defines the class label without specifying the confidence level, generating realistic images with low FID scores is considerably easier. Furthermore, we observe that global CF generation exhibits greater consistency with the classifier but lags behind in terms of realism when compared to the local CF generation process. The KDE plots for each model's four training runs are displayed in \cref{fig:KDEMnist}. These are the models used in \cref{tab:consistency}.

\subsection{Analysis of Baseline Models}\label{supp:BaselineAnalysis}

We conducted explorative parameter tuning for all baseline models, initially using the parameterization from the original implementation. Subsequently, we fine-tuned the parameters based on the results to achieve a balanced performance across various metrics and datasets. Results regarding ProtoVAE can be found in \cref{tab:protoVAEParams} and the results for the counterfactual methods are presented in \cref{tab:baselineModels,tab:consistencySimple}. Additionally, \cref{tab:consistencyRanges} presents the GdVAE realism results for various ranges of query confidences for comparison with \cref{tab:baselineModels}. In the main paper, the FID scores for $p_c\in [0,1]$ are reported.

\begin{table*}[!htbp]
\caption{CF explanations across diverse baseline model configurations. Mean values, with standard deviation, are reported across four training runs with different seeds. The configurations used in the main paper's experiments (\cref{tab:consistency}) are \textbf{bolded}. }
\label{tab:baselineModels}
	\centering
  		\renewcommand{\arraystretch}{1.0}
  		\resizebox{\textwidth}{!}{\begin{tabular}{cc|ccc|ccc}
			\cline{2-8}  
        & \multicolumn{1}{c|}{\multirow{2}{*}{Setup}} & \multicolumn{3}{c|}{Consistency} & \multicolumn{3}{c}{Realism (FID) $\downarrow$}\\
         &	&   {$\rho_p$ $\uparrow$} & $ACC\%$ $\uparrow$& $MSE$ $\downarrow$ & $\widehat{\pdf}_\class\notin[0.1,0.9]$  &   $\widehat{\pdf}_\class \in[0.1,0.9]$  &   $\widehat{\pdf}_\class \in[0,1]$ \\
		\cline{2-8}
    & \multicolumn{7}{c}{\textbf{GANalyze \cite{Goetschalckx_2019_ICCV}}} \\
   \cline{2-8}

  \parbox[t]{16mm}{\multirow{3}{*}{\rotatebox[origin=c]{0}{{\specialcell[c]{MNIST\\ Binary 0/1}}}}} 
 
  & A & 0.48\footnotesize$\pm$0.04 & 2.8\footnotesize$\pm$0.8 &  19.33\footnotesize$\pm$1.31 & 79.88\footnotesize$\pm$10.04 & 110.36\footnotesize$\pm$8.54 & 55.77\footnotesize$\pm$7.46 \\
  & \textbf{B1} & 0.84\footnotesize$\pm$0.04 & 5.5\footnotesize$\pm$1.3 & 6.75\footnotesize$\pm$1.27  & 51.65\footnotesize$\pm$5.25 & 96.20\footnotesize$\pm$8.93 & 54.89\footnotesize$\pm$4.19\\
  & B2 & 0.93\footnotesize$\pm$0.03 & 25.9\footnotesize$\pm$16.8 & 2.09\footnotesize$\pm$1.35  & 122.16\footnotesize$\pm$17.74 & 158.01\footnotesize$\pm$27.19 & 132.36\footnotesize$\pm$23.35\\
	\cline{2-8}
     & \multicolumn{7}{c}{\textbf{UDID \cite{voynov2020unsupervised}}} \\
   \cline{2-8}

  \parbox[t]{16mm}{\multirow{5}{*}{\rotatebox[origin=c]{0}{{\specialcell[c]{MNIST\\ Binary 0/1}}}}} 
 
    & A & 0.87$\pm$0.02 & 13.1$\pm$5.9 & 5.91$\pm$1.04 & 159.16$\pm$21.48 & 181.54$\pm$29.72 & 155.37$\pm$17.39 \\
    & B1 & 0.87\footnotesize$\pm$0.03 & 0.6\footnotesize$\pm$0.2 & 8.47\footnotesize$\pm$0.12 & 46.74\footnotesize$\pm$4.34 & 104.99\footnotesize$\pm$12.26 & 42.98\footnotesize$\pm$3.93 \\
    & \textbf{B2} & 0.85\footnotesize$\pm$0.01 & 1.2\footnotesize$\pm$0.3 &  8.82\footnotesize$\pm$0.18  & 42.01\footnotesize$\pm$1.84 & 104.11\footnotesize$\pm$10.33 & 38.89\footnotesize$\pm$2.01 \\
    & B3 & 0.80\footnotesize$\pm$0.01 & 2.9\footnotesize$\pm$0.5 & 10.45\footnotesize$\pm$0.16 & 45.90\footnotesize$\pm$1.01 & 104.46\footnotesize$\pm$7.61 & 42.40\footnotesize$\pm$1.14\\
    & C & 0.79\footnotesize$\pm$0.05 & 4.7\footnotesize$\pm$1.5 & 8.82\footnotesize$\pm$0.18 & 74.71\footnotesize$\pm$21.85 & 106.04\footnotesize$\pm$22.54 & 72.64\footnotesize$\pm$24.42\\
	\cline{2-8}
	\parbox[t][][t]{10mm}{\multirow{2}{*}{\rotatebox[origin=c]{0}{{\specialcell[c]{CelebA\\Smiling}}}}}
	
 	& A & 0.98\footnotesize$\pm$0.01 & 71.5\footnotesize$\pm$6.8 & 0.24\footnotesize$\pm$0.13 & 411.09\footnotesize$\pm$22.61 & 381.26\footnotesize$\pm$24.01 & 370.69\footnotesize$\pm$19.59 \\
    & \textbf{B2} & 0.86\footnotesize$\pm$0.06 & 15.8\footnotesize$\pm$9.2 & 4.22\footnotesize$\pm$2.17 & 146.58\footnotesize$\pm$43.18 & 217.36\footnotesize$\pm$96.52 & 178.23\footnotesize$\pm$75.84 \\
	\cline{2-8}
     & \multicolumn{7}{c}{\textbf{EBPE \cite{Singla2020Explanation}}} \\
   \cline{2-8}
	\parbox[t][][t]{10mm}{\multirow{8}{*}{\rotatebox[origin=c]{0}{{\specialcell[c]{CelebA\\Smiling}}}}}
	
 	& A & 0.91\footnotesize$\pm$0.01 & 26.3\footnotesize$\pm$1.3 & 1.64\footnotesize$\pm$0.19 & 347.88\footnotesize$\pm$190.56 & 463.90\footnotesize$\pm$35.91 & 425.76\footnotesize$\pm$15.80 \\
    & B1 & 0.86\footnotesize$\pm$0.02 & 20.0\footnotesize$\pm$4.6 & 3.21\footnotesize$\pm$0.92 & 217.41\footnotesize$\pm$0.76 & 252.24\footnotesize$\pm$12.78 & 241.99\footnotesize$\pm$8.13 \\
    & B2 & 0.00\footnotesize$\pm$0.01 & 0.0\footnotesize$\pm$0.0 & 32.18\footnotesize$\pm$2.14 & 372.58\footnotesize$\pm$19.57 & n/a & 372.58\footnotesize$\pm$19.57 \\
    & B3 & 0.99\footnotesize$\pm$0.01 & 80.8\footnotesize$\pm$3.3 & 0.09\footnotesize$\pm$0.05 & 387.15\footnotesize$\pm$11.34 & 403.74\footnotesize$\pm$9.36 & 391.91\footnotesize$\pm$16.79  \\
    & \textbf{C} & 0.94\footnotesize$\pm$0.01 & 41.9\footnotesize$\pm$3.1 & 1.22\footnotesize$\pm$0.16 & 193.99\footnotesize$\pm$20.44 & 191.90\footnotesize$\pm$20.66 & 191.67\footnotesize$\pm$20.51 \\
    & D1 & 0.97 & 54.39 & 0.56 & 185.17 & 185.16 & 184.96 \\
    & D2 & 0.96 & 53.62 & 0.57 & 148.02 & 146.75 & 146.73 \\
    & D3 & 0.96 & 53.71 & 0.57 & 120.45 & 120.14 & 120.00 \\

	\cline{2-8}
     & \multicolumn{7}{c}{\textbf{C3LT \cite{Khorram_2022_CVPR}}} \\
   \cline{2-8}

  \parbox[t]{16mm}{\multirow{2}{*}{\rotatebox[origin=c]{0}{{\specialcell[c]{MNIST\\ Binary 0/1}}}}} 
 
    & A & 0.84\footnotesize$\pm$0.08  & 2.0\footnotesize$\pm$0.8 & 4.71\footnotesize$\pm$2.21 & 94.13\footnotesize$\pm$12.06 & 106.90\footnotesize$\pm$14.25 & 92.30\footnotesize$\pm$19.59\\
    & \textbf{B} & 0.89\footnotesize$\pm$0.03  & 3.6\footnotesize$\pm$0.8 &  6.32\footnotesize$\pm$1.39& 63.49\footnotesize$\pm$8.73 & 96.11\footnotesize$\pm$17.22 & 57.09\footnotesize$\pm$10.78\\
	\cline{2-8}
 	\parbox[t][][t]{10mm}{\multirow{2}{*}{\rotatebox[origin=c]{0}{{\specialcell[c]{CelebA\\Smiling}}}}}
	
 	& A & 0.89\footnotesize$\pm$0.04 & 3.5\footnotesize$\pm$2.5 & 3.52\footnotesize$\pm$0.71 & 122.29\footnotesize$\pm$12.06 & 138.82\footnotesize$\pm$15.33 & 133.13\footnotesize$\pm$13.56 \\
    & \textbf{B} & 0.90\footnotesize$\pm$0.01 & 11.8\footnotesize$\pm$5.5&  3.94\footnotesize$\pm$0.66 & 96.65\footnotesize$\pm$4.97 & 113.70\footnotesize$\pm$19.52 & 101.46\footnotesize$\pm$11.56\\
	\cline{2-8}
\end{tabular}}
\end{table*}
\begin{table}[!htbp]
\caption{Hyperparameter analysis of ProtoVAE. We report classifier's accuracy (ACC) and mean squared error (MSE) of the reconstructions. MSE is scaled by a factor of $10^2$. Mean values, including standard deviation, are reported over four training processes with different seeds. Configurations for experiments in \cref{tab:gdvaeEMvsIS} are \textbf{bolded}.}
\label{tab:protoVAEParams}
	\centering
		\renewcommand{\arraystretch}{0.9}
		\begin{tabular}{cccc}
			\cline{2-4}
			&Setting	&  $ACC\%$ $\uparrow$& $MSE$ $\downarrow$ \\
		\cline{2-4}
		\parbox[t]{0mm}{\multirow{4}{*}{\rotatebox[origin=c]{90}{MNIST}}} 
	&\textbf{A: ProtoVAE} \cite{Gautam22ProtoVAE} & {99.1\footnotesize$\bm \pm$0.17} & 1.51\footnotesize$\pm$0.23 \\
		&B1  & 98.0\footnotesize$\pm$0.12 & {1.00\footnotesize$\bm \pm$0.01} \\ 
  	    &B2     & 94.8\footnotesize$\pm$0.37 & 1.01\footnotesize$\pm$0.01   \\ 
		&B3& 94.8\footnotesize$\pm$0.31 & 0.99\footnotesize$\pm$0.01   		\\ 
    \cline{2-4}
	\parbox[t]{0mm}{\multirow{4}{*}{\rotatebox[origin=c]{90}{CIFAR-10}}} 
	&\textbf{A: ProtoVAE} \cite{Gautam22ProtoVAE} & {76.6\footnotesize$\bm \pm$0.35} & 2.69\footnotesize$\pm$0.02 \\ 
	&B1 & 58.4\footnotesize$\pm$1.31 & 0.92\footnotesize$\pm$0.02   			 \\ 
    &B2   & 31.2\footnotesize$\pm$2.18 & {0.85\footnotesize$\bm \pm$0.05}   			 \\ 
	&B3  & {30.0}\footnotesize$\pm$3.25 & 0.37\footnotesize$\pm$0.02   		\\ 
\cline{2-4}
\parbox[t]{9mm}{\multirow{4}{*}{\rotatebox[origin=c]{90}{\specialcell[t]{CelebA\\Gender}}}}
&\textbf{A: ProtoVAE} \cite{Gautam22ProtoVAE} & 96.6\footnotesize$\pm$0.24 & 1.32\footnotesize$\pm$0.10 \\ 
&B1 &               96.0\footnotesize$\pm$0.22 & 0.76\footnotesize$\pm$0.02 \\ 
&B2 &               88.9\footnotesize$\pm$0.80 & 0.75\footnotesize$\pm$0.05 \\ 
&B3  &               85.8\footnotesize$\pm$0.50 & 0.74\footnotesize$\pm$0.03 \\ 
\cline{2-4}
\end{tabular}

\end{table}
\begin{table*}[!t]
\caption{FID evaluation of GdVAE's CF explanations for different confidence ranges.} 
\label{tab:consistencyRanges}
\vspace{-0.1cm}
	\centering
    \small
		\renewcommand{\arraystretch}{1.0}
        \begin{tabular}{cl|ccc}
			\cline{2-5}  
        & \multicolumn{1}{l|}{\multirow{2}{*}{Method}} & \multicolumn{3}{c}{Realism (FID) $\downarrow$}\\
         &	&   $\widehat{\pdf}_\class\notin[0.1,0.9]$  &   $\widehat{\pdf}_\class \in[0.1,0.9]$  &   $\widehat{\pdf}_\class \in[0,1]$ \\
    	\cline{2-5}
    \parbox[t]{16mm}{\multirow{2}{*}{\rotatebox[origin=c]{0}{{\specialcell[c]{MNIST\\ Binary 0/1}}}}} 
        &\textbf{Ours} (global)  & 126.93\footnotesize$\pm$8.73 & 140.11\footnotesize$\pm$9.04 &  125.45\footnotesize$\pm$11.32 \\ 
       & \textbf{Ours} (local-L2) &  91.80\footnotesize$\pm$10.35 & 101.25\footnotesize$\pm$11.07 & 91.22\footnotesize$\pm$11.04 \\
    \cline{2-5}
 	\parbox[t][][t]{10mm}{\multirow{2}{*}{\rotatebox[origin=c]{0}{{\specialcell[c]{CelebA\\Smiling}}}}}
  
	&\textbf{Ours} (global)  & 118.40\footnotesize$\pm$5.15 & 138.79\footnotesize$\pm$6.11 & 128.93\footnotesize$\pm$4.94 \\
    &\textbf{Ours} (local-L2) & {86.01\footnotesize$\pm$2.60} & {86.59\footnotesize$\pm$2.47} & {85.52\footnotesize$\pm$2.37} \\
	\cline{2-5}
\end{tabular}
\vspace{-0.5cm}
\end{table*}

\noindent\textbf{ProtoVAE \cite{Gautam22ProtoVAE}.} The hyperparameter analysis for ProtoVAE is detailed in \cref{tab:protoVAEParams}. Setting A represents the original parameterization of the loss function as introduced by \cite{Gautam22ProtoVAE}. In an effort to enhance results, we modified the loss function based on our GdVAE settings, outlined in \cref{tab:initalTests}. Initially, we adjusted the scale of the reconstruction loss using $\pdf_\param(\rvxr|\rvyr,\rvzr)=\gaussian\left(\mu_\rvxr(\rvyr,\rvzr ;\param), \Sigma_\rvxr(\rvyr,\rvzr ;\param)\right)$, with $\Sigma_\rvxr:=\Sigma_\rvxr(\rvyr,\rvzr ;\param)=0.6^2\cdot I$ for likelihood calculation (B1), maintaining the original weighting of the other loss terms. Setting B2 involves adjusting the weight of the reconstruction and classification loss according to our GdVAE learning objective. Finally, setting B3, an extension of B2, incorporates a change in the learning rate from $0.001$ to the one used by our GdVAE, which is $0.0005$. As intended, we achieved a consistent reduction in reconstruction error; however, this improvement came at the cost of lower classification accuracy. Hence, we retained the parameterization from the original implementation.
 
\noindent\textbf{GANalyze \cite{Goetschalckx_2019_ICCV}.} All hyperparameter tuning for GANalyze was done on the MNIST binary dataset and the results can be found in \cref{tab:baselineModels}. In the initial setting (A), we employed the loss function from the original implementation, which included the normalization of counterfactuals. In the second setting (B1), we recreated the loss as described in the paper, omitting the normalization and achieving enhanced results. To further explore the efficacy of setting B1, we extended the training to 48 epochs (setting B2) instead of the original 24 epochs. 

\noindent\textbf{UDID \cite{voynov2020unsupervised}.} The hyperparameter tuning for UDID primarily focused on optimizing the proximity loss and selecting the appropriate classification loss. In setting A, we assessed the model with no proximity loss, defined as $\loss=\loss^{UD}+\gamma \cdot\loss_{prox}(\rvxr,\rvxr^\alpha)$, where $\gamma=0$. Settings B1, B2, and B3 were evaluated with a proximity loss weighted by $\gamma \in \{0.1,1.0,10.0\}$. In the final configuration, C, based on B2, we replaced the mean square error classification loss with cross-entropy loss. For the CelebA dataset, we revisited settings A and B2 from the MNIST experiments. Setting A yielded the highest accuracy and lowest MSE (consistency), but the generated counterfactuals lacked coherence and showed no resemblance to the input data, as indicated by the high FID score. The FID played a crucial role in selecting setting B2, which includes the proximity loss. 

\noindent\textbf{AttFind \cite{Lang_2021_ICCV}.} As AttFind was not part of the main paper's comparative study, we applied it directly to our GdVAE's latent space without optimization. Refer to \cref{tab:consistencySimple} for the results.

\begin{table}[!t]
\caption{Evaluation of CF explanations for the simple consistency task. Methods marked with a $\dagger$ exclusively induce class changes, rendering accuracy incomparable to other methods. These $\dagger$ methods closely adhere to the original implementation. We use ACC to assess the classification consistency for generated CFs. The Fr\'{e}chet Inception Distance (FID) is employed to gauge CF realism. Proximity is measured with MSE (scaled by $10^2$). Mean values, with standard deviation, are reported across four training runs with different seeds. It is evident that all methods marked with $\dagger$ achieve 100\% accuracy in altering the class of the query image. The methods without a $\dagger$ represent the modified versions discussed in the main paper. These methods enable users to pre-define a confidence value when generating CFs.} 
\label{tab:consistencySimple}

\vspace{-0.1cm}
	\centering
    \small
		\begin{tabular}{cccc}
         \cline{1-4}
 \multicolumn{4}{c}{{{MNIST - Binary 0/1}}}  \\
        \cline{1-4}
         \multicolumn{1}{c}{\multirow{2}{*}{Method}} & \multicolumn{1}{c}{Consistency} & \multicolumn{1}{c}{Realism (FID) $\downarrow$} &\multicolumn{1}{c}{Proximity} \\
         	& $ACC\%$ $\uparrow$& $\widehat{\pdf}_\class \in[0,1]$&$MSE$ $\downarrow$ \\
   \cline{1-4}
		AttFind$^\dagger$ \cite{Lang_2021_ICCV}  & 100.0\footnotesize$\pm$0.0 & 120.14\footnotesize$\pm$21.88& 11.92\footnotesize$\pm$6.69 \\  
  	C3LT$^\dagger$ \cite{Khorram_2022_CVPR}  & 100.0\footnotesize$\pm$0.0 &  88.67\footnotesize$\pm$11.01 & 14.92\footnotesize$\pm$0.85\\
		C3LT \cite{Khorram_2022_CVPR}  & 3.6\footnotesize$\pm$0.8 &  57.09\footnotesize$\pm$10.78& 5.83\footnotesize$\pm$1.47\\
 \cline{1-4}
  \multicolumn{4}{c}{{{CelebA - Smiling}}} \\
         \cline{1-4}
         \multicolumn{1}{c}{\multirow{2}{*}{Method}} & \multicolumn{1}{c}{Consistency} & \multicolumn{1}{c}{Realism (FID) $\downarrow$} &\multicolumn{1}{c}{Proximity} \\
         	& $ACC\%$ $\uparrow$& $\widehat{\pdf}_\class \in[0,1]$&$MSE$ $\downarrow$ \\
   		\cline{1-4}
    
		AttFind$^\dagger$ \cite{Lang_2021_ICCV} &  100.0\footnotesize$\pm$0.0 & 109.18\footnotesize$\pm$14.82& 2.32\footnotesize$\pm$1.37\\ 
        C3LT$^\dagger$ \cite{Khorram_2022_CVPR} & 100.0\footnotesize$\pm$0.0 & 171.72\footnotesize$\pm$7.97 & 8.59\footnotesize$\pm$0.84\\
		C3LT \cite{Khorram_2022_CVPR}  & 11.8\footnotesize$\pm$5.5&  101.46\footnotesize$\pm$11.56 & 3.97\footnotesize$\pm$0.86\\ 
\cline{1-4}
\end{tabular}
\vspace{-0.2cm}
\end{table}

\noindent\textbf{EBPE \cite{Singla2020Explanation}.} To evaluate EBPE, we experimented with different hyperparameters and activation functions in the decoder's output layer. The initial implementation, designed exclusively for CelebA, used the parameters detailed in \cref{tab:EBPEParams} ("Original") as our starting configuration. However, this configuration yielded suboptimal performance on both MNIST and CelebA. For MNIST, setting A from \cref{tab:EBPEParams} was chosen for its balance between consistency and realism.

EBPE encountered challenges in generating good reconstructions for the CelebA dataset with our backbone network and the original settings (see \cref{tab:EBPEParams}). To address this issue, we explored different activation functions to prevent image saturation, particularly as our GdVAE architecture employed a ReLU function in the output layer. In setting A, we employed a Sigmoid function, while method B1 utilized a hyperbolic tangent (tanh) function---the latter being used in the original GAN model by the authors of EBPE. Hence, all subsequent settings adopted the tanh function in the decoder architecture. Emphasis was placed on optimizing the reconstruction quality and FID scores by varying the weight of the conditional GAN loss (discriminator loss). Specifically, settings B1, B2, and B3 utilized weights of $\lambda_{cGAN}\in \{0.01, 0.1, 0.001\}$ for the discriminator loss.

Building on these findings, we enhanced the FID of version B1 by employing a higher weight of $\lambda_{rec}=\lambda_{cyc}=10^3$ for the reconstruction loss in setting C, deviating from the previously used weight of $10^2$. To validate our implementation, we used EBPE to explain a separate discriminative classifier. The objective was to assess whether further improvements could be achieved by extending the training time beyond the 24 epochs that was used by all other methods, as the original implementation employed 300 epochs. Consequently, methods D1, D2, and D3 were executed with 24, 48, and 96 epochs, respectively. To expedite training, we employed a single discriminative classifier instead of the GdVAE, and as a result, standard deviations are not reported. It's evident that EBPE requires more training time compared to the GdVAE, and with four times the training duration, it converges toward similar FID values as our global method. The results of this hyperparameter analysis are shown in \cref{tab:baselineModels}. 

\noindent\textbf{C3LT \cite{Khorram_2022_CVPR}.} 
We examined two loss function settings on MNIST and CelebA. Setting A employs the mean squared error, as utilized by GANalyze (\cref{eq:lossGANalyze}), while setting B is founded on the cross-entropy loss for $\loss_{cf}$. The results are summarized in \cref{tab:baselineModels}.

\begin{table*}[ht]
    \caption{Hyperparameters for EBPE.}
    \label{tab:EBPEParams}
    \centering
    \begin{tabular}{ccccccc}
    \cline{1-7}
    \multicolumn{7}{c}{{MNIST}}\\
    \cline{1-7}
    Setting  & $\lambda_{cGAN}$  & $\lambda_{rec}$ & $\lambda_{cyc}$ & $\lambda_{cfCls}$ & $\lambda_{recCls}$ & Activation\\
    \cline{1-7}
        A  & 0.01 & 1 & $10$ &10 & 0.1  & ReLU \\
    \cline{1-7}
   \multicolumn{7}{c}{{CelebA}}\\
    \cline{1-7}
    Setting  & $\lambda_{cGAN}$  & $\lambda_{rec}$ & $\lambda_{cyc}$ & $\lambda_{cfCls}$ & $\lambda_{recCls}$ & Activation\\
    \cline{1-7}
        Original & 1 & $10^2$ & $10^2$ & 1& 1& tanh \\
        \cdashline{1-7}
        A &  0.01 &  $10^2$ & $10^2$ & 1& 1&Sigmoid \\
        B1 & 0.01 &  $10^2$ & $10^2$ & 1& 1&tanh\\
        B2 & 0.1 & $10^2$ & $10^2$ & 1& 1&tanh\\
        B3 & 0.001& $10^2$ & $10^2$ & 1& 1& tanh\\
        C  & 0.01 & $10^3$ &$10^3$ & 1 &1  & tanh \\
    \cline{1-7}
    \end{tabular}
\end{table*}

\subsection{Qualitative Results}\label{supp:qualitativeResults}
\begin{figure*}[!t]
	\centering
		\begin{overpic}[grid=false, width=0.75\textwidth]{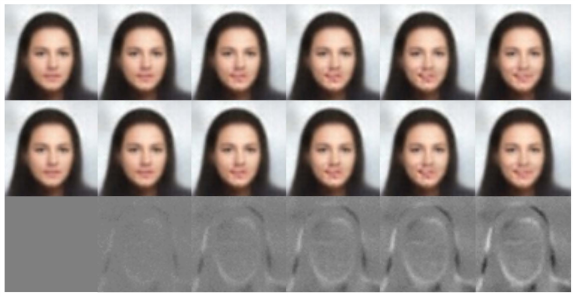}%
 		\put(1.5,47.8) {\textcolor{black}{a)}}%
 		\put(1.5,31.1) {\textcolor{black}{b)}}%
 		\put(1.5,14.4) {\textcolor{black}{c)}}%
 		\end{overpic}
\caption{{a) Local-L2 CFs, b) Local-M CFs, and c) the difference, with white and black indicating a deviation of approximately $\pm3\%$.}}
	\label{fig:L2vsM}
\end{figure*}

\begin{figure*}[!ht]
  \centering 
 	\begin{subfigure}{0.8\textwidth}
     \centering
   		\begin{overpic}[grid=false, width=1.0\textwidth]{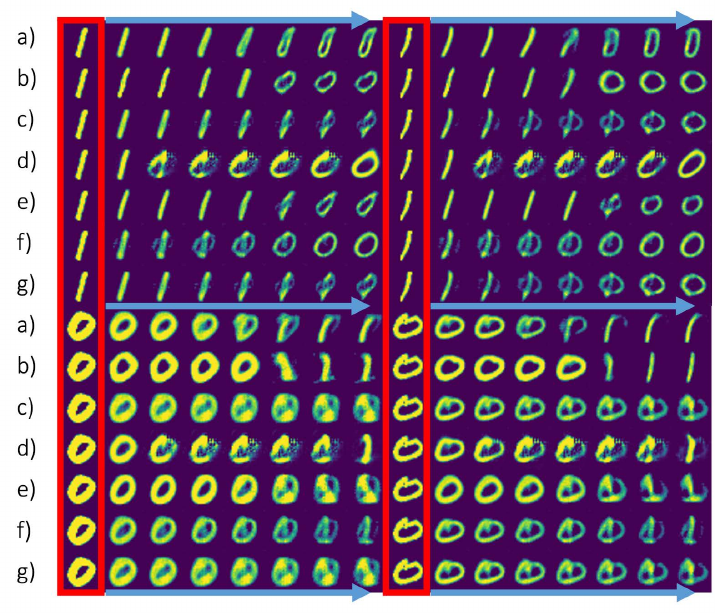}%
 		
 		\put(11,84.5) {\color{black} \Large {$\rvxr$} }%
 		\put(56,84.5) {\color{black} \Large {$\rvxr$} }%
 		\put(34,84.5) {\color{black} \Large {$\rvxr^\delta$} }%
 		\put(79.5,84.5) {\color{black} \Large {$\rvxr^\delta$} }%
 	\end{overpic}
 	\end{subfigure}
\caption{MNIST CFs.\,We generate CFs ($\rvxr^\delta$) linearly for the input, with decreasing confidence for the true class from left to right. On the leftmost side of each section, $\rvxr$ denotes the input. We generate samples for $\pdf_c=[0.99, 0.95. 0.75, 0.5, 0.25,0.05,0.01]$. a) GANalyze, b) UDID, c) ECINN, d) EBPE, e) C3LT, f) Ours (global), g) Ours (local). }
 	\label{figsupp:counterExamplesMNIST}
 \end{figure*}

\begin{figure*}[!ht]
  \centering 
 	\begin{subfigure}{0.85\textwidth}
 
     \centering
 		\begin{overpic}[grid=false, width=1.0\textwidth]{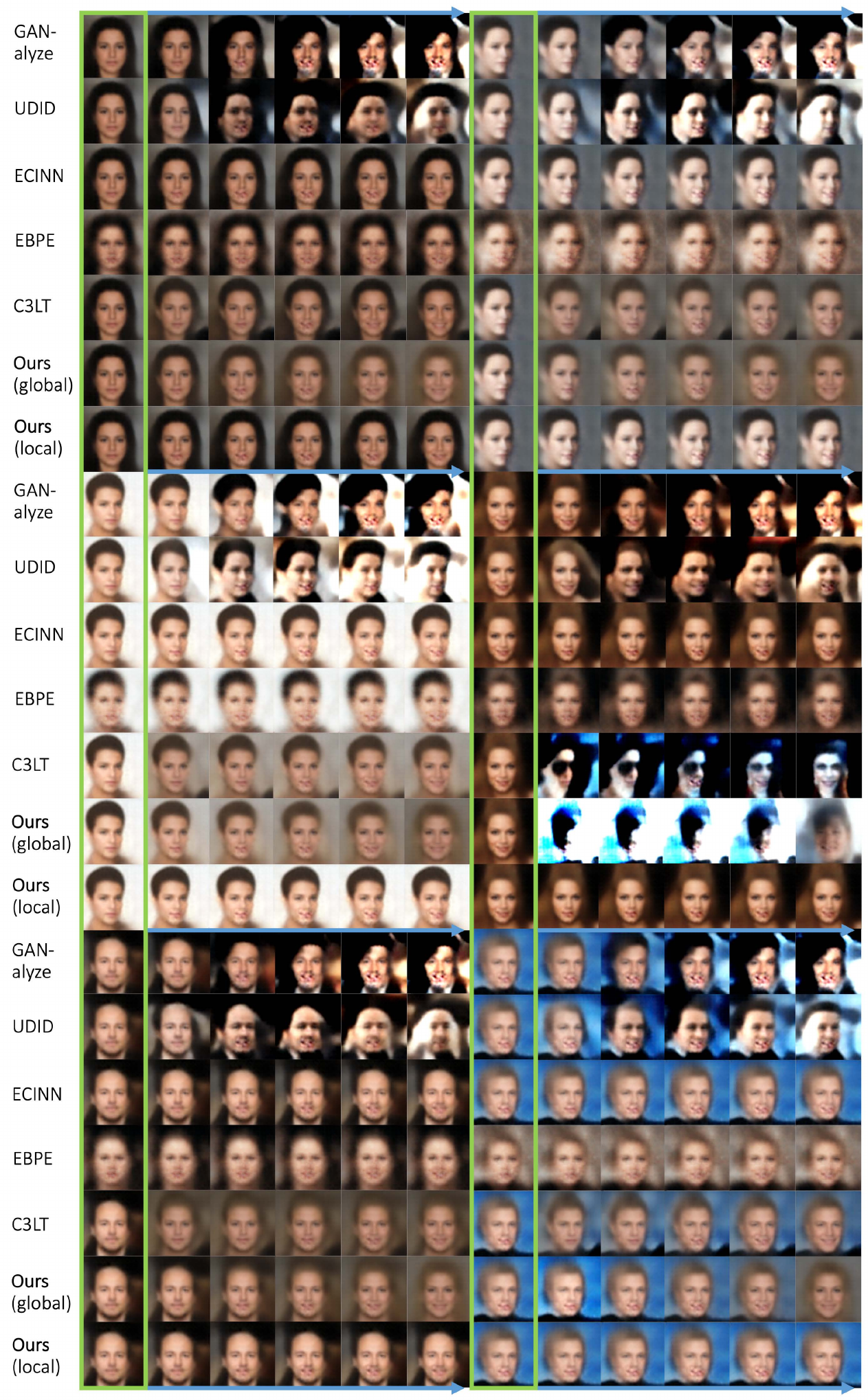}%
 	\end{overpic}
 	\end{subfigure}
\caption{CelebA CFs}
\label{figsupp:counterExamplesCelebA}
\end{figure*}

\begin{figure*}[!ht]
  \centering 
 	\begin{subfigure}{0.9\textwidth}
 
     \centering
   		\begin{overpic}[grid=false, width=1.0\textwidth]{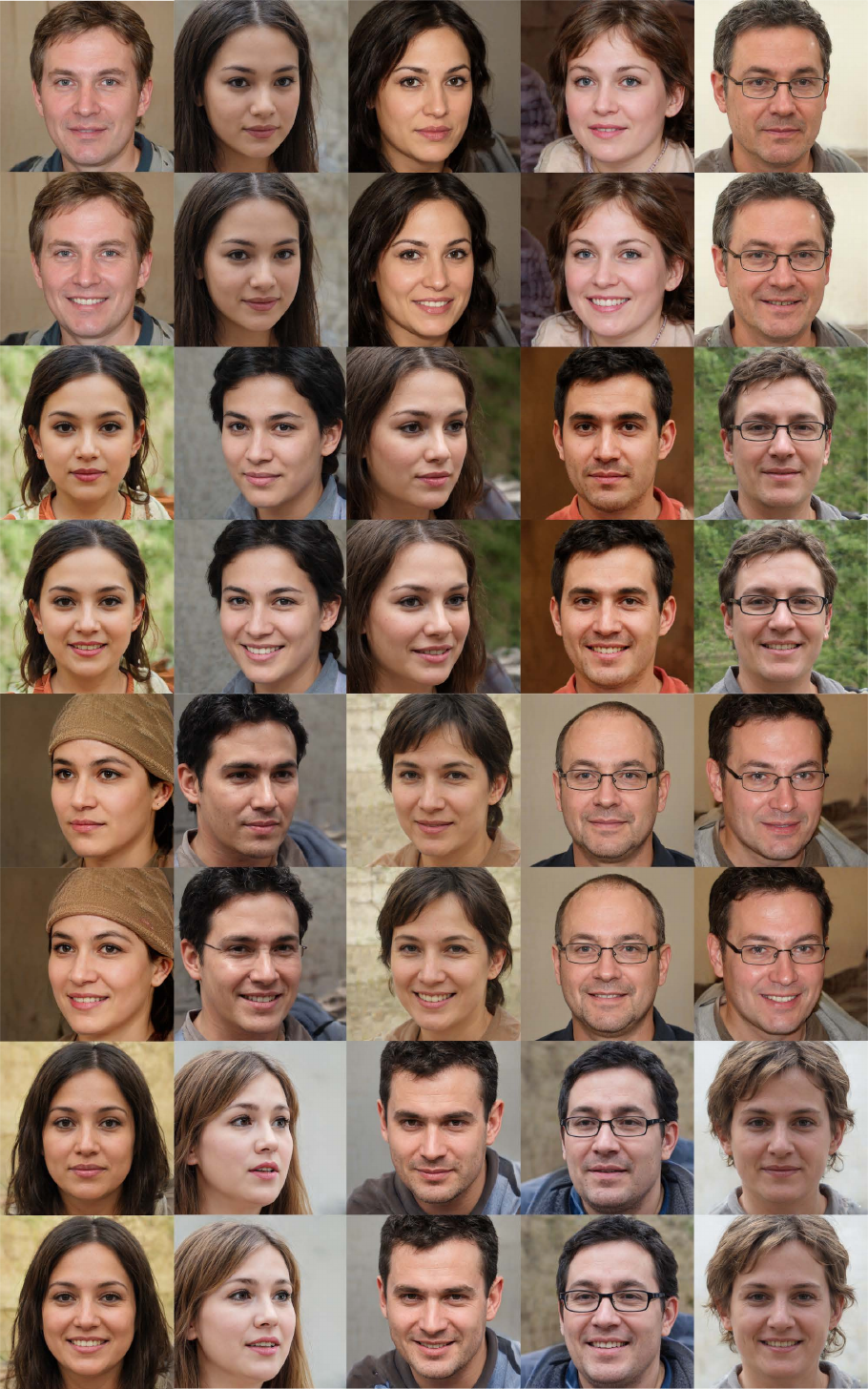}%
 	\end{overpic}
 	\end{subfigure}
\caption{FFHQ CFs}
 	\label{fig:suppFFHQ}
 \end{figure*}

\begin{figure*}[!ht]
  \centering 
 	\begin{subfigure}{0.8\textwidth}
 
     \centering
  
 		\begin{overpic}[grid=false, width=1.0\textwidth]{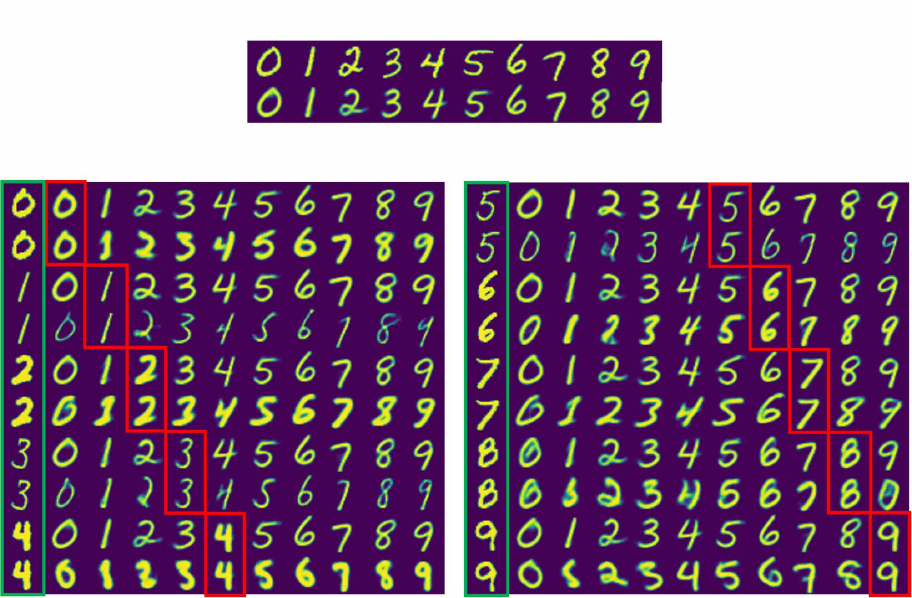}%

 		\put(22,62.5) {\color{black} {Prototypes and closest input image $x$} }%
 		\put(38,47) {\color{black} {Counterfactuals} }%
 		\put(22,58) {\color{black} {$\rvxr$} }%
 		\put(20,54) {\color{black} {$\mean_{\rvxr|y}$} }%
 		
 	\end{overpic}
 	\end{subfigure}
\caption{MNIST multi-class CFs.\,CFs for the simpler consistency task by swapping the logits of the predicted and counterfactual classes. The green rectangle indicates the input image, while the red rectangles highlight the image reconstructions. Global (top) and local (bottom) CFs are each positioned to the left and right of the reconstructions, respectively. The local CFs maintain key image characteristics, such as line thickness. This CF strategy changes class predictions $100\%$ of the time for the global method, and $99\%$ of the time for the local method.}
 	\label{fig:suppMNISTMulti}
 \end{figure*}

\begin{figure*}[!ht]
  \centering 
 	\begin{subfigure}{0.85\textwidth}
 
     \centering
  		\begin{overpic}[grid=false, width=1.0\textwidth]{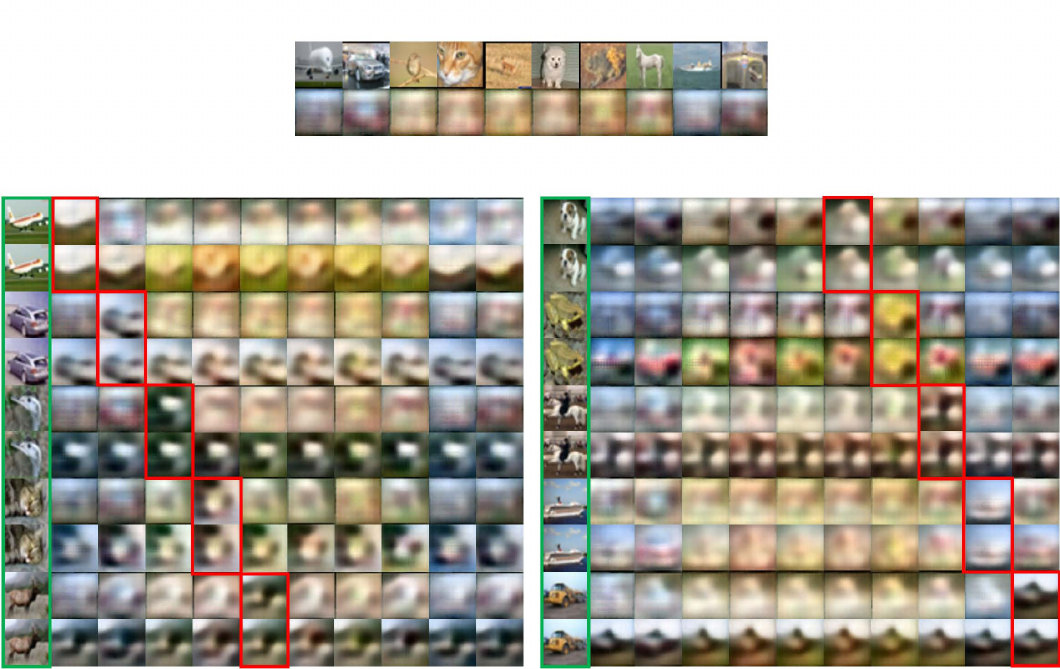}%
		\put(25,61) {\color{black} {Prototypes and closest input image $x$} }%
 		\put(40,46) {\color{black} {Counterfactuals} }%
 		\put(23.5,56) {\color{black} {$\rvxr$} }%
 		\put(21.5,52) {\color{black} {$\mean_{\rvxr|y}$} }%
 	\end{overpic}
 
 	\end{subfigure}
\caption{CIFAR-10 multi-class CFs.\,CFs by swapping the logits of the predicted and counterfactual classes. The green rectangle indicates the input image, while the red rectangles highlight the image reconstructions. Global (top) and local (bottom) CFs are each positioned to the left and right of the reconstructions, respectively. The CFs are generated primarily through color adjustments and minor shape adaptations. This CF strategy changes class predictions $97\%$ of the time for the global method, and $89\%$ of the time for the local method.}
 	\label{fig:suppCIFARMulti}
 \end{figure*}

Additional results: MNIST in \cref{figsupp:counterExamplesMNIST}, CelebA in \cref{figsupp:counterExamplesCelebA}, FFHQ in \cref{fig:suppFFHQ}, and multi-class CFs for MNIST and CIFAR in \cref{fig:suppMNISTMulti,fig:suppCIFARMulti}. All local explanations were generated using the L2-based method. A comparison of local-L2 and local-M explanations is provided in \cref{fig:L2vsM}.




\end{document}